%% file: main.tex
\documentclass{article}

     \PassOptionsToPackage{numbers, sort&compress}{natbib}

\usepackage[final]{neurips_2021}

\usepackage[utf8]{inputenc} 
\usepackage[T1]{fontenc}    
\usepackage{hyperref}       
\usepackage{url}            
\usepackage{booktabs}       
\usepackage{amsfonts}       
\usepackage{amsmath}
\usepackage{nicefrac}       
\usepackage{microtype}      
\usepackage{xcolor}         
\usepackage{comment}
\usepackage{tabularx}
\usepackage{multirow, multicol}

\usepackage{graphicx}
\usepackage[normalem]{ulem}
\usepackage{algorithm}
\usepackage{enumitem}
\usepackage{algpseudocode}
\usepackage[normalem]{ulem}
\usepackage{cancel}
\usepackage{soul}

\newcommand{\etal}{\textit{et al.}}
\newcommand{\ie}{\textit{ie.}~}
\newcommand{\eg}{\textit{eg.,}~}

\title{How Modular Should Neural Module Networks Be \\
for Systematic Generalization? 
}

\author{Vanessa D'Amario \textsuperscript{\rm 1, 3} \And 
  Tomotake Sasaki \textsuperscript{\rm 2, 3} \And 
  Xavier Boix \textsuperscript{\rm 1, 3} \AND  
   \textsuperscript{\rm 1}  \mbox{\rm Massachusetts Institute of Technology, USA \;\;\; \textsuperscript{\rm 2} Fujitsu Limited, Japan} \\    
  \textsuperscript{\rm 3} Center for Brains, Minds and Machines, USA\\
  \texttt{vanessad@mit.edu, tomotake.sasaki@fujitsu.com, xboix@mit.edu} 
}

\begin{document}

\maketitle

\begin{abstract}

Neural Module Networks (NMNs) aim at Visual Question Answering (VQA) via
composition of modules that tackle a sub-task. NMNs are a promising strategy to
achieve systematic generalization, \ie overcoming biasing factors in the training
distribution. However, the aspects of NMNs that facilitate systematic generalization
are not fully understood. In this paper, we demonstrate that the degree
of modularity of the NMN have large influence on systematic generalization. In
a series of experiments on three VQA datasets (VQA-MNIST,
SQOOP, and CLEVR-CoGenT), our results reveal that tuning the degree of modularity, especially at the image encoder stage, reaches substantially
higher systematic generalization. These findings lead to new NMN architectures
that outperform previous ones in terms of systematic generalization.

\end{abstract}

\input{01_intro}

\input{02_specialization}

\input{03_dataset}
\input{04_results}

\input{05_clevr}

\input{06_conclusions}

\begin{ack}
We would like to thank Moyuru Yamada, Hisanao Akima, Pawan Sinha and Tomaso Poggio  for useful discussions and insightful advice provided during this project, and Ramdas Pillai and Atsushi Kajita from Fixstars Solutions for their assistance executing the  large scale  computational experiments.
This work has been supported by the Center for Brains, Minds and Machines (funded by NSF STC award CCF-1231216), the R01EY020517 grant from the National Eye Institute (NIH) and Fujitsu Limited (Contract No. 40008819 and 40009105). 
\end{ack}

\section*{Code and Data Availability}
Code and data can be found in the following github repository: \\ \url{https://github.com/vanessadamario/understanding_reasoning.git}.

{
\small
\bibliography{bibliography}
}

\clearpage
\appendix


\renewcommand{\thetable}{App.\arabic{table}}%
\renewcommand{\theHtable}{arabicsection.\thetable}
\renewcommand\thefigure{App.\arabic{figure}}    
\renewcommand{\theHfigure}{arabicsection.\thefigure}
\renewcommand{\theequation}{App.\arabic{equation}}%
\renewcommand{\theHequation}{arabicsection.\theequation}
\setcounter{equation}{0}
\setcounter{table}{0}
\setcounter{figure}{0}

\input{A1_appendix_data_generation}

\input{A2_more_networks}
\input{A3_cogent}

\end{document}

%% file: 01_intro.tex
\section{Introduction}

The combinatorial nature of vision \cite{krishna2017visual} is an open challenge for learning machines tackling Visual Question Answering (VQA)~\cite{johnson2017clevr, antol2015vqa,  agrawal2018don, bahdanau2019closure, teney2020value, kafle2017visual}. The amount of possible combinations of object categories, attributes, relations, context and visual tasks is exponentially large, and any training set contains a limited amount of such combinations. The ability to generalize to combinations not included in the training set, \ie the so-called \emph{systematic generalization}~\cite{lake2018generalization, loula2018rearranging, fodor1988connectionism}, is a hallmark of intelligence.

Recent research has shown that Neural Module Networks (NMNs) are a promising approach for systematic generalization in VQA~\cite{bahdanau2018systematic, bahdanau2019closure}.
NMNs infer a program layout containing a sequence of sub-tasks, where each sub-task is associated to a module consisting of a neural network~\cite{andreas2016neural, johnson2017inferring}.

The aspects that facilitate systematic generalization in NMNs are not fully understood. Recent works demonstrate that modular approaches are helpful for object recognition tasks with novel combinations of category and viewpoints~\cite{madan2020capability} and also novel combinations of visual attributes~\cite{purushwalkam2019task}. Bahdanau~\etal~\cite{bahdanau2018systematic} have demonstrated the importance of using an appropriate program layout for systematic generalization, and Bahdanau~\etal~\cite{bahdanau2019closure} in another paper have shown that the module's network architecture also impacts systematic generalization. These results highlight that using modules to tackle sub-tasks and using compositions of these modules facilitates systematic generalization. However, there is a central question at the heart of modular approaches that remains largely unaddressed: what sub-tasks should be tackled by a module? Namely, is it best for systematic generalization to use a large degree of modularity such that modules tackle very specific sub-tasks, or is it best to use a smaller degree of modularity?

In this paper, we investigate what sub-tasks a module should perform in order to facilitate systematic generalization in VQA. We analyse different degrees of modularity as in the example of Figure~\ref{fig:ood_modules}, where NMNs could have three different degrees of modularity: (i) a unique module for tackling all sub-tasks, (ii) modules tackling groups of sub-tasks, or (iii) many modules each one for tackling sub-tasks at the highest degree of modularity. Also, we investigate the degree of modularity at all the stages of the network, as NMNs have a structure divided in three stages: an \emph{image encoder} for the extraction of visual features, a set of \emph{intermediate modules} arranged based on the program layout, and a \emph{classifier} that provides the output answer.

We examined the degree of modularity in three families of VQA datasets: VQA-MNIST, the SQOOP dataset~\cite{bahdanau2018systematic}, and the Compositional Generalization Test (CoGenT) of CLEVR~\cite{johnson2017clevr}. 
Our results reveal two main findings that are consistent across the tested datasets: (i) an intermediate degree of modularity by grouping sub-tasks leads to much higher systematic generalization
than for the NMNs' modules
introduced in previous works, and 
(ii) modularity is mostly effective when
is defined at the image encoder stage, which is rarely done in the literature. Furthermore, we show that these findings are directly applicable to improve the systematic generalization of state-of-the-art NMNs.

%% file: 02_specialization.tex
\section{Libraries of Modules}\label{sec:specialization} 

NMNs consist of three components: a program generator, a library of modules and an execution engine. The program generator translates VQA questions into program layouts, consisting of compositions of sub-tasks. The library of modules contains neural networks that perform sub-tasks, and these modules are used by the execution engine to put the program layout into action.

\begin{figure*}[!t]
    \centering
    \begin{tabular}{c}
        \includegraphics[width=1\linewidth]{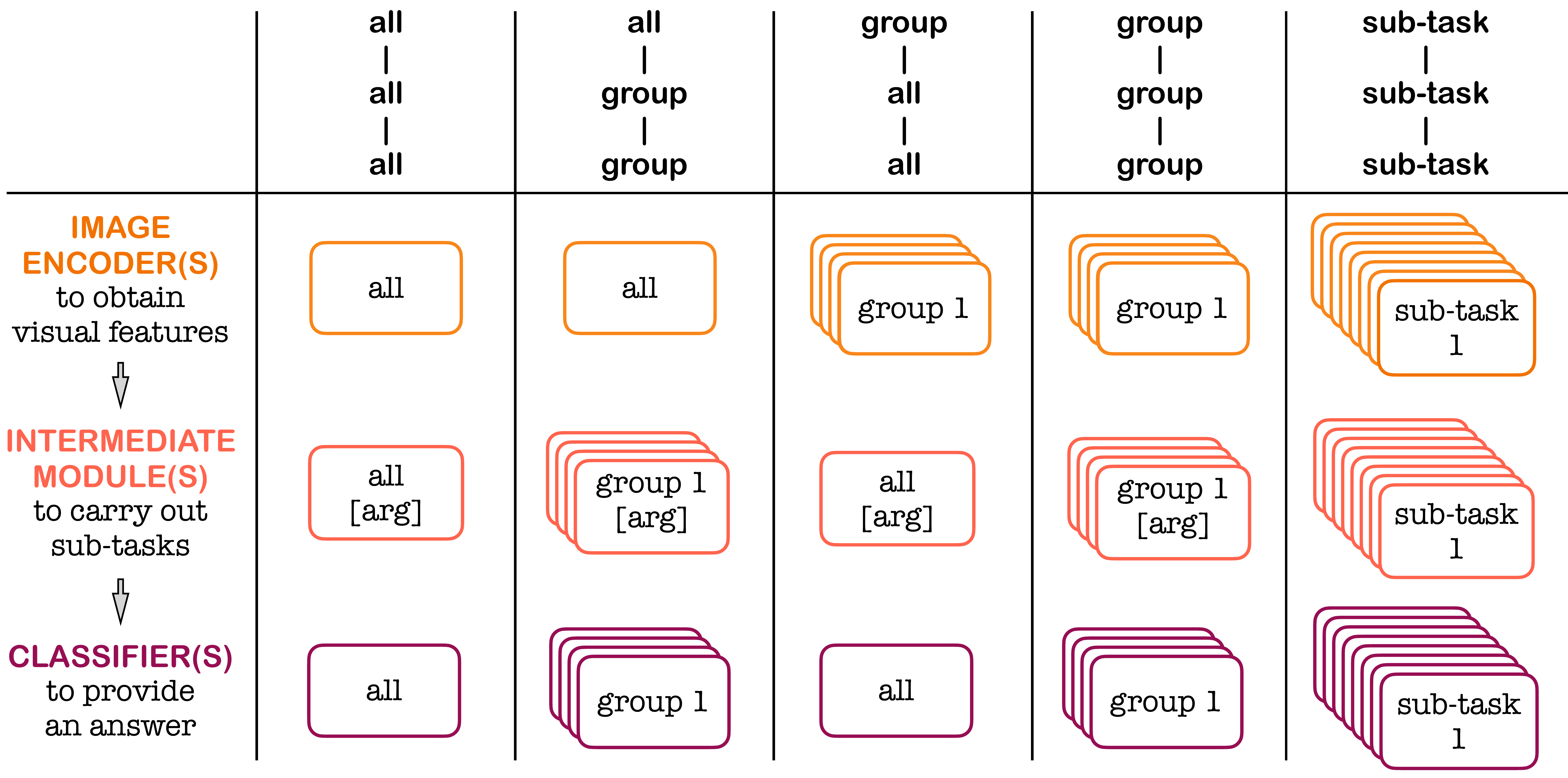} \\
        (a) \vspace{.7cm}\\
        \includegraphics[width=1\linewidth]{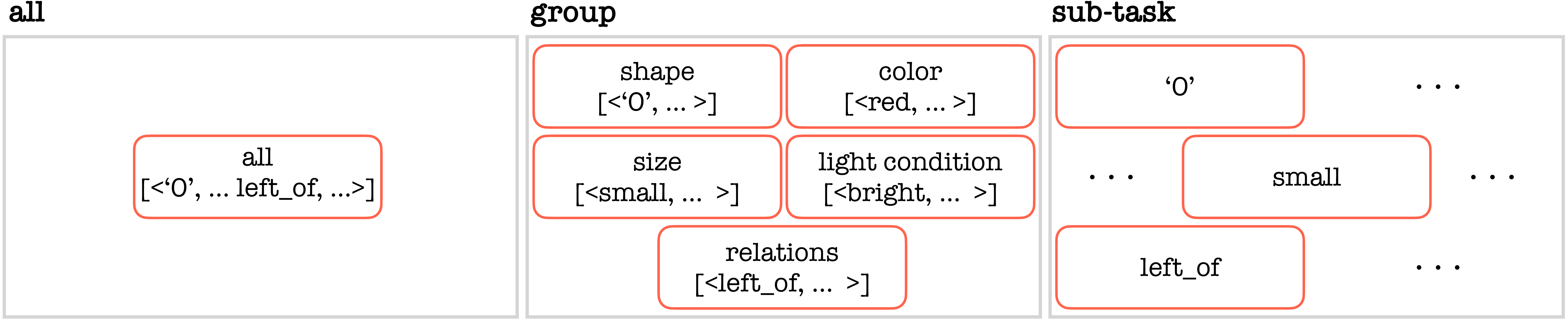} \\
        (b) \vspace{.7cm}\\
    \hspace{-0.2cm}\includegraphics[width=1\linewidth]{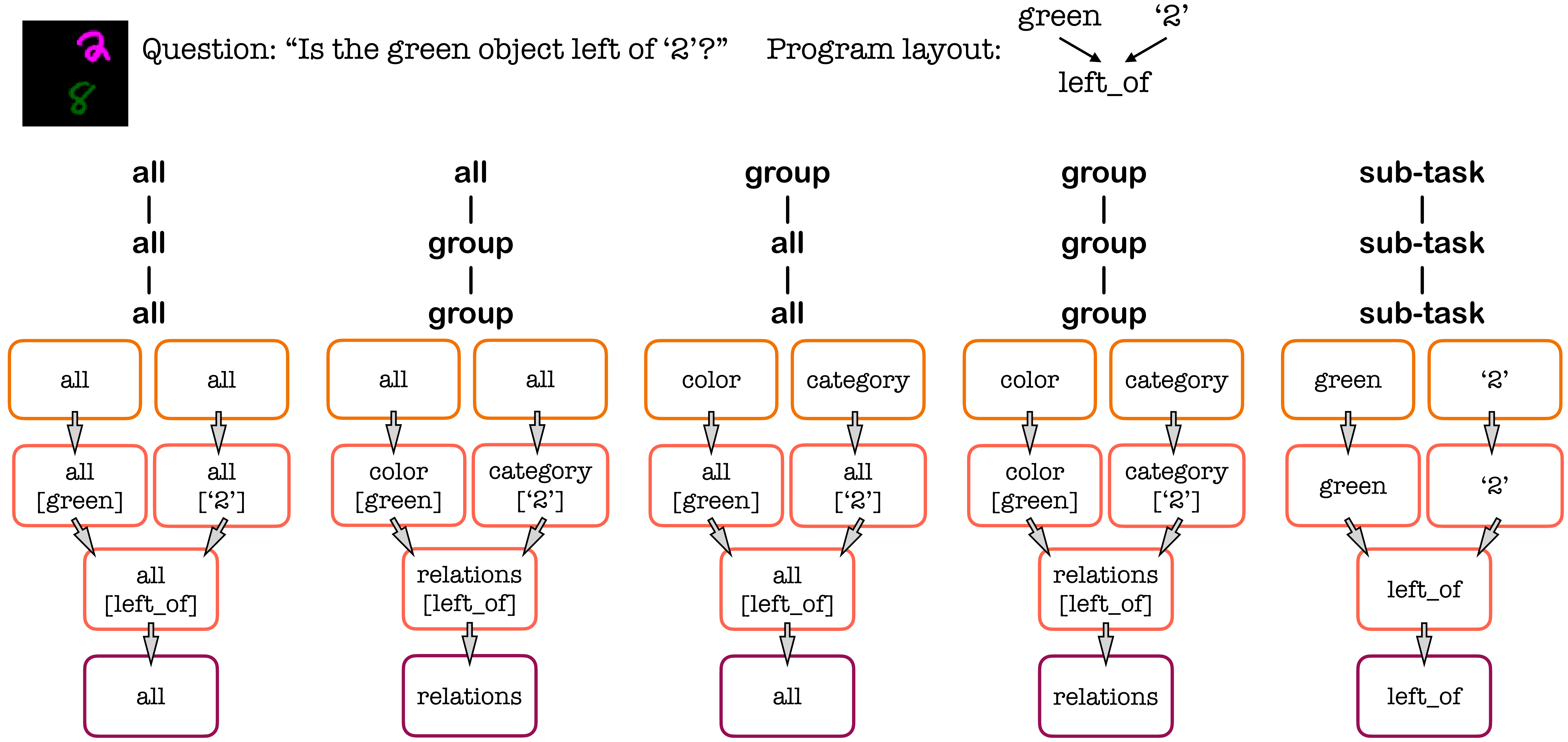} \\
        (c) \\
    \end{tabular}
    \caption{\emph{Libraries of modules.} (a) Example of five libraries with different degrees of modularity (\emph{all}, \emph{group}, and \emph{sub-task}) analyzed in this paper. Some modules have additional input arguments to indicate specifics of the sub-task.  (b) Intermediate modules from libraries with \emph{all}, \emph{group}, and \emph{sub-task} degrees of modularity, for the VQA-MNIST dataset. (c) Composition of modules leveraging on the libraries in (a), given the question \emph{``Is the green object left of `2'?''} and an image containing a pair of objects (the former identified by its color, the latter by its category).}
    \label{fig:ood_modules}
\end{figure*}

Given that the use of a correct program layout plays a fundamental role to achieve systematic generalization with NMNs~\cite{johnson2017inferring,bahdanau2018systematic}, we follow the approach in previous works that fix the program layout to the ground-truth program, such that it does not interfere in the analysis of the library~\cite{bahdanau2018systematic, bahdanau2019closure}. Thus, we follow a reductionist approach, which facilitates gaining an understanding of the crucial aspects of NMNs in systematic generalization. As we gain such understanding, we will be in a position to study the interactions between these aspects in future works.


 Here, we aim at understanding whether the choice of a library with a specific degree of modularity facilitates systematic generalization in NMNs.
Figure~\ref{fig:ood_modules}a shows 
examples of several libraries of modules (one library per column), characterized by different degrees of modularity at different stages of the network.
Each row identifies a stage in the network, while the use of single or multiple modules identifies the degree of modularity at each stage. In particular, a library with a single module shared across all sub-tasks is called \emph{all}.
A library with modules tackling groups of sub-tasks is called \emph{group}. Finally, a library with modules that tackle each single sub-task is called \emph{sub-task}. These different ways of defining the degree of modularity in the library are detailed in Figure~\ref{fig:ood_modules}b. 

Note that in the intermediate stage, it is sometimes necessary to use modules that operate differently depending on an input argument related to the sub-task to be performed by the module. For example, a module whose sub-task is to determine whether an object is of a specific color or not, could be implemented by a module specialized to detect colors with an input argument that indicates the color to be detected. It could also be implemented by a module specialized to detect a specific color, which would not require an additional input argument.

In Figure~\ref{fig:ood_modules}c, we provide an example of usage of the libraries given the question \emph{``Is the green object left of `2'?''}.  
The ground-truth program layout is a tree structure with two leaves and a root that identify respectively the pair of objects and the spatial relation. In the figure we depict how the same program layout can be implemented with libraries with different degrees of modularity. Note that the degree of modularity does not change the final number of modules in the program layout, but the number of modules in the library.

In the following, we first introduce the definitions of the modules in the library
for the image encoder and the classifier stages, and then we introduce existing implementations of intermediate modules and propose a new one.

\subsection{Libraries of Modules at the Image Encoder and Classifier Stages}

The image encoder is usually a single convolutional neural network module  common to all sub-tasks (first two columns of Figure~\ref{fig:ood_modules}a). 
We analyze  image encoders defined per group of sub-tasks, as in the third and forth columns in Figure~\ref{fig:ood_modules}a. 
The library with maximum degree of modularity at the image encoder stage corresponds to having an image encoder per sub-task of the VQA problem, as depicted in the last column of Figure~\ref{fig:ood_modules}a.

For a VQA task with binary answers, libraries with different degrees of modularity can be equally defined on the classifier. 

\subsection{Libraries of Modules  at the Intermediate Modules Stage}

The architecture of the intermediate modules varies across the literature. We consider two architectures which previous works \cite{bahdanau2018systematic, bahdanau2019closure} have focused on: the \emph{Find}~\cite{hu2017learning} and the \emph{Residual}~\cite{johnson2017inferring} modules. 
Let $f(k, s_x, s_y)$ be the function that represents an intermediate module, where each sub-task is denoted by an index $k$ that represents the input argument of the module, and ($s_x$, $s_y$) are the data inputs to the module, which can be the representation of the input image or the output of the precedent intermediate modules according to the program layout. If a module requires a single input for the given sub-task, $s_y$ is the output of the image encoder given a null input image. 

In the following, we show that the \emph{Find module} can be interpreted as a library with a \emph{single intermediate module} shared across all sub-tasks, with the neural representation modulated by an embedding related to the sub-task. Then, we show that the \emph{Residual module} corresponds to a library with maximum degree of modularity, 
in which each sub-task corresponds a single intermediate module. Finally, we introduce a new module  based on these two, which has an intermediate degree of modularity. 

\noindent \paragraph{A single intermediate module (aka. Find module~\cite{hu2017learning}).}
The definition is the following:
\begin{align}
    & \gamma_k = \text{Embedding}(k),  \\ 
    & f(k, s_x, s_y) = \text{ReLU}(W_1 * (\gamma_k \odot \text{ReLU}(W_2 * [s_x; s_y] + b_2)) + b_1), \label{eq:modulation}
\end{align}
where $[s_x; s_y]$ is the concatenation of the two inputs and $[W_1;b_1;W_2;b_2]$ are the weights and bias terms of the convolutional layer. 
These weights and bias terms are the same across all sub-tasks,~\ie all the sub-tasks are tackled with the same module.
The module performs a sub-task by ``modulating'' the neural activity via the element-wise product involving $\gamma_k$. 
Figure~\ref{fig:ood_modules}b depicts the Find module at the first column.

\noindent \paragraph{One module per sub-task (aka. Residual module~\cite{johnson2017inferring}).}
This represents a library with a much finer degree of modularity than the previous one. For each sub-task $k$, we have an independent set of convolutional weights indexed by $k$,~\ie $[W_1^k; b_1^k; W_2^k; b_2^k; W_3^k; b_3^k]$, as in the following:
\begin{align}
    &\tilde{s}_k = \text{ReLU}(W_3^k*[s_x; s_y] + b_3^k),   \\
    & f(k, s_x, s_y) = \text{ReLU}(\tilde{s}_k + W_1^{k} * \text{ReLU}(W_2^k * \tilde{s}_k + b_2^k) + b_1^k).
    \label{eq:residual}
\end{align}
Note that here the mechanism to tackle a sub-task is to use an independent set of weights per sub-task, as depicted in Figure~\ref{fig:ood_modules}b, at the third column. 

\noindent \paragraph{One module per group of sub-tasks.}
We introduce a new library of modules that has a degree of modularity halfway between the two presented above (depicted in Figure~\ref{fig:ood_modules}b, second column). 
Each modules tackles a group of sub-tasks via a Find module.    We use $[W^g_1;b^g_1;W^g_2;b^g_2]$ to denote the weights and bias terms of a module for the group of sub-tasks indexed by $g$. Let $g=\mbox{Group}(k)$ be a mapping from the sub-task $k$ to the index of the group, $g$. Thus, this library of modules is defined as follows:
\begin{align}
    & g=\mbox{Group}(k), \;\; \gamma_k = \text{Embedding}(k), \label{eq:def_group} \\ 
    & f(k, s_x, s_y) = \text{ReLU}(W^g_1 * (\gamma_k \odot \text{ReLU}(W^g_2 * [s_x, s_y] + b^g_2)) + b^g_1).  \label{eq:spec_type}
\end{align}
Note that this library of modules uses both Find and Residual architectures to tackle a sub-task---the division of sub-tasks in groups of separated weights reflects the design of the Residual architecture, while the modulation mechanism that allows the modules to adjust their representation among sub-tasks in the same group reflects the design of the Find architecture.

%% file: 03_dataset.tex
\section{Datasets for Analysing Systematic Generalization in VQA}\label{sec:datasets_mnist_sqoop}

In this section, we introduce the VQA-MNIST and the SQOOP datasets to study the effect of libraries with different degrees of modularity on systematic generalization.

\subsection{VQA-MNIST: limited combinations of visual attributes}
We introduce VQA-MNIST, a family of four VQA datasets of which two of them are related to attribute extraction, while the other two deal with comparisons of attributes and spatial positions. All tasks require a binary answer (`yes' or `no'). 
Each object is an MNIST digit~\cite{lecun1998gradient} with multiple attributes: category, color, size and illumination condition among respectively ten, five, three, and three different options. This leads to a total amount of attribute instances equal to $21$, and a total amount of combinations of attribute instances equal to $450$. See Appendix~\ref{sec:attribute_relations_mnist} for more details.

\begin{figure}[!t]
    \centering
    \begin{tabular}{cccccc}
    \multicolumn{6}{c}{\hspace{-0.2cm}\includegraphics[width=1\linewidth]{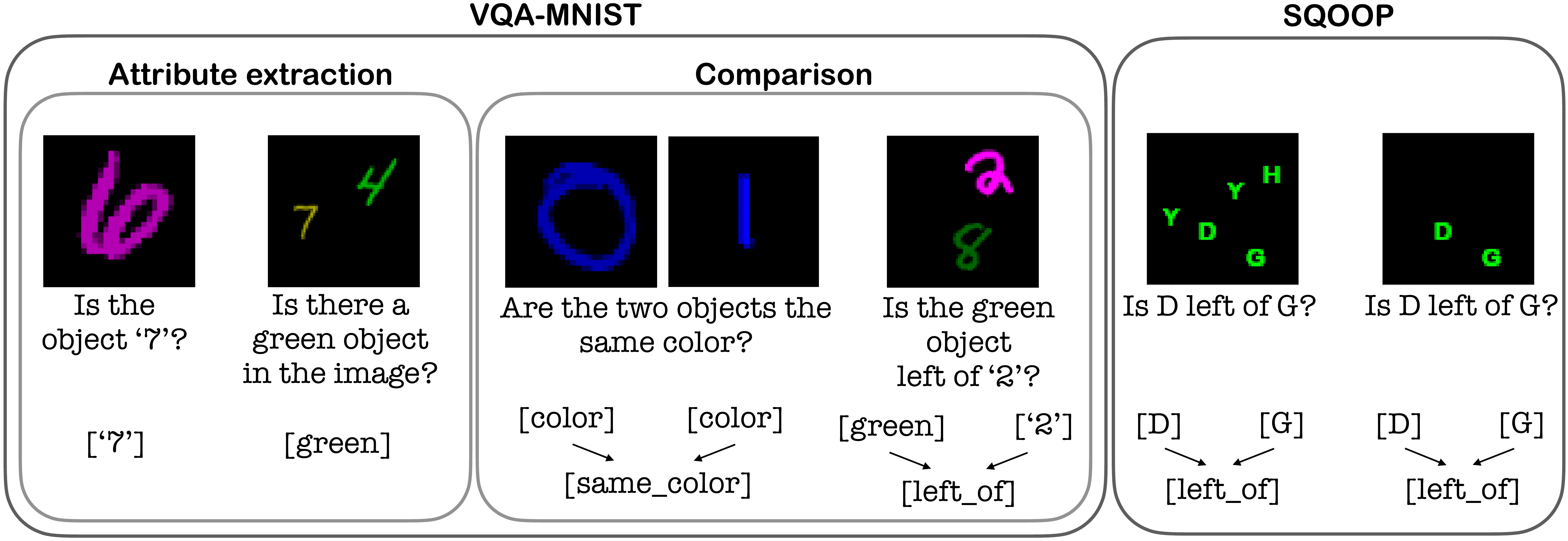}} \\
    \end{tabular} 
    \caption{\emph{VQA datasets to evaluate systematic generalization.} From top to bottom: Visual input, question and corresponding program layout. The VQA-MNIST contains objects that are combinations of four attributes (category, color, size and illumination condition). The tasks consist on attribute extraction  (from a single object and from multiple objects), and attribute comparisons (between separated object pairs and spatial locations). The training distribution has a limited amount of attribute combinations. The SQOOP dataset contains objects at different positions and the task consists on comparing the position of two given objects. The training distribution has a limited number of co-occurring objects in the image. The original SQOOP dataset contains five objects per image~\cite{bahdanau2018systematic}, and we further limit the training distribution by only allowing two objects per image. }
    \label{fig:synthetic_datasets}
\end{figure}

To generate a limited set of combinations of attributes, we fix a parameter $r$ and we randomly extract combinations among all the $450$ possibilities, with the requirement that each of the $21$ attribute instances appears at least $r$ times (see Appendix~\ref{sec:section_algo} for details). In order to study the effect of training data diversity, we compare NMNs trained with datasets with different number of attribute combinations and for a fair comparison, we evaluate them under the same testing conditions, as described next.  We define \emph{data-run} as six different training datasets originated from training combinations obtained with six different $r$'s ($[1,2,5,8,10,20]$), such that the set of training combinations extracted for smaller values of $r$ is included in the training combinations for larger values of $r$. We evaluate systematic generalization for all datasets in the data-run on the same images that come from novel combinations, 
obtained by fixing $r=5$. We generate the training and test examples as described in Appendix~\ref{sec:gen_train_test_examples}. The amount of training examples for each dataset in the data-run is fixed to $210$K for training, and $42$K for testing systematic generalization. In this way, all NMNs in the data-run are trained with the same amount of data,~\ie we evaluate different amounts of training data diversity for the same number of training images.

In Figure~\ref{fig:synthetic_datasets}, we depict the four VQA tasks divided into \emph{attribute extraction} and \emph{attribute comparison}, which are summarized in the following:  

\begin{itemize}[leftmargin=*]
    \item  \emph{Attribute extraction.} Starting from the left of Figure~\ref{fig:synthetic_datasets}, we show two datasets, 
\emph{attribute extraction from single object} and \emph{attribute extraction from multiple objects}. In the former case, the image 
contains a single object (image size is $28\times 28$ pixels), in the latter, it always contains a pair of objects (image size is $64\times 64$ pixels). The second object plays the role of a confounding factor. In both datasets, the questions inquire about one of the $21$ possible attribute instances. The ground-truth program layout corresponds to a single sub-task, which is identified by the attribute instance in the question. 
Thus, the program layout 
in the intermediate module corresponds to a single module identifying the sub-task (the sub-tasks in Figure~\ref{fig:synthetic_datasets} are identified by \emph{`7'} and \emph{green}). 

\item \emph{Attribute comparison.} At the center of Figure~\ref{fig:synthetic_datasets} we show two datasets for attribute comparison. These have a ground-truth program layout that is a tree structure. The first dataset consists on \emph{attribute comparison between pairs of separated objects}, in which comparisons are done between a pair of objects contained in separated images (of size $28\times 28$ pixels). The comparison are with respect to an attribute of the dataset. The two leaves in the program layout extract the attribute and the root tackles the comparison between them.
The second dataset consists on a \emph{comparison between spatial positions}, as shown in Figure~\ref{fig:synthetic_datasets} (image size $64\times 64$ pixels).  The leaves of the program layout identify the objects in the image, and the root tackles the spatial comparison (\ie above, below, left of, or right of).  

\end{itemize}

\subsection{SQOOP: limited amount of co-occurrence of objects in the images}
The SQOOP dataset tackles the problem of spatial relations between two objects~\cite{bahdanau2018systematic}. Namely, the questions consist on comparing the position of two objects, \emph{e.g., Is D left of G?}  The dataset contains four spatial relations (above, below, left of, or right of)  on  thirty-six possible objects.
All objects can appear in the image, but specific pairs of objects are not compared during training. By limiting the possible pairs of objects compared during training, systematic generalization is measured as the ability to compare novel pairs of objects. 

In the original SQOOP dataset~\cite{bahdanau2018systematic}, each image contains five objects, as in the left example of the SQOOP block in Figure~\ref{fig:synthetic_datasets}. We increase the bias of the SQOOP dataset, to make it more challenging for NMNs. To do so, we reduce the objects in the image to the pair in the question. We generate a dataset with the smallest amount of objects co-occurring with another object in an image, such that each object only co-occurs with another object. Given a generic training question \emph{e.g., ``Is D left of G?''}, all the training images related to this question contain the \emph{D} and \emph{G} objects only. This procedure substantially reduces the training data diversity of the original SQOOP datasets with five objects per image.

%% file: 04_results.tex
\section{Results}
\label{sec:results}


In this section, we report the systematic generalization accuracy of NMNs with different libraries of modules on the VQA-MNIST and the SQOOP datasets. 

\subsection{VQA-MNIST}

To analyze the impact of the degree of modularity of the library in systematic generalization, we report results using the following libraries, which are depicted in Figure~\ref{fig:ood_modules}a:
 
\begin{itemize}[leftmargin=*]
    \item \textit{all - all - all:}  
    This is a library with
    a single image encoder module, a single classifier module, and intermediate module defined as the Find module.  This library is commonly used in NMN~\cite{hu2017learning}, while the following ones are explored for the first time in this paper.
    
    \item \textit{all - group - group:} 
    This library contains a single image encoder, and intermediate modules and classifiers separated per group. We divide sub-tasks related to object categories, colors, spatial relations and so forth in different groups.

    \item \textit{group - all - all:}
    This is a library with multiple image encoders separated per group, where sub-tasks related to object categories, colors, spatial relations and so forth belong to different groups. It has a single intermediate module and a single classifier shared across all sub-tasks, respectively.
    
    \item  \textit{group - group - group:} This library has one image encoder module per each group of sub-tasks, as in the image encoder of \textit{group - all - all}. The intermediate modules and the classifier also tackle groups of sub-task, as in the intermediate modules and classifier of \textit{all - group - group}.

\item \textit{sub-task - sub-task - sub-task:} 
    This is the library
    with the maximum degree of modularity. The modules address a single sub-task at each stage of the NMN.
\end{itemize}

Other libraries of modules are analyzed in  Appendix~\ref{app:supplemental_results_vqa_mnist}, such as other state-of-the-art NMN implementations and different libraries that help to isolate the effect of the degree of modularity at the different stages. The analysis of the libraries in the Appendix serves to strengthen the evidence that support the conclusions obtained with the libraries analysed in this section.

\begin{figure*}[!t]
    \centering
    \begin{tabular}{@{\hspace{-0.1cm}}cc}
    \multicolumn{2}{c}{\includegraphics[width=0.95\linewidth]{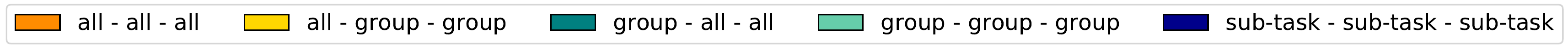} \hspace{1cm} } \\
    \hspace{-0.4cm} 
    \includegraphics[width=.5\linewidth]{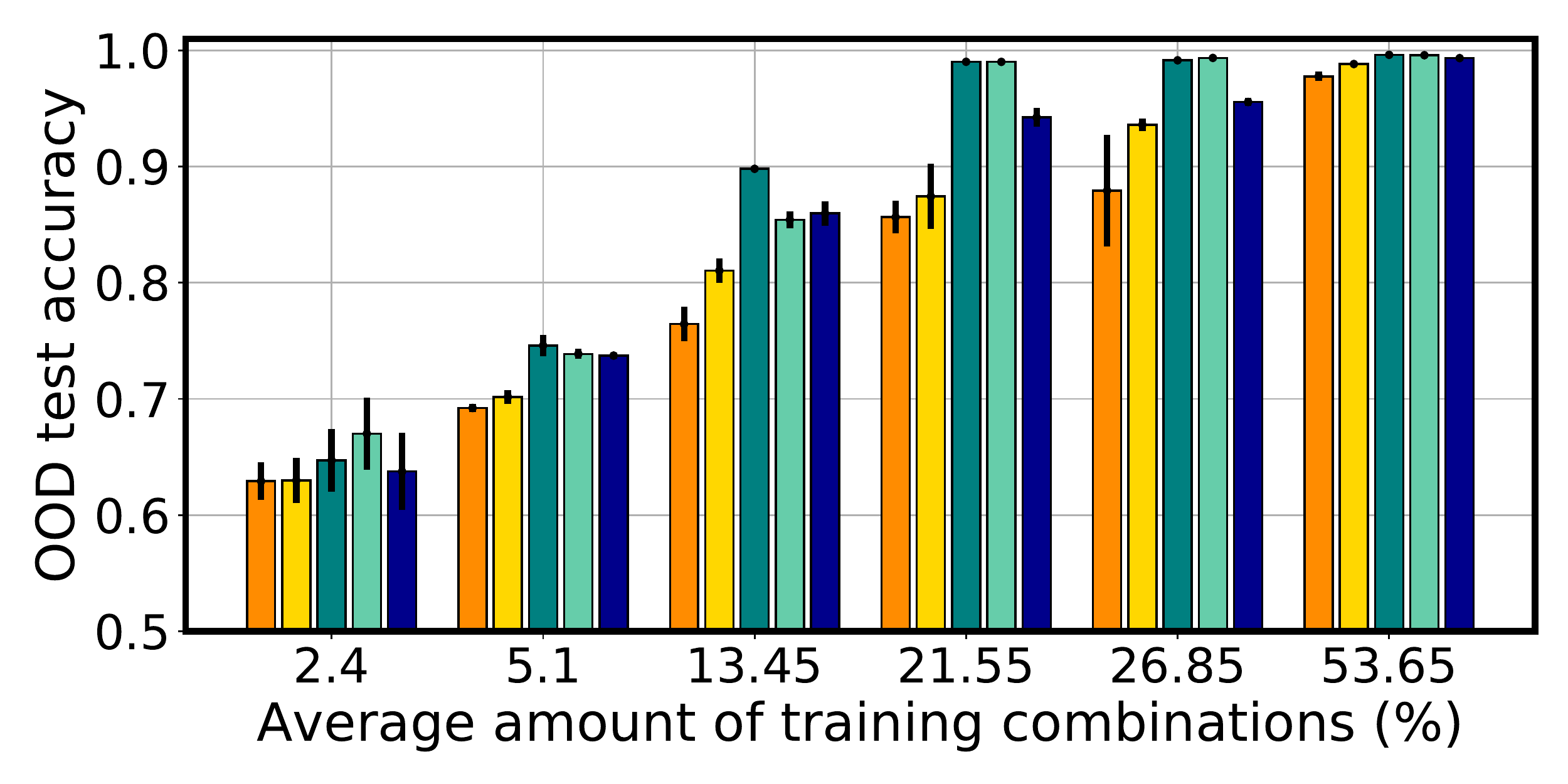} 
    & \hspace{-1.6cm} 
    \includegraphics[width=.5\linewidth]{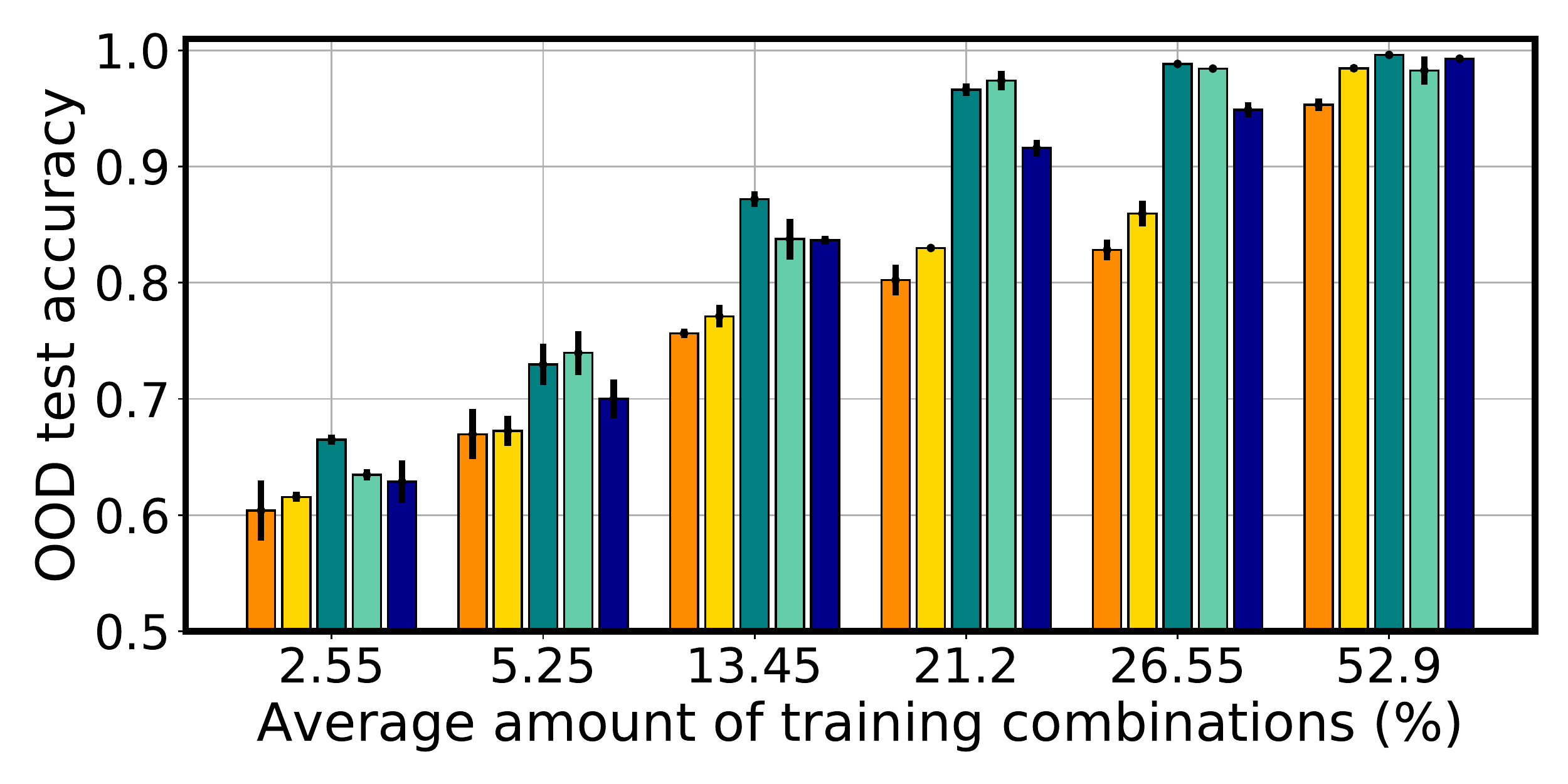} \\
    \vspace{0.5cm}
    (a) & (b) \\
    \hspace{-0.2cm}
    \includegraphics[width=.5\linewidth]{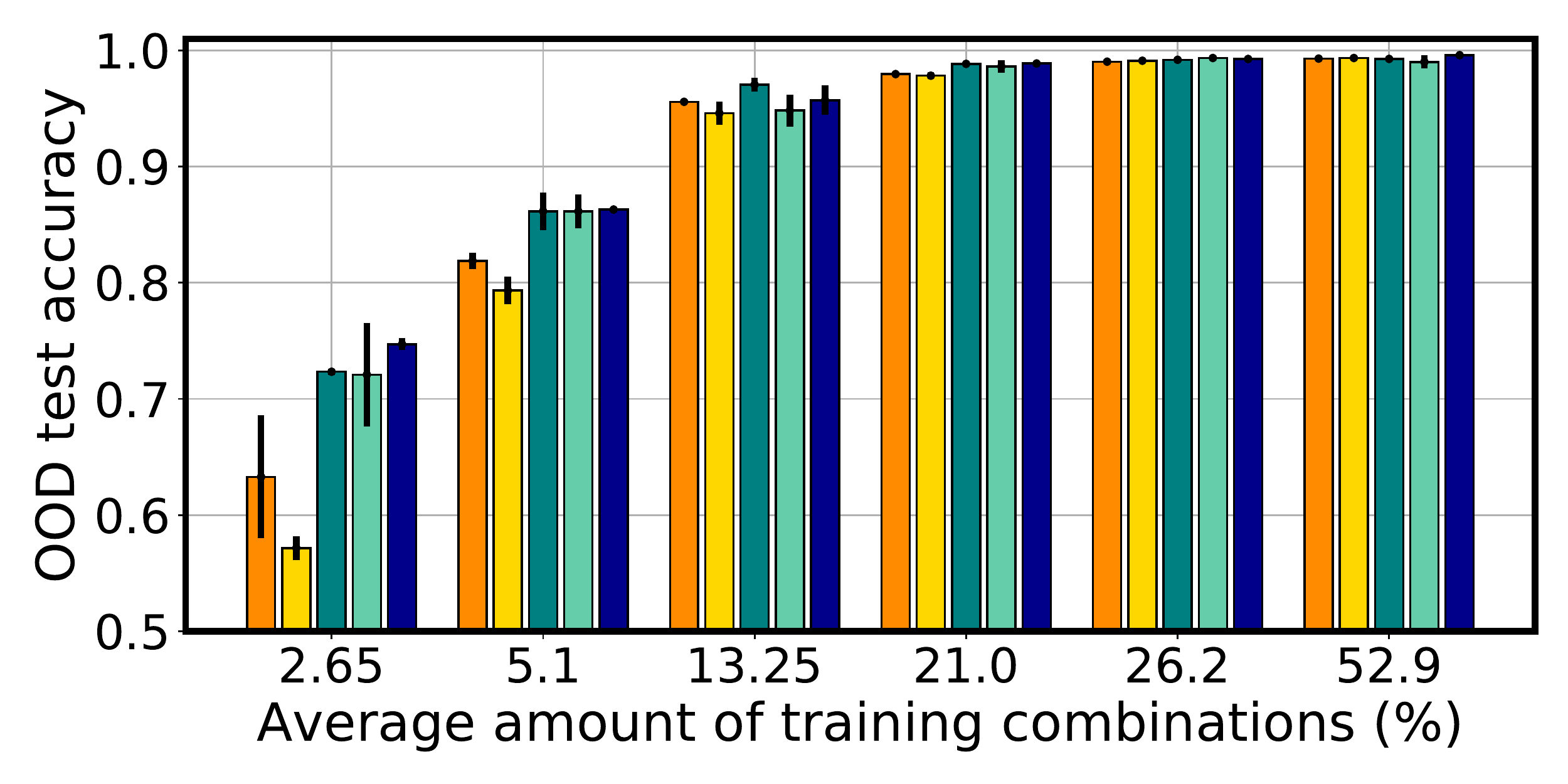}
    & \hspace{-1.6cm}
    \includegraphics[width=.5\linewidth]{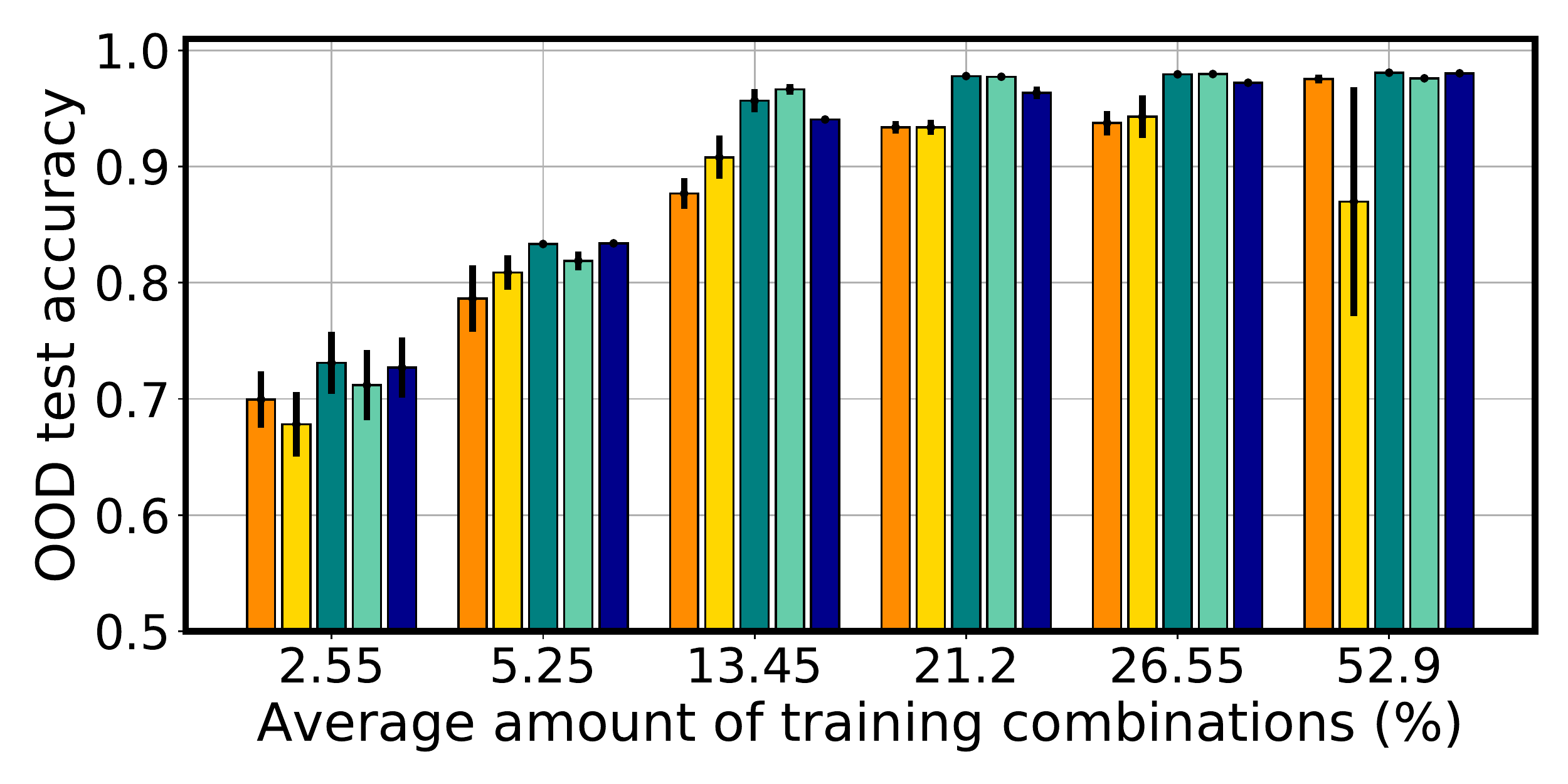} \\
    (c) & (d) \\
    \end{tabular}
    \caption{\emph{Results in VQA-MNIST.} Systematic generalization accuracy (referred as OOD test accuracy in the plot)  that evaluates VQA on novel combinations of visual attributes, for different amount of training combinations (\%),~\ie training data diversity. Results compare  different libraries of modules in the four types of tasks of VQA-MNIST: (a) attribute extraction from single object, (b) attribute extraction from multiple objects, (c) attribute comparison between pairs of separated objects, (d) comparison between spatial positions.}
    \label{fig:exps}
\end{figure*}

To measure systematic generalization accuracy, it needs to be  guaranteed that the novel combinations of attributes used for testing are not used in any stage of learning~\cite{teney2020value}. We perform a grid-search over the NMNs' hyper-parameters, and evaluate the accuracy using an in-distribution validation split at the last iteration of the training. The model with highest in-distribution validation accuracy is selected, such that the novel combinations of attributes are not used for hyper-parameter tuning (see Appendix~\ref{app:opt_details_multi_attribute} for more details).

Figure~\ref{fig:exps} shows the systematic generalization accuracy to novel combinations of attributes on the four tasks in VQA-MNIST. Across all plots, each value on the horizontal-axis displays the average percentage of training combinations for a fixed $r$ (recall that $r$ is the minimum amount of times an attribute instance appears in the training combinations) across two data-runs, where the bars are grouped based on the value of $r$. In the legend, we specify the library of modules used in the NMN. On the vertical-axis, we report mean and standard deviation of the systematic generalization accuracy across the two data-runs. 

\noindent \paragraph{Systematic generalization compared to in-distribution generalization is much more challenging.}  Even though NMNs achieve the highest  systematic generalization in the literature~\cite{bahdanau2018systematic, bahdanau2019closure},
 Figure~\ref{fig:exps} shows that the  systematic generalization accuracy  across VQA tasks can be very low when the amount of training combinations is small. This highlights the difficulty to generalize to novel combinations of attributes when the training datasets are biased. Recall that VQA-MNIST is a synthetic dataset which could be considered one of the simplest. Yet, state-of-the-art approaches in systematic generalization have a remarkable low accuracy in this dataset.  Appendix~\ref{sec:in-out} shows the in-distribution accuracy, which is in stark contrast with the systematic generalization accuracy, as it is close to $100\%$ accuracy in all evaluated cases. 
We underline that the low systematic generalization accuracy depends on the amount of training combinations,~\ie the training data diversity, which has been reported in other contexts beyond VQA, such as in object recognition \cite{purushwalkam2019task, madan2020capability} and in navigation tasks \cite{lake2018generalization, ruis2020benchmark}.

\noindent \paragraph{Libraries with an intermediate degree of modularity, especially in the image encoder stage, substantially improve systematic generalization.} 
The \emph{group - all - all} and the \emph{group - group - group} libraries achieve the highest systematic generalization accuracy across all VQA tasks. These libraries also clearly outperform the \emph{sub-task - sub-task - sub-task} library, which has the highest degree of modularity. The library with worst systematic generalization is \emph{all - all - all}, which has the lowest amount of modularity. Observe that the improvements of systematic generalization accuracy are large in some cases (there are gaps of about $15\%$ accuracy between libraries of modules). Also, note that the \emph{group - all - all} library is clearly superior to \emph{all - group - group}, which suggests that a higher degree of modularity is more helpful at the image encoder stage.   Taking these results together, we can conclude that a library with an intermediate degree of modularity (\emph{group}), specially at the image encoder stage, is crucially important to improve systematic generalization for novel combinations of visual attributes. 
In Appendix~\ref{app:supplemental_results_vqa_mnist}, we report several control experiments that support this conclusion. Namely, in Appendix~\ref{sec:results_soa} we validate the conclusions with other libraries of modules commonly used in the literature (such as \emph{all - sub-task - all}, and variants with batch normalization applied to the \emph{all} libraries, which is when it is only possible). In Appendix~\ref{sec:result_modular_image_encoder},~\ref{sec:results_subtasks} and ~\ref{sec:result_modular_classifier} we further validate the conclusion analysing variants for the image encoder,  intermediate modules and image classifier, respectively.

\noindent \paragraph{Higher systematic generalization is not reachable with larger amounts of training examples.}
Since using libraries with a larger number of modules may lead to modules being trained with lower number of training examples, we controlled that our conclusions are not affected by increasing the amount of training examples. We consider the task of attribute extraction from single object and attribute comparison between objects in separated images, and increased the training set size of a factor ten (for a total of $2.1$M training examples), while keeping the same percentage of training combinations of the previous experiments in Figure~\ref{fig:exps} (a,c). Results are reported in Appendix~\ref{app:more_datapoints}, which show that the systematic  generalization accuracy is not improved when increasing the number of training examples, and conclusions are consistent to those in Figure~\ref{fig:exps} (a,c). Thus, the amount of training examples in these experiments already  achieves the highest possible systematic generalization accuracy by increasing the dataset size,~\ie our conclusions are not a consequence of having libraries with data-starved modules.

\subsection{SQOOP}

We use the two libraries of modules used in previous work \cite{bahdanau2018systematic}: \emph{all - all - all} and \emph{all - sub-task - all}. To measure systematic generalization, we consider the model at its last training iteration using the hyper-parameters in previous works (batch normalization is used in the image encoder and classifier stages).

\begin{table}[t!]
  \caption{\emph{Results in the SQOOP dataset.} Systematic generalization accuracy (\%) for NMNs with tree program layout trained on SQOOP. Top row: SQOOP dataset with five objects per image, bottom row: SQOOP dataset with two objects per image. The \emph{all - sub-task - all} library reaches higher systematic generalization in the more difficult case of two objects per image.}
  \centering
  \begin{tabularx}{\textwidth}{XXX}
  \toprule
    & \emph{all - all - all}  & \emph{all - sub-task - all} \\
    \hline 
    Five objects per image \newline (as in previous work~\cite{bahdanau2018systematic})
    & $\mathbf{99.8 \pm 0.2}$ 
    & $\mathbf{99.96 \pm 0.06}$ \\
    \hline
    Two objects per image 
    & ${84\pm2}$   
    & $\mathbf{88.5\pm0.5}$ \\
    \hline
    \bottomrule 
    \end{tabularx}
    \label{table:sqoop_2objs}
\end{table}

In Table~\ref{table:sqoop_2objs}, we show the systematic generalization accuracy (mean and standard deviation across five trials).   Results are consistent with the conclusions obtained in VQA-MNIST.
On the original SQOOP dataset with five objects per image, both libraries achieve almost perfect systematic generalization over novel test questions. For the dataset with two objects in the image, the dataset is more biased because of the lower amount of co-occurring objects, and hence, it increases the difficulty of generalizing to novel question. In this case, the \emph{all - sub-task - all} library achieves significantly higher systematic generalization performance than the \emph{all - all - all} library. 
Finally, in Appendix~\ref{sec:further_sqoop} we report  the performance of other VQA models that do not use modules (\ie FiLM \cite{perez2018film} and MAC \cite{hudson2018compositional,bahdanau2018systematic}). The results provide re-assurance that NMNs achieve higher systematic generalization.

%% file: 05_clevr.tex
\section{Application on CLEVR-CoGenT Split for Systematic Generalization}\label{sec:clevr_systematic}

In this section, we apply the findings in the previous section on existing NMN architectures used in practice. To do so, we use CLEVR, a diagnostic dataset in VQA~\cite{johnson2017clevr}. This dataset consists of complex 3D scenes with multiple objects and ground-truth program layouts formed by compositions of sub-tasks. This dataset comes with additional splits to test systematic generalization, namely the Compositional Generalization Test (CLEVR-CoGenT). CLEVR-CoGenT is divided in two conditions where cubes and cylinders appear in a limited amount of colors, that are inverted between training and testing (see Appendix~\ref{sec:cogent_description} for details). In this way, we can measure systematic generalization to novel combinations of visual attributes (shape and color in this case).

We now show that the analysis in previous sections is helpful to improve state-of-the-art NMNs. We focus on Vector-NMN, which recently has been shown to outperform all previous NMNs~\cite{bahdanau2019closure}.   Vector-NMN uses a single image encoder shared across all sub-tasks.
The Vector-NMN's image encoder takes as input the features extracted from a pre-trained ResNet-101~\cite{he2016deep}, and consists of two trainable convolutional layers shared among all sub-tasks.   The details of Vector-NMN definition can be found in Appendix~\ref{app:vectornmn}. Our variant of Vector-NMN is based on tuning the degree of modularity in the encoder stage, as we have previously shown this can substantially improve the systematic generalization capabilities. Thus, we modify the library of Vector-NMN  by using a higher degree of modularity at the image encoder stage,~\ie one module for each group of sub-tasks.
Concretely, we define the groups as: (i) counting tasks, (ii) tasks related to colors, (iii) materials, (iv) shapes, (v) sizes, (vi) spatial relations, (vii) logical operations, (viii) input scene (for details, see Appendix~\ref{sec:subtask_division}).

\begin{table}[t!]
  \caption{\emph{Results in the CLEVR-CoGenT dataset.} Mean and standard deviation of test accuracy (\%) across five repetitions. NMNs are tested on in-distribution and out-of-distribution (systematic generalization).}
  \centering
  \begin{tabularx}{\textwidth}{XXXX}
  \toprule
    & Tensor-NMN \newline with \emph{all} \newline image encoder  & Vector-NMN \newline with \emph{all} \newline image encoder & Vector-NMN \newline with \emph{group} \newline image encoder ({\bf ours}) 
    \\
    \hline
    in-distribution
    & $97.9 \pm 0.1$
    & $\mathbf{98.0 \pm 0.2}$ 
    & $94.4 \pm 0.3$  \\
    \hline
    syst. generalization
    & $72.7 \pm 0.5$ 
    & $73.2 \pm 0.2$ 
    & $\mathbf{77.3 \pm 1.3}$ \\
    \hline
    \bottomrule
  \end{tabularx}
  \label{table:cogent}
\end{table}

The hyper-parameters are fixed to the ones in the previous work~\cite{bahdanau2019closure}, and the ground-truth program layout is given. 
Tables~\ref{table:cogent} and~\ref{tab:nmn_cogent} show the mean and standard deviation accuracy (across five runs with different network initialization each) of the original Vector-NMN, our version of Vector-NMN with a more modular (\emph{group}) image encoder, and also Tensor-NMN, which is another NMN commonly used in the literature that here serves as baseline.  Table~\ref{table:cogent} shows the  in-distribution accuracy and the systematic generalization accuracy in objects with novel combinations of shape and color. Results shows the wide gap between in-distribution  and out-of-distribution for all NMNs. Our Vector-NMN with  a more modular (\emph{group}) image encoder  achieves the best systematic generalization performance. This comes with a reduction of in-distribution generalization, which can be explained by the trade-off  between in-distribution and out-of-distribution generalization reported in previous works~\cite{madan2020capability,zaidi2020robustness}. 

Table~\ref{tab:nmn_cogent} shows the break-down of systematic generalization for each type of question. Note that the systematic generalization accuracy varies depending on questions types. The questions that are mostly affected by the limited amount of object shapes and color combinations in CLEVR-CoGenT,~\ie \texttt{query\_shape} and \texttt{query\_color}, is where a higher degree of modularity in the image encoder brings an improvement of the accuracy of $12\%$ and $7\%$, respectively. A broader comparison of our  approach with other non-modular approaches such as FiLM~\cite{perez2018film} and MAC~\cite{hudson2018compositional,bahdanau2019closure}, and other NMN variants is reported in Appendix~\ref{sec:app_cogent_more_models}. These results further demonstrate the higher systematic generalization accuracy of our approach. 

\begin{table}
\centering
    \caption{\emph{Breakdown of the results in the CLEVR-CoGenT dataset.} Breakdown by question type. Systematic generalization accuracy (\%) is reported, and it is the average across five trials.}
\begin{tabularx}{\textwidth}{XXXX}
\toprule
& Tensor-NMN \newline with \emph{all} \newline image encoder  & Vector-NMN \newline with \emph{all} \newline image encoder & Vector-NMN \newline with \emph{group} \newline image encoder ({\bf ours}) \\
\midrule
    \texttt{count  }
    & $69.7 \pm 0.8$ 
    & $70.4 \pm 0.4$ 
    & $\mathbf{71\pm 1}$ \\
    \hline
   \texttt{equal\_color }   
    &   $75.6 \pm 0.8$ 
    &   $74 \pm 1$ 
    &  $\mathbf{80 \pm 1}$ \\
    \hline
    \texttt{equal\_integer }
    &   $82.7 \pm 0.3$ 
    &   $78 \pm 2$ 
    &   $\mathbf{85 \pm 2}$ \\
    \hline
    \texttt{equal\_material }
    &   $74 \pm 2$ 
    &   $74.2 \pm 0.7$ 
    &   $\mathbf{84 \pm 2}$ \\
    \hline
    \texttt{equal\_shape   } 
    &  $\mathbf{91 \pm 2}$ 
    &  $89 \pm 3$ 
    &  $79 \pm 2$ \\
    \hline
    \texttt{equal\_size } 
    &   $75 \pm 1$ 
    &   $75 \pm 1$ 
    &   $\mathbf{88 \pm 2}$ \\
    \hline
    \texttt{exist    }
    &  $84.2 \pm 0.4$ 
    &  $\mathbf{84.4 \pm 0.4}$ 
    &  $84.4 \pm 0.5$ \\
    \hline
    \texttt{greater\_than  } 
    & $83.8 \pm 0.6$ 
    & $83.6 \pm 0.4$ 
    & $\mathbf{89 \pm 1}$ \\
    \hline
    \texttt{less\_than }     
    & $80.7 \pm 0.9$ 
    & $82.0 \pm 0.5$ 
    & $\mathbf{87 \pm 2}$ \\
    \hline
    \texttt{query\_color }
    &   $58 \pm 1$ 
    &   $60 \pm 1$ 
    &   $\mathbf{67 \pm 4}$ \\
    \hline
    \texttt{query\_material }
    &  $84.1 \pm 0.9$ 
    &  $84.7 \pm 0.4$ 
    &  $\mathbf{88.2 \pm 0.8}$ \\
    \hline
    \texttt{query\_shape }
    &  $37 \pm 1$ 
    &   $40 \pm 3$ 
    &  $\mathbf{52 \pm 3}$ \\
    \hline
    \texttt{query\_size    } 
    &    $83.5 \pm 0.6$ 
    &   $84.7 \pm 0.7$ 
    &  $\mathbf{89.5 \pm 0.5}$ \\
    \hline
\bottomrule
\end{tabularx}
\label{tab:nmn_cogent}
\end{table}

%% file: 06_conclusions.tex
\section{Conclusions and Discussion}\label{sec:conclusions}

Our results demonstrate that NMNs with a library of modules with an  intermediate degree of modularity, especially at the image encoder stage, substantially improves systematic generalization. This finding is consistent across datasets of different complexity and NMN architectures, and it is easily applicable to improve state-of-the-art NMNs by using modular image encoders.

This work also has led to new research questions. While we have shown that modularity has a large impact in systematic generalization, we have tuned the degree of modularity at hand and in practical applications it is desirable to do this automatically. 
Also, it is unclear how other types of bias, such as bias in the program layout, could affect systematic generalization. To motivate follow-up work in other types of bias beyond the ones analyzed in this work, in Appendix~\ref{app:closure_results} we report a pilot experiment that shows that bias in the program layout could degrade systematic generalization of the NMNs we introduced. This suggests that there may be a trade-off between the degree of modularity and bias in the program layout, which will be investigated in future works.

Finally, we are intrigued about the neural mechanisms that facilitate systematic generalization and how the degree of modularity  affects those mechanisms. Some hints towards an answer have been pointed out by~\cite{madan2020capability}, which shows that the emergence of invariance to nuisance factors at the individual neuron level improves systematic generalization for object recognition.

%% file: A1_appendix_data_generation.tex
\section{VQA-MNIST Dataset}
\label{app:data_generation}

In Appendix~\ref{sec:attribute_relations_mnist}, we describe the attributes and the comparisons in the VQA-MNIST datasets. The algorithm for the generation of the training and testing combinations is in Appendix~\ref{sec:section_algo}. Given the
combinations, in Appendix~\ref{sec:gen_train_test_examples} we define the procedure to generate training and testing examples.

\subsection{Attributes and Comparisons}
\label{sec:attribute_relations_mnist}

\begin{figure}[!ht]
    \centering
    \includegraphics[width=0.5\textwidth]{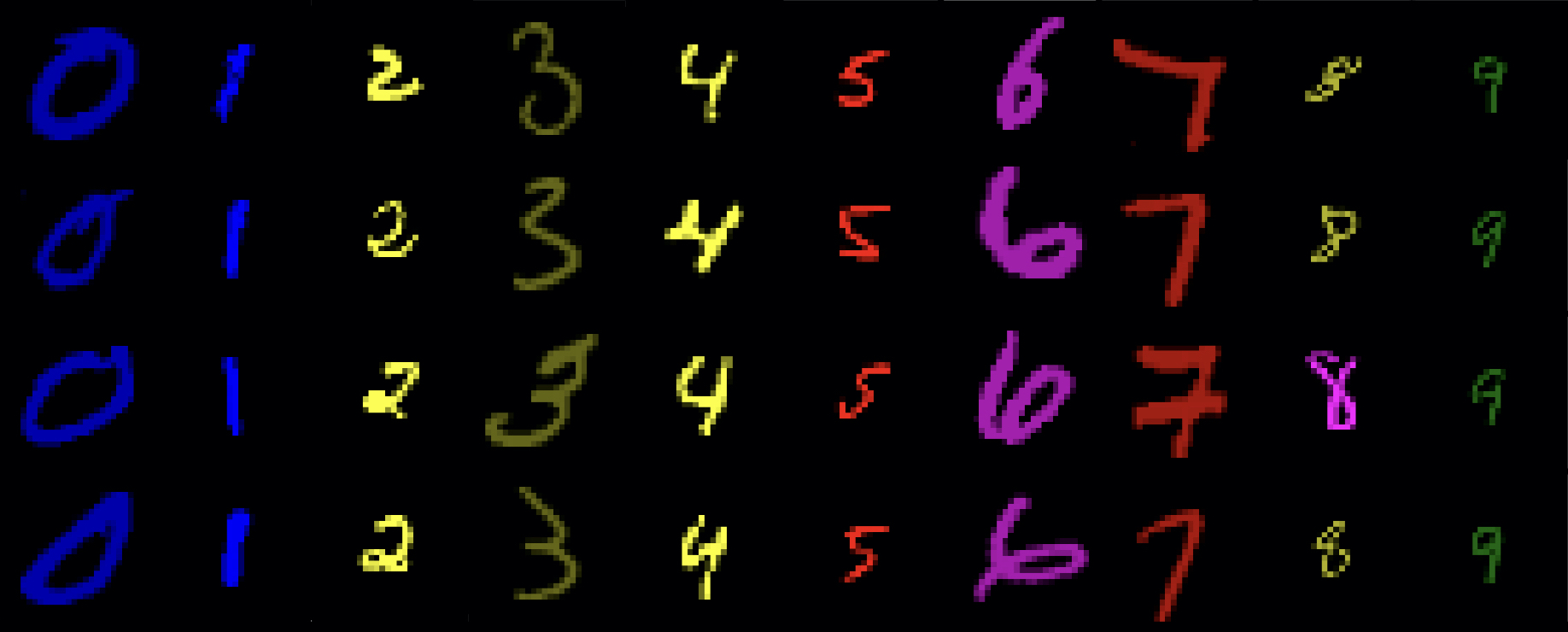}
    \caption{VQA-MNIST: some examples of training images, with a limited amount of attributes combinations. }
    \label{fig:example_multiattribute_mnist}
\end{figure}

\paragraph{Visual attributes.} Each object in VQA-MNIST is characterized by a category, the original MNIST category from 0 to 9, a color among red, green, blue, yellow and pink, a light condition among bright, half-bright, and dark, correspondent to a multiplicative factor $(1, \ 0.7, \ 0.4)$ applied on the whole image, and a size among large, medium, small, correspondent to three scale factors $(1, \ 5/7, \ 1/2)$ of the original digit. This corresponds to a total of 21 attribute instances.  Figure~\ref{fig:example_multiattribute_mnist} shows some objects with a reduced amount of combination of attributes.

\paragraph{Attribute comparisons.} The sub-tasks used for attribute comparison between two separated objects are in Table \ref{tab:division_comparisons}. The sub-tasks for comparison of spatial relations are in Table~\ref{tab:division_spatial}.

\begin{table}[!h]
    \centering
    \caption{Sub-tasks for attribute comparison between pairs of separated objects. The sub-tasks are divided in groups.}
    \begin{tabular}{c|l}
    \toprule
        \emph{group of sub-tasks} & \emph{sub-tasks} \\ 
        \hline
         category & \texttt{category} \\
         \hline
         color & \texttt{color} \\
         \hline
         light & \texttt{light} \\
         \hline
         size & \texttt{size} \\
         \hline
         comparison of category &  \texttt{same\_category};  \texttt{different\_category}\\
         \hline
         comparison of color &   \texttt{same\_color}; \texttt{different\_color} \\
         \hline
         comparison of light condition & \texttt{same\_light}; \texttt{different\_light}; \texttt{brighter}; \texttt{darker} \\
         \hline
         comparison of size & \texttt{same\_size}; \texttt{different\_size}; \texttt{larger}; \texttt{smaller} \\
         \hline
         \bottomrule
    \end{tabular}
      \label{tab:division_comparisons}
\end{table}

\begin{table}[!h]
    \centering
    \caption{Sub-tasks for comparison of spatial positions between two objects. The sub-tasks are divided in groups.}
    \label{tab:division_spatial}
    \begin{tabular}{c|l}
    \toprule
    \emph{group of sub-tasks} & \emph{sub-tasks} \\ 
    \hline
     category & \texttt{0}; \texttt{1}; \texttt{2}; \texttt{3}; \texttt{4}; \texttt{5}; \texttt{6}; \texttt{7}; \texttt{8}; \texttt{9} \\
    \hline
    color & \texttt{red}; \texttt{green}; \texttt{blue}; \texttt{yellow}; \texttt{pink}  \\
    \hline
    light & \texttt{bright}; \texttt{half-bright}; \texttt{dark}  \\
    \hline
    size & \texttt{large}; \texttt{medium}; \texttt{small}  \\
    \hline
    spatial relations 
    & \texttt{left\_of}; \texttt{right\_of}; \texttt{above}; \texttt{below} \\
    \hline
    \bottomrule
    \end{tabular}
\end{table}

\subsection{Algorithm for the Generation of Training and Test Combinations}\label{sec:section_algo}

To generate the training and test combinations across all the VQA-MNIST datasets, we fix the set of attribute types: category (attr1), color (attr2), light condition (attr3), and size (attr4); and we define a list of attribute instances for each type. The list $\mathrm{attr1}=[\texttt{0,\dots, 9}]$ contains the ten categories, the list
$\mathrm{attr2} = [\mathrm{\texttt{red, green, blue, yellow, pink}}]$ 
the five colors, the list $\mathrm{attr3} = [\mathrm{\texttt{bright, half-bright, dark}}]$ 
the three light conditions, and the list $\mathrm{attr4} = [\mathrm{\texttt{small, medium, large}}]$ 
the three sizes. We define a single comprehensive list of all attributes as $\mathrm{attributes} = \mathrm{attr1} + \mathrm{attr2} + \mathrm{attr3} + \mathrm{attr4} $, of length 21.
Two integers are required for the generation of training and test combinations: $\mathrm{n\_combinations\_train}$ and $\mathrm{n\_combinations\_test}$, which define the minimum amount of times (indicated with the $r$ parameter) each attribute will appear in the training and testing sets, respectively. We also define $\mathrm{train\_combinations}$ and $\mathrm{test\_combinations}$ as two empty lists. Through a stochastic procedure described in Algorithm~\ref{algo:comb_generation}, we fill these lists with the objects for the training and the testing sets.

\begin{algorithm}
\begin{algorithmic}[1]
\Require $\mathrm{n\_train\_combinations},\ \mathrm{n\_test\_combinations}$
\State $\mathrm{train\_combinations,\ test\_combinations} = [\ ], [\ ]$, $k=0$, $\mathrm{random\_seed}=n$

\For{$k<\mathrm{n\_test\_combinations}$} 

\\
  \State{$\mathrm{taken} = \mathrm{dict}() $}
   
   \For{$a\_$ in $\mathrm{attributes}$} 
   \State{ $\mathrm{taken}[a\_] = False$} 
    \EndFor
    
        \Comment{At every iteration in the loop we set the dictionary of seen attribute to False}    

\\
    \State $\mathrm{missing\_attributes} = True$
    
    \While{$\mathrm{missing\_attributes}$} 
    
        \Comment{Until we see all the attributes at the $k$-th iteration} 
        
        \State $\mathrm{candidate}= [\mathrm{choice}(\mathrm{attr1}), \mathrm{choice}(\mathrm{attr2}), \mathrm{choice}(\mathrm{attr3}), \mathrm{choice}(\mathrm{attr4})]$
        
        \Comment{random extraction using uniform distribution}
        
        \State $\mathrm{exists = candidate \ in  \ test\_combinations}$

        \State $\mathrm{reject} = True$ 
        \Comment{We assume that all attributes in candidate already appeared}
            \For{$c\_$ in $\mathrm{candidate}$} 
                \State{$\mathrm{reject} \ *= \mathrm{taken}[c\_]$}
            \EndFor
            
        \If {$\mathrm{not \ reject \  \& not \ exists}$}
            \State $\mathrm{test\_combinations} \gets \mathrm{candidate}$
            \For{$c\_$ in $\mathrm{candidate}$} 
                \State{$\mathrm{taken}[c\_] = True$}
            \EndFor
        \EndIf
        
        \If {all values of $\mathrm{taken}$ are \emph{True}}
             \State$\mathrm{missing\_attributes} \gets False$
             \State$k \gets k+1$
            \Comment{All attributes appeared once, next iteration}
        \EndIf
    \EndWhile
\EndFor
\State To fill the list $\mathrm{train\_combinations}$, substitute in line 2 $\mathrm{n\_test\_combinations}$ to $\mathrm{n\_train\_combinations}$. Change declaration of line 12 into line 30. Repeat from 2-28. 
\State $\mathrm{exists}$ = ($\mathrm{candidate}$ in $\mathrm{test\_combinations}$) OR ($\mathrm{candidate}$ in $\mathrm{train\_combinations}$)
\end{algorithmic}
\caption{Generation of Training and Test Combinations}\label{algo:comb_generation}
\end{algorithm}

The additional variable $\mathrm{reject}$ and the loop in 14-16 is to exclude a new tuple if it contains a list of attributes which have already appeared, \emph{e.g.,} if $\mathrm{test\_combinations}=[(3, \mathrm{yellow, bright, small}),\ (4, \mathrm{blue, dark, large})]$, the candidate $(3,\mathrm{ blue, dark, small})$ will be rejected. This constraint allows to guarantee that all attributes appear at least $r$ times while minimizing the number of combinations used to do so.

\subsection{Procedure for the Generation of Training and Test Examples}
\label{sec:gen_train_test_examples}

Across all the VQA-MNIST datasets, we leverage on Algorithm~\ref{algo:comb_generation} to generate the combinations of attributes. The digits of MNIST used for the training, validation, and test splits have been fixed a priori, with $50$K training digits, $10$K validation digits, and the $10$K test digits of the original MNIST. Depending on the VQA task, we differentiate the data generation process.

\paragraph{Attribute Extraction.}
Given a split (training or test), and the VQA question with its corresponding sub-task, we divide the objects in the split into those providing a positive and negative example. We randomly select a digit from the MNIST split which belongs to the category of the example, and we introduce the additional attributes to the image.
We generate evenly examples for both cases, so to have a balanced dataset.
The program layout corresponds to one of the 21 attribute instances.

\paragraph{Attribute Comparison.} The data generation repeats identically at training and testing. 
Given a VQA question about a relation and a split, we first collect all the pairs of objects from that split that provide a positive answer to the question. We repeat the same for the negative pairs. Then, we generate the examples based on the attributes for the pair of objects, by extracting images from the MNIST split which match with the categories of the objects. We generate an even number of positive and negative examples for each question, to have a balanced dataset.

For the tasks of attribute comparison, the program layout has a tree structure. 
For attribute comparison between pairs of separated objects, the sub-tasks at the leaves can be category, color, light condition, and size. These sub-tasks match with the relational questions (\eg if we are asking if two objects are the same color, the leaves tackle the sub-task color). The sub-tasks and their division in groups are detailed in Table~\ref{tab:division_comparisons}.
For comparison between spatial positions, the sub-tasks at the leaves can assume the form of one of the 21 attribute instances, while the root coincides with one of the four 2D spatial relations among below, above, left, and right. The sub-tasks and their division in groups are detailed in Table~\ref{tab:division_spatial}.

%% file: A2_more_networks.tex
\newpage

\section{Implementation Details and Supplemental Results on VQA-MNIST}
\label{app:supplemental_results_vqa_mnist}

In Section~\ref{sec:definition_img_encoder_class}, we give the definition of image encoder and classifier architectures, common to all the NMNs trained on VQA-MNIST.
We present the experimental setup and optimization details for the experiments on VQA-MNIST in Section \ref{app:opt_details_multi_attribute}.
The in-distribution generalization of the main libraries on VQA-MNIST are in Section~\ref{sec:in-out}.

Then, we introduce a series of control libraries to prove the generality of our findings. In particular, in Section~\ref{sec:results_soa} we compare libraries with modular image encoder to state-of-the-art libraries in the NMN literature. In Section~\ref{sec:result_modular_image_encoder} we analyze the effect of different degrees of modularity at the image encoder stage. In Section~\ref{sec:results_subtasks}, we further verify the impact of different architectures for the intermediate modules on systematic generalization. In Section~\ref{sec:result_modular_classifier}, we analyze the effect of different degrees of modularity at the classifier stage. Finally, in Section~\ref{app:more_datapoints}, we report the systematic generalization of NMNs trained ten times larger datasets than those in Figure~\ref{fig:exps}.

\subsection{Description of the Image Encoder and Classifier Architectures}
\label{sec:definition_img_encoder_class}

\begin{table}[!ht]
    \centering
        \caption{Image encoder module VQA-MNIST. The first two parameters in Conv2d refer to the number of input and output channels. For those libraries without batch normalization, the BatchNorm2d and BatchNorm1d should not be considered.}
    \begin{tabular}{c|c|c}
    \toprule
    \multicolumn{3}{c}{{Image Encoder}} \\
    \toprule
    & Layer & Parameters \\
    \hline
    (0) & Conv2d     & (3, 64, kernel\_size=(3, 3), stride=(1, 1), padding=(1, 1), bias=False)  \\
    \hline
    (1) & BatchNorm2d & (64, eps=1e-05, momentum=0.1, affine=True, track\_running\_stats=True) \\
    \hline
    (2) & ReLU &  \\
    \hline
    (3) & Conv2d & (64, 64, kernel\_size=(3, 3), stride=(1, 1), padding=(1, 1), bias=False) \\
    \hline
    (4) & BatchNorm2d & (64, eps=1e-05, momentum=0.1, affine=True, track\_running\_stats=True) \\
    \hline
    (5) & ReLU & \\
    \hline
    (6) & MaxPool2d & (kernel\_size=2, stride=2, padding=0, dilation=1, ceil\_mode=False) \\
        \hline
    (7) & Conv2d & (64, 64, kernel\_size=(3, 3), stride=(1, 1), padding=(1, 1), bias=False) \\
        \hline
    (8) & BatchNorm2d & (64, eps=1e-05, momentum=0.1, affine=True, track\_running\_stats=True) \\   
    \hline
    (9) & ReLU & \\
        \hline
    (10) & Conv2d &(64, 64, kernel\_size=(3, 3), stride=(1, 1), padding=(1, 1), bias=False) \\
        \hline
    (11) & BatchNorm2d & (64, eps=1e-05, momentum=0.1, affine=True, track\_running\_stats=True) \\
        \hline
    (12) & ReLU & \\ 
        \hline
    (13) & MaxPool2d & (kernel\_size=2, stride=2, padding=0, dilation=1, ceil\_mode=False) \\
        \hline
    (14) & Conv2d & (64, 64, kernel\_size=(3, 3), stride=(1, 1), padding=(1, 1), bias=False) \\
        \hline
    (15) & BatchNorm2d & (64, eps=1e-05, momentum=0.1, affine=True, track\_running\_stats=True) \\
        \hline
    (16) & ReLU & \\
        \hline
    (17) & Conv2d &  (64, 64, kernel\_size=(3, 3), stride=(1, 1), padding=(1, 1), bias=False)\\
        \hline
    (18) & BatchNorm2d & (64, eps=1e-05, momentum=0.1, affine=True, track\_running\_stats=True)\\
        \hline
    (19) & ReLU & \\
    \hline
    \bottomrule
    \end{tabular}
    \label{tab:image_encoder_hyper_params_nmn_multi_attribute_mnist}
\end{table}

\begin{table}[!ht]
    \centering
        \caption{Classifier module for libraries VQA-MNIST. The first two parameters in Conv2d refer to the number of input and output channels. For those libraries without batch normalization, the BatchNorm2d and BatchNorm1d should not be considered.}
    \begin{tabular}{c|c|c}
    \toprule
    \multicolumn{3}{c}{{Binary Classifier}} \\
    \toprule
    & Layer & Parameters \\
    \hline
    (0) & Conv2d & (64, 512, kernel\_size=(1, 1), stride=(1, 1)) \\
    \hline
    (1) & BatchNorm2d & (512, eps=1e-05, momentum=0.1, affine=True, track\_running\_stats=True) \\
    \hline
    (2) & ReLU &  \\
    \hline
    (3) & MaxPool2d & (kernel\_size=7, stride=7, padding=0, dilation=1, ceil\_mode=False) \\
    \hline
    (4) & Flatten &  \\
    \hline
    (5) & Linear & (in\_features=512, out\_features=1024, bias=True) \\
    \hline
    (6) & BatchNorm1d & (1024, eps=1e-05, momentum=0.1, affine=True, track\_running\_stats=True) \\
    \hline
    (7) & ReLU & \\ 
    \hline
    (8) & Linear & (in\_features=1024, out\_features=2, bias=True) \\
    \hline
    \bottomrule
    \end{tabular}
    \label{tab:classifier_hyper_params_nmn_multi_attribute_mnist}
\end{table}

The architectures of the image encoder and the classifier modules are shared across all the experiments on VQA-MNIST. All the libraries in Section~\ref{sec:results}, as well as most of those in this section, do not have batch normalization, except for \emph{all(bn) - all - all(bn)} and \emph{all(bn) - sub-task - all(bn)} libraries. We report the implementation of an image encoder module and a classifier module, respectively in Table~\ref{tab:image_encoder_hyper_params_nmn_multi_attribute_mnist} and Table~\ref{tab:classifier_hyper_params_nmn_multi_attribute_mnist}, where we include the batch normalization. 

For those image encoders and classifiers without batch normalization, the BatchNorm2d and BatchNorm1d layers need to be excluded (respectively transformations 1, 4, 8, 11, 15, and 18 in the image encoder and transformations 1 and 6 in the classifier). Notice that in Tables~\ref{tab:image_encoder_hyper_params_nmn_multi_attribute_mnist} and \ref{tab:classifier_hyper_params_nmn_multi_attribute_mnist}, the first two parameters for the Conv2d define the number of input and output channels.

\subsection{Optimization Details}
\label{app:opt_details_multi_attribute}

Across all the experiments on VQA-MNIST on $210K$ training examples, we fix the number of training steps to $200K$. For the task of attribute extraction from a single object, the grid of hyper-parameters includes batch size with candidate values $[128,\ 256]$, and the learning rates with values $[10^{-4},\ 10^{-3},\ 0.005,\ 0.01]$. 
Given the high computational cost of training, for all the remaining experiments we reduce the number of models by fixing the batch size to $64$ and the learning rate hyper-parameters to the values $[10^{-5}, 10^{-4}, 10^{-3}, 0.005, 0.01]$. 

During training, the learning rate is kept constant. We notice that in some cases, especially on comparison tasks, the training loss function, as well as the training accuracy, often presents several plateaus alternated to jumps, despite the absence of a decaying learning rate. This did not allow us to use early stopping as a strategy to accelerate the experiments.

\subsection{In-distribution generalization}
\label{sec:in-out}

\begin{figure}[!ht]
    \centering
    \begin{tabular}{@{\hspace{-0.1cm}}cc}
    \multicolumn{2}{c}{\includegraphics[width=0.95\textwidth]{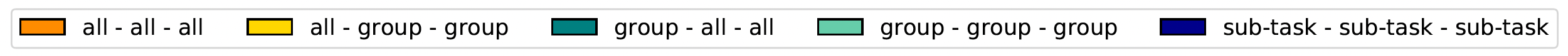}}  \\
    \includegraphics[width=0.5\textwidth]{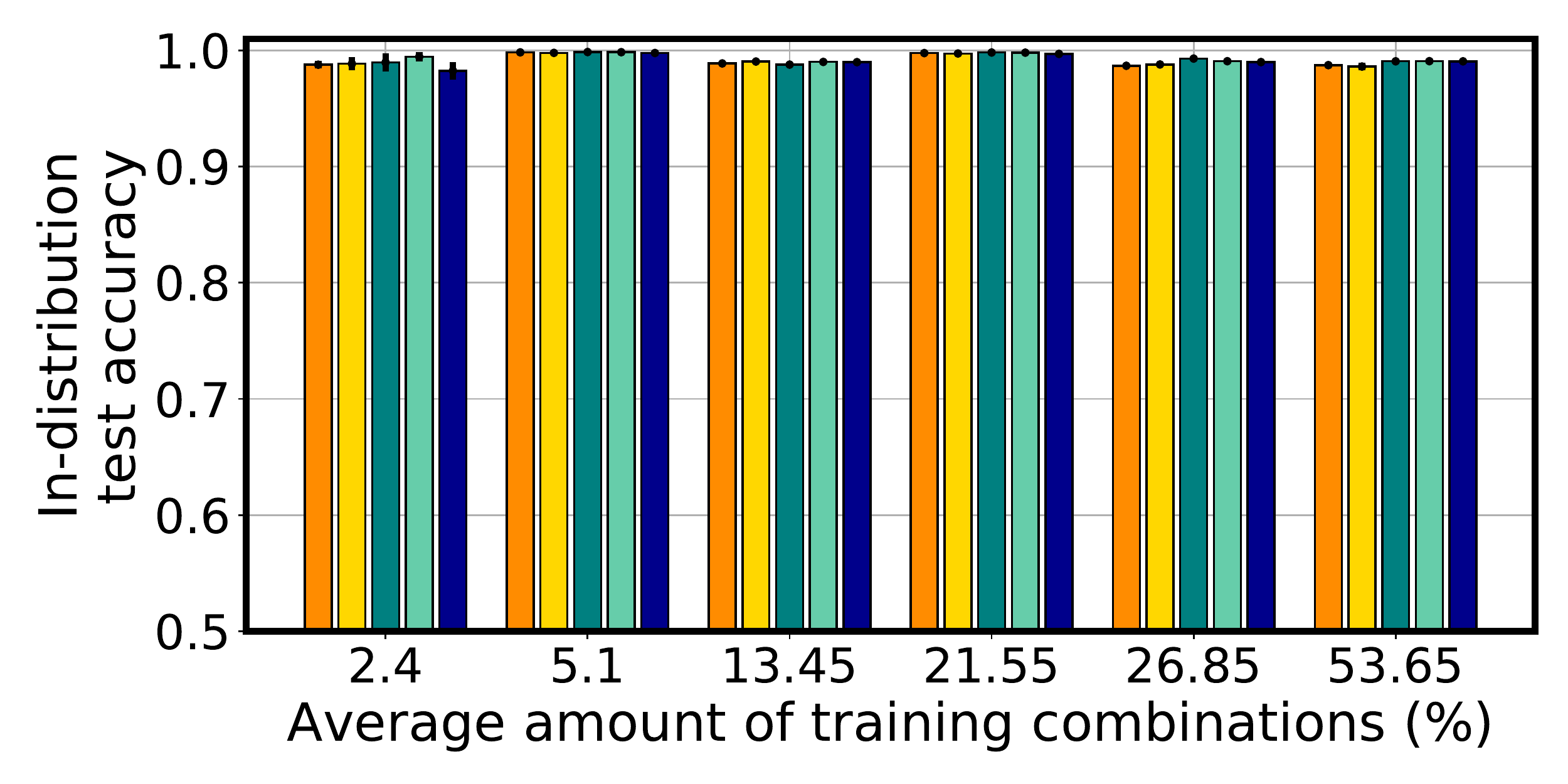} 
    & \hspace{-0.55cm}
    \includegraphics[width=0.5\textwidth]{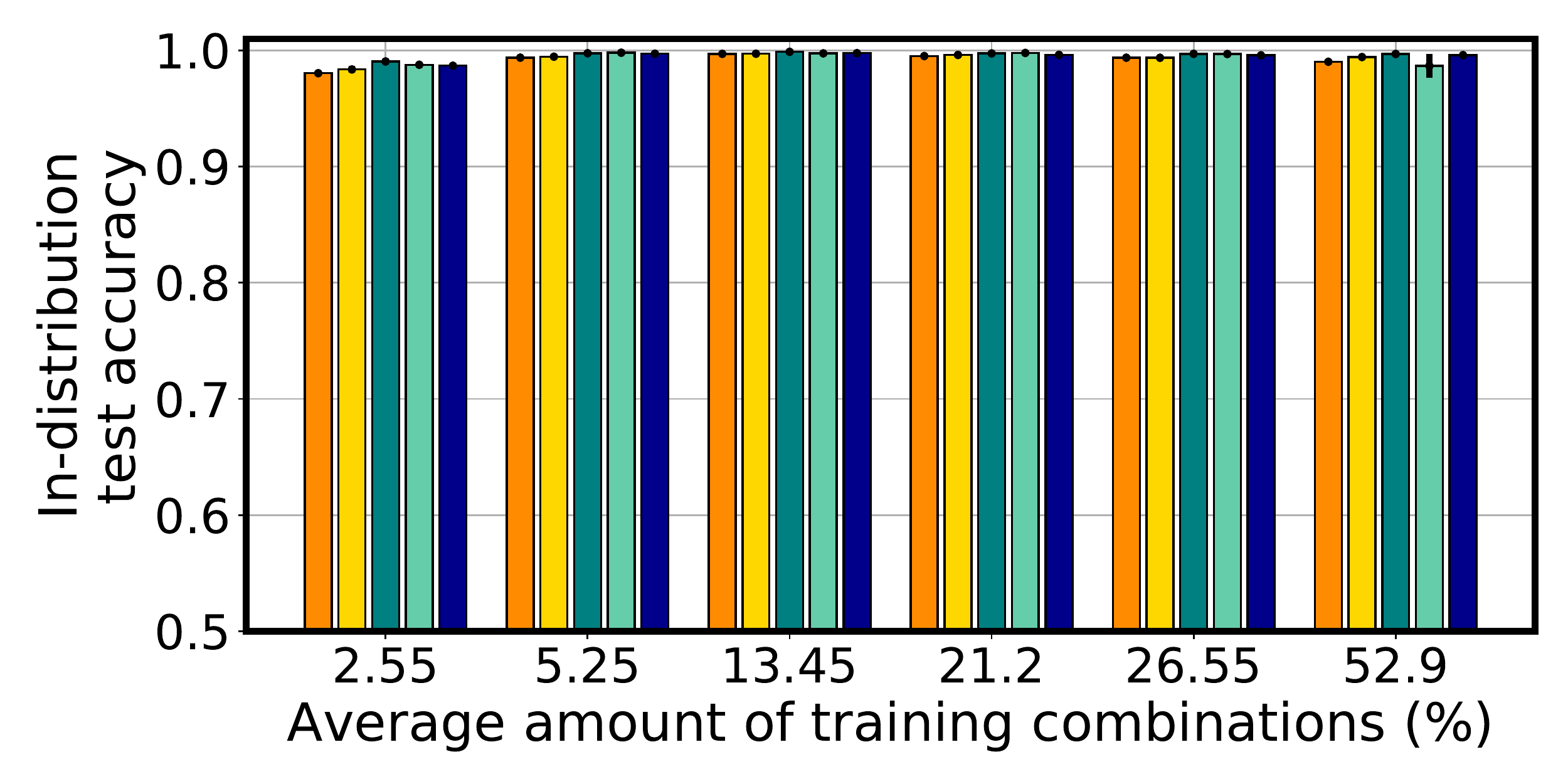}   \\
    \vspace{0.4cm}
    (a) & (b) \\
    \includegraphics[width=0.5\textwidth]{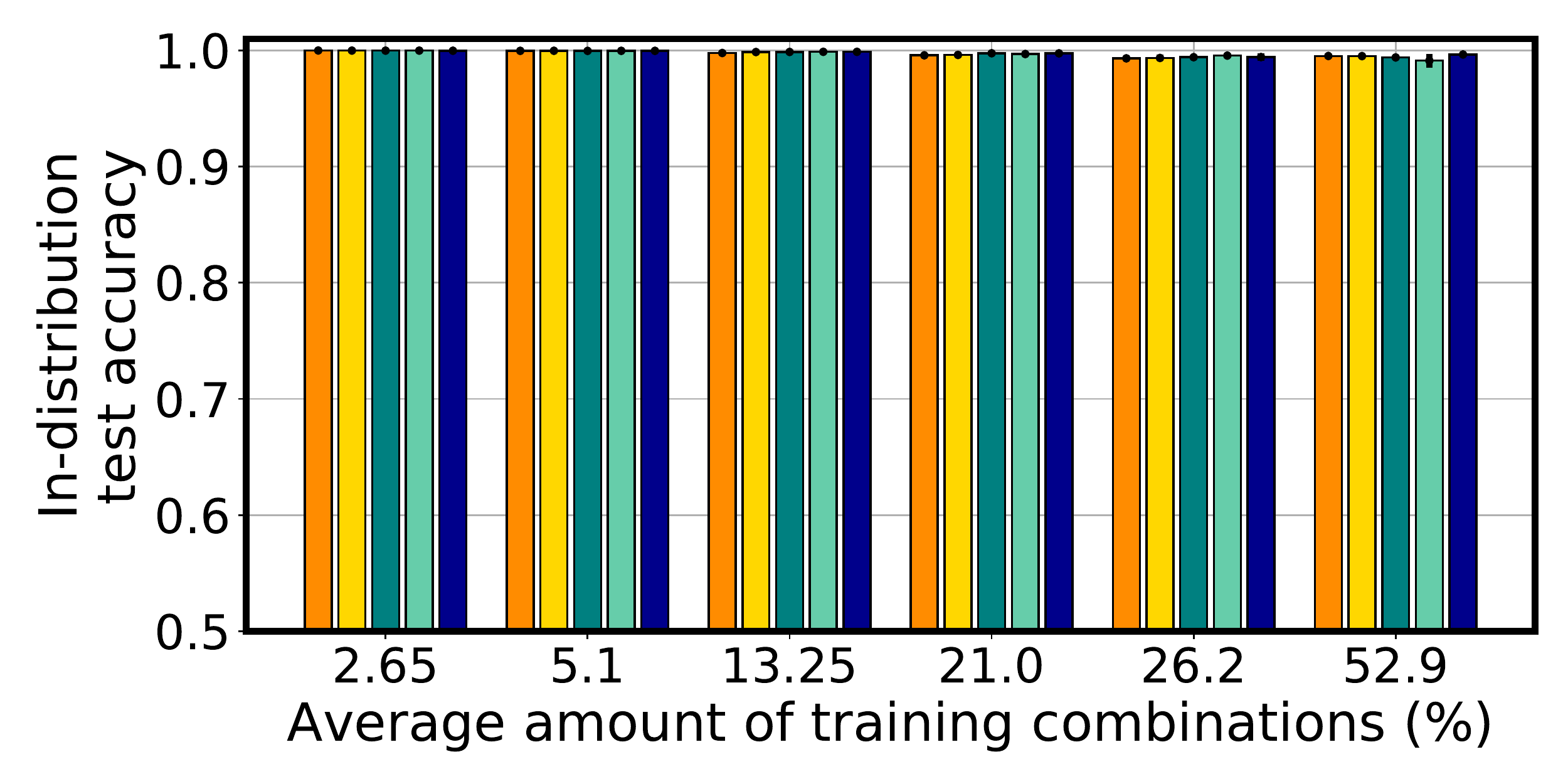} 
    & \hspace{-0.55cm}
    \includegraphics[width=0.5\textwidth]{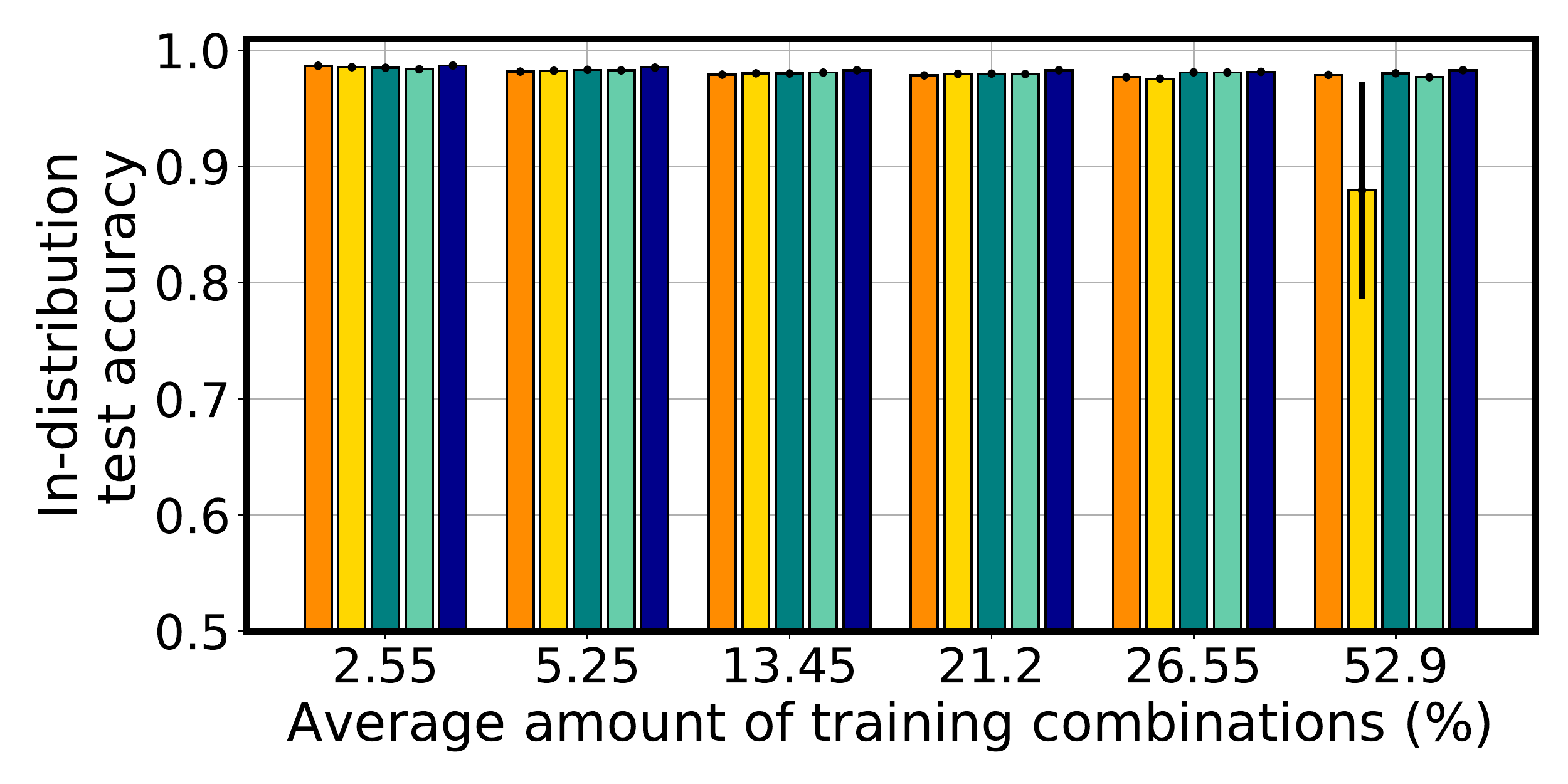}   \\
    (c) & (d)  \\ 
    \end{tabular}
    \caption{In-distribution generalization on the VQA-MNIST dataset, for (a) attribute extraction from single object, (b) attribute extraction from multiple objects, (c) attribute comparison between pairs of separated objects, and (d) comparison between spatial positions.}
    \label{fig:generalization_indistribution_main}
\end{figure}

We aim at comparing the in-distribution and systematic generalization trends for the libraries previously tested on novel combinations, in Figure~\ref{fig:exps}.
The plots in Figure~\ref{fig:generalization_indistribution_main} shows the in-distribution generalization for (a) attribute extraction from single object, (b) attribute extraction from multiple objects, (c) attribute comparison between pairs of separated objects, and (d) comparison between spatial positions. The disparity between testing in-distribution generalization or systematic generalization is dramatic. In absence of a bias, the tasks of attribute extraction and comparison, with the given amount of training examples looks easy. Indeed, the plots across the four VQA tasks show that all the libraries represent equivalently a good choice to achieve generalization within in-distribution testing examples. This result emphasizes the importance of measuring systematic generalization as a strategy to highlight the limited capability of models to generalize to novel combinations.

\subsection{Results for Additional state-of-the-art Libraries and Comparison with Libraries with Modular Image Encoder}\label{sec:results_soa} 

\begin{figure}[!t]
    \centering
    \begin{tabular}{@{\hspace{-0.1cm}}cc}
    \multicolumn{2}{c}{\includegraphics[width=1\textwidth]{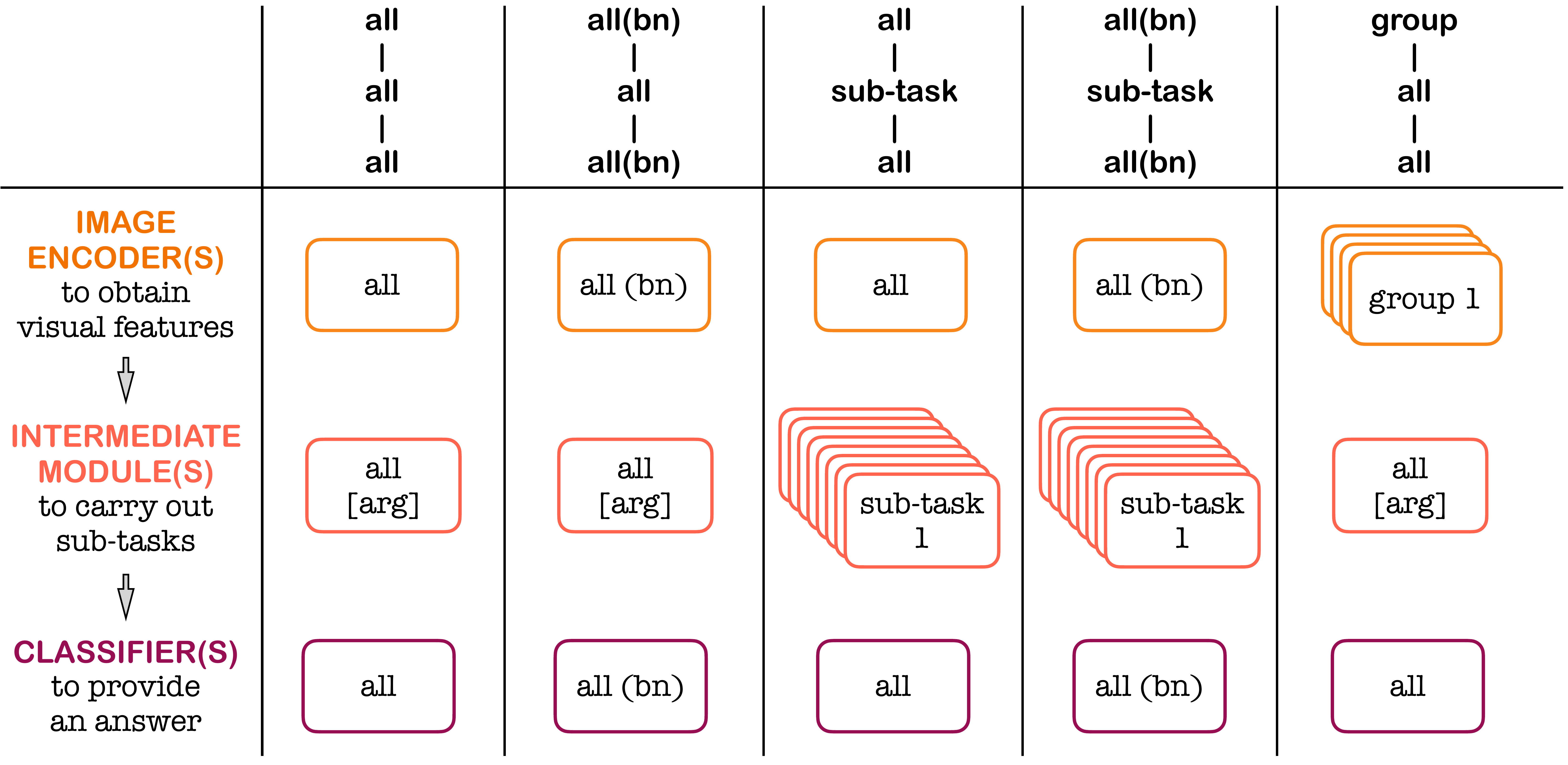}}  \\
    \multicolumn{2}{c}{(a)}  \vspace{0.4cm} \\
    \multicolumn{2}{c}{\includegraphics[width=1\textwidth]{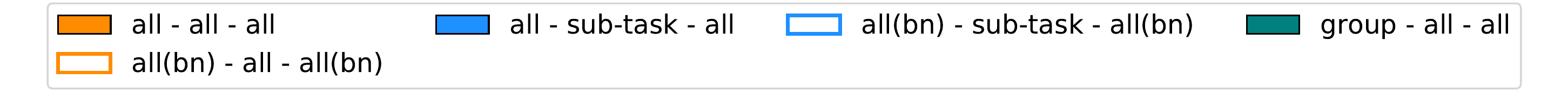}}  \\
    \includegraphics[width=0.5\textwidth]{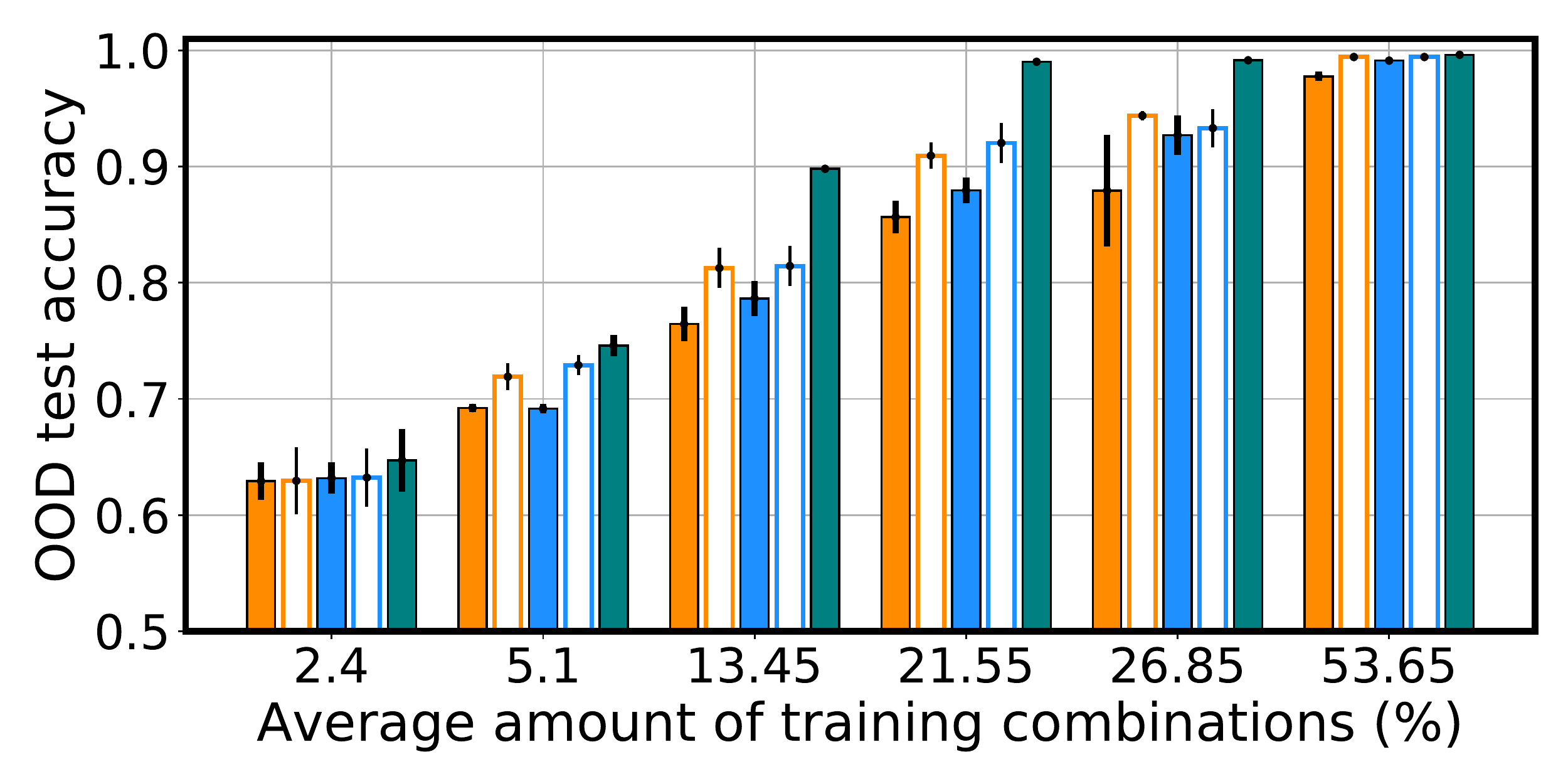} &
    \hspace{-0.8cm}
    \includegraphics[width=0.5\textwidth]{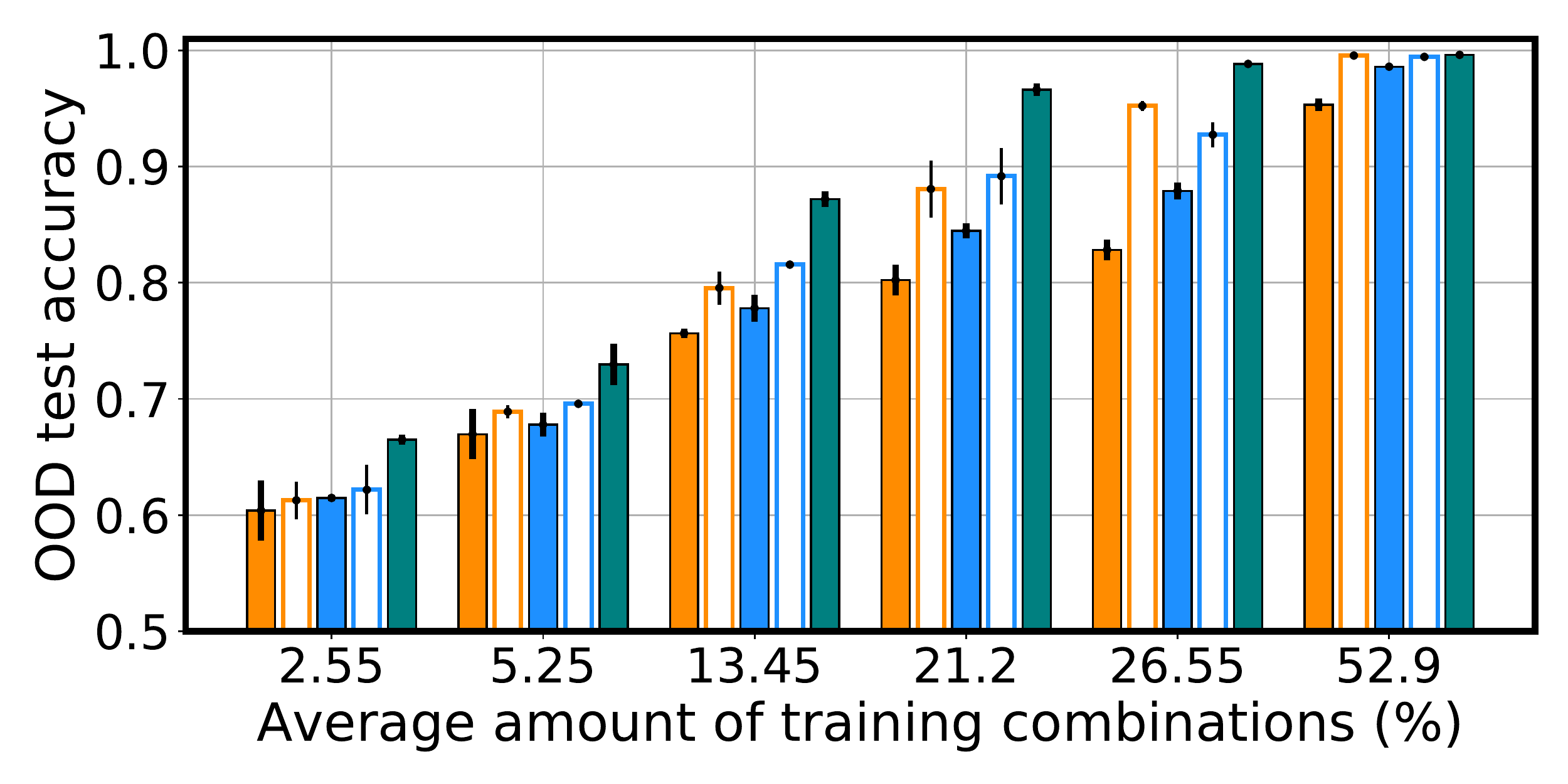} \\
    \vspace{0.4cm}
    (b) & (c) \\
    \includegraphics[width=0.5\textwidth]{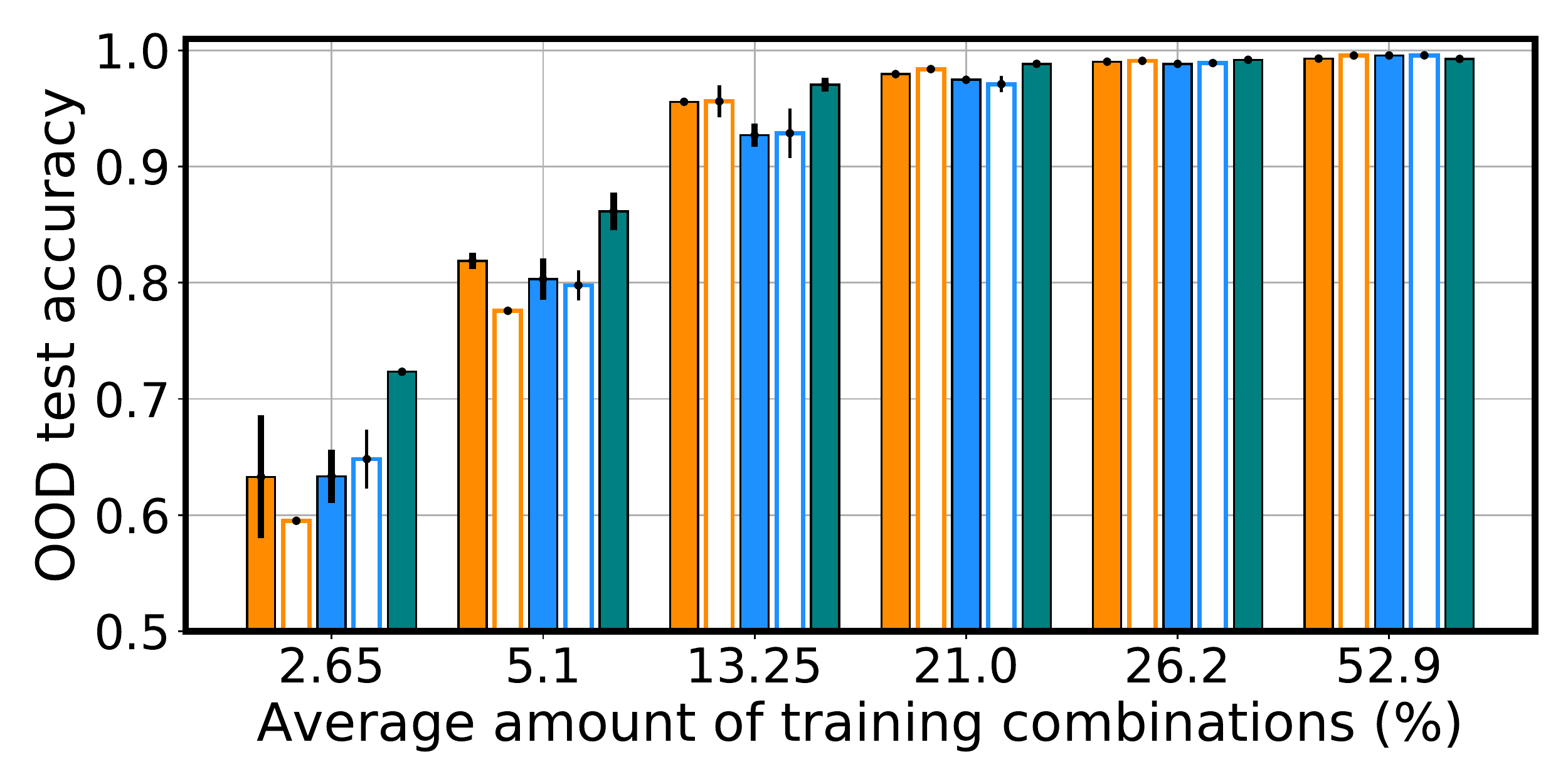} &
    \hspace{-0.8cm}
    \includegraphics[width=0.5\textwidth]{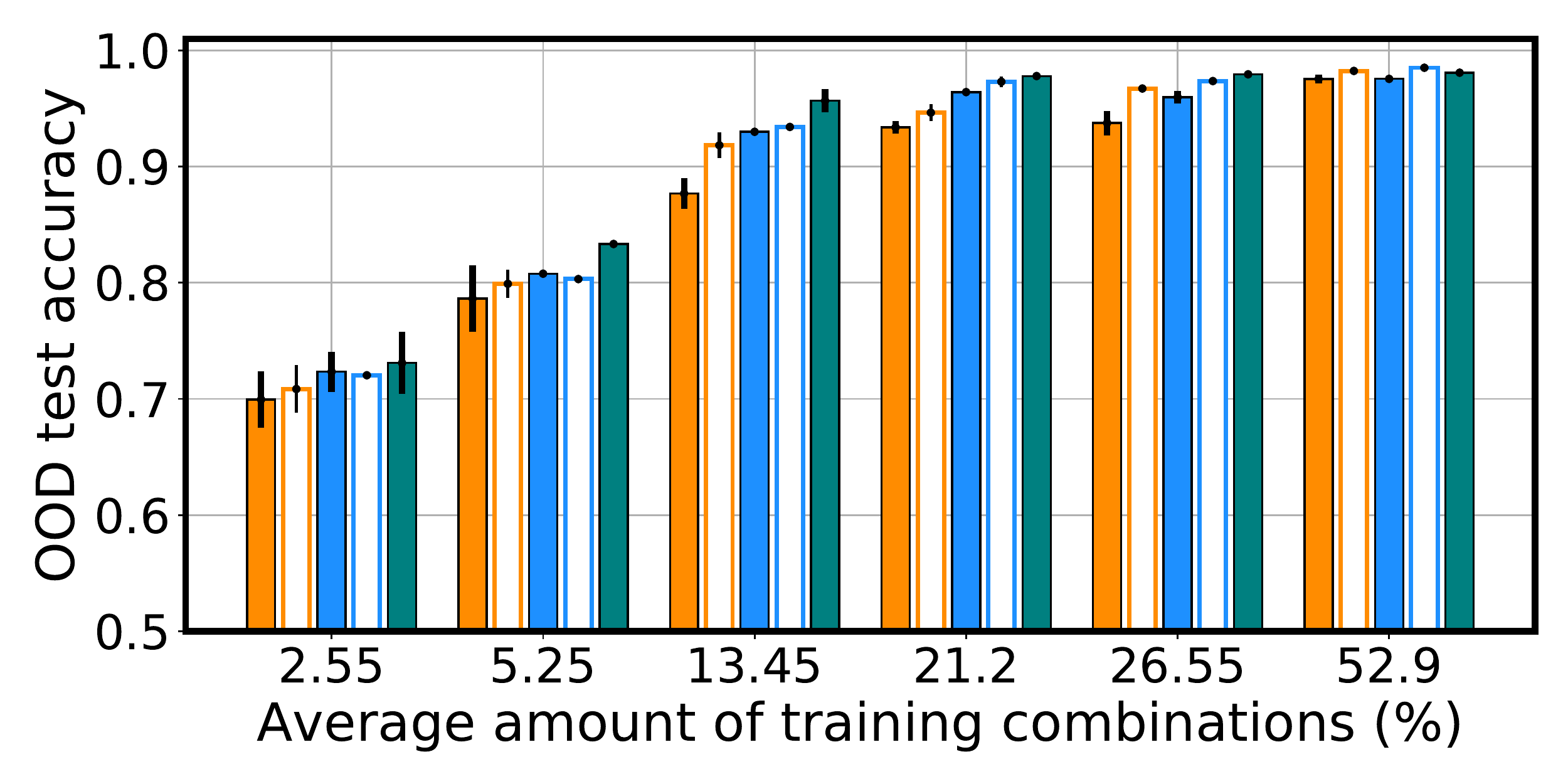}   \\
    (d) & (e)  \\ 
    \end{tabular}
    \caption{Comparison of libraries with modular image encoder and state-of-the-art libraries.  OOD test accuracy refers to systematic generalization accuracy. (a) State-of-the-art libraries and library with modular image encoder (\emph{group - all - all}). Systematic generalization accuracy on (b) attribute extraction from single object, (c) attribute extraction from multiple objects, (d) attribute comparison between pairs of separated objects, and (e) comparison between spatial positions.}
    \label{fig:soa_nmn}
\end{figure}

With the exception of the \emph{all - all - all} library, the libraries in Figure~\ref{fig:ood_modules}a  are not the standard ones from the literature. To confirm our findings on the superiority of the modular image encoder (\emph{group~-~all -~all} in particular) with previous works,
we add the \emph{all~-~sub-task~-~all} library used in \cite{johnson2017inferring}. Its implementation consists of a 
single image encoder and a single classifier, and Residual intermediate modules. Given the usage of a single module at the image encoder and the classifier stages, \emph{all - all - all} library and \emph{all - sub-task - all} library have  variants with batch normalization at both the image encoder and classifier~\cite{bahdanau2018systematic}. We denote these libraries as  \emph{all(bn) - all - all(bn)} and \emph{all(bn) - sub-task - all(bn)}. See Figure~\ref{fig:soa_nmn}a for the depictions of these libraries of modules.

We measure the systematic generalization of those libraries. In Figure~\ref{fig:soa_nmn}b-e, we reported the performance on (b) attribute extraction from single object, (c) attribute extraction from multiple objects, (d) attribute comparison between pairs of separated objects, and (e) comparison between spatial positions. The systematic generalization accuracy for state-of-the-art methods remains clearly below the performance of the \emph{group - all - all} library.

\subsection{Results for Libraries with Modular Image Encoder}
\label{sec:result_modular_image_encoder}

\begin{figure}[!t]
    \centering
    \begin{tabular}{@{\hspace{-0.1cm}}cc}
    \multicolumn{2}{c}{\includegraphics[width=0.9\textwidth]{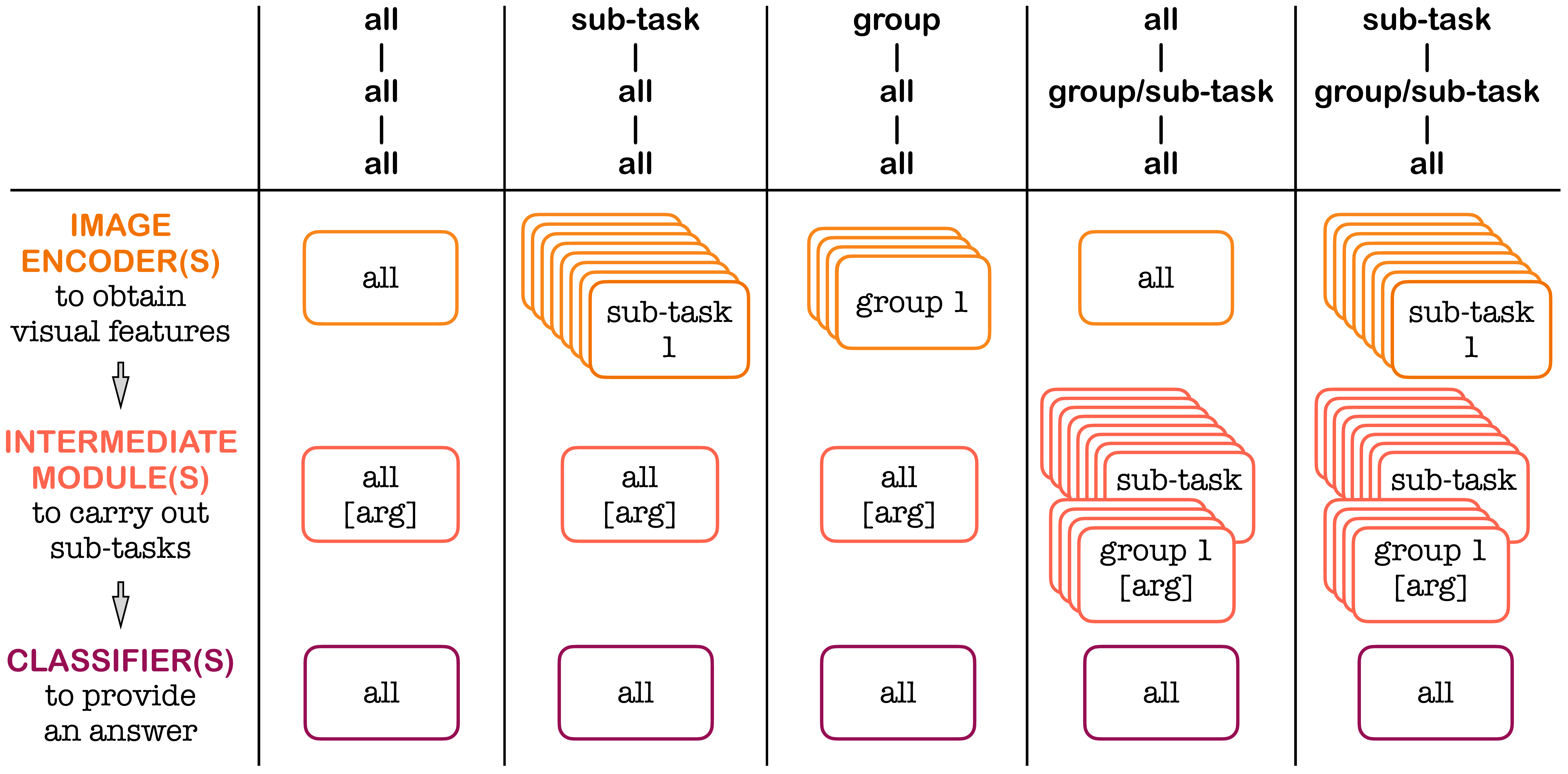}} \\
    \multicolumn{2}{c}{(a)} \vspace{0.4cm}\\
    \multicolumn{2}{c}{\includegraphics[width=0.40\textwidth]{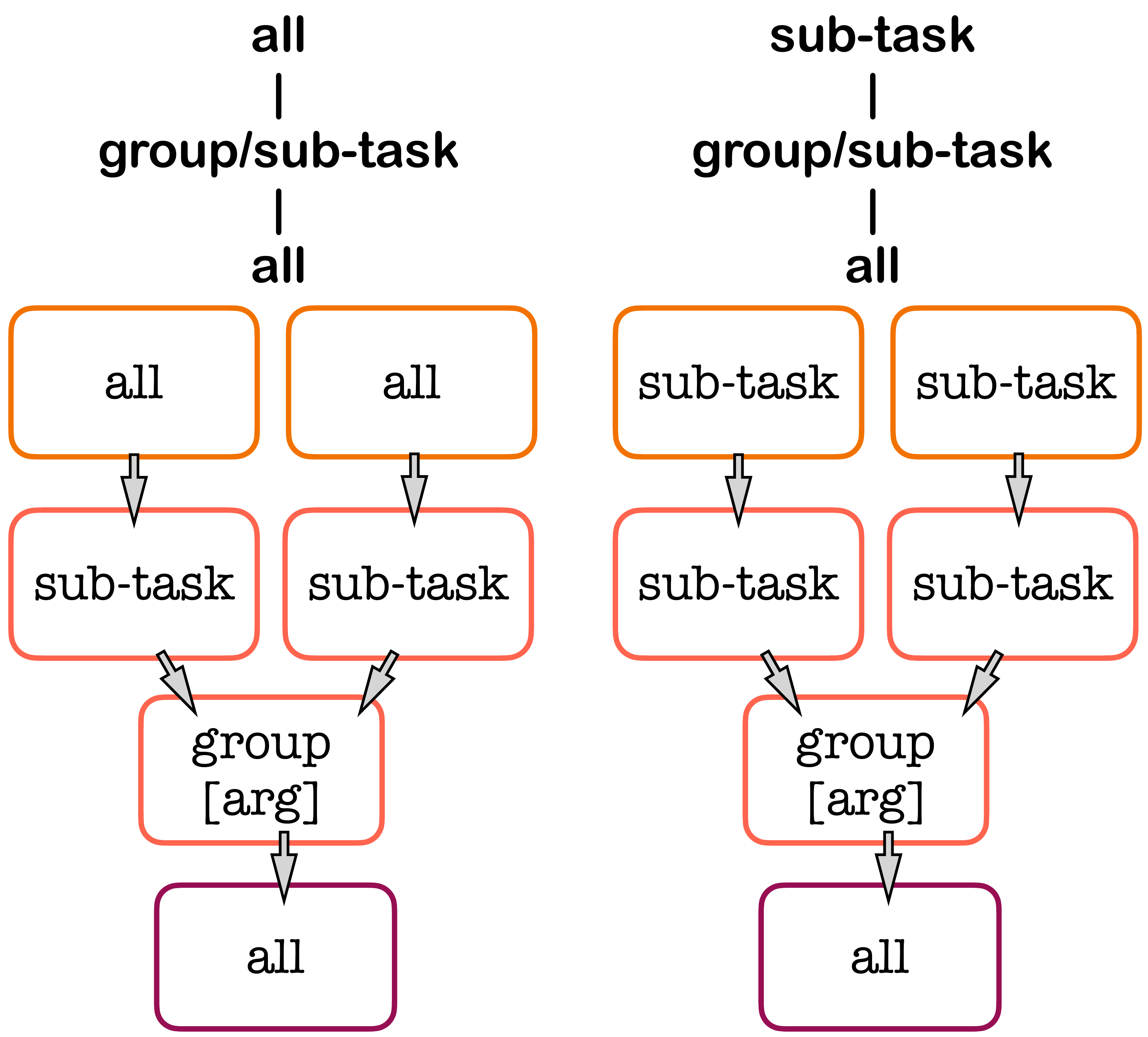}}  \\
    \multicolumn{2}{c}{(b)} \vspace{0.4cm}\\
    \end{tabular}
    \caption{Comparison between libraries with different degrees of modularity at the image encoder stage. (a) Libraries with different degrees of modularity at the image encoder stage, (b) Implementation of libraries with group/sub-task libraries at intermediate module stage. }
    \label{fig:mod_image_encoder_libraries}
\end{figure}

Results in Figure~\ref{fig:exps} highlight the importance of a library of modules at the image encoder stage, together with a medium degree of modularity. To further investigate this aspect, we introduce the \emph{sub-task - all - all} library, an additional library with image encoder modules per each sub-task. We further define two additional libraries for the task of attribute comparison, reported in Figure~\ref{fig:mod_image_encoder_libraries}b. There, given the tree program layout for this task, we rely on libraries with different degrees of modularity at the intermediate modules stage: at the leaves, we use a module per sub-task, while the root has an intermediate module per group of sub-task, where the division of sub-tasks follows the one in Table~\ref{tab:division_comparisons}. The main difference in the libraries of Figure~\ref{fig:mod_image_encoder_libraries}b is at the image encoder stage. On the left, the \emph{all - group/sub-task - all} library relies on a single image encoder module. On the right, the \emph{sub-task - group/sub-task - all} library relies on the largest library of image encoder modules. These two libraries have a single classifier module.

\begin{figure}[!t]
    \centering
    \begin{tabular}{@{\hspace{-0.1cm}}cc}
    \multicolumn{2}{c}{\includegraphics[width=0.95\textwidth]{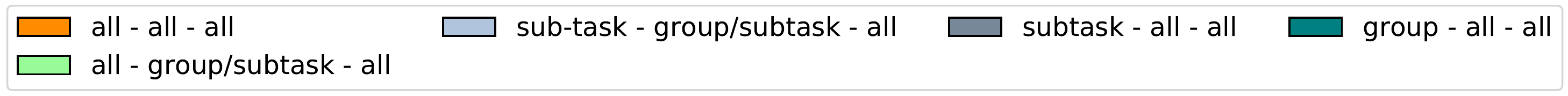}}  \\
    \includegraphics[width=0.5\textwidth]{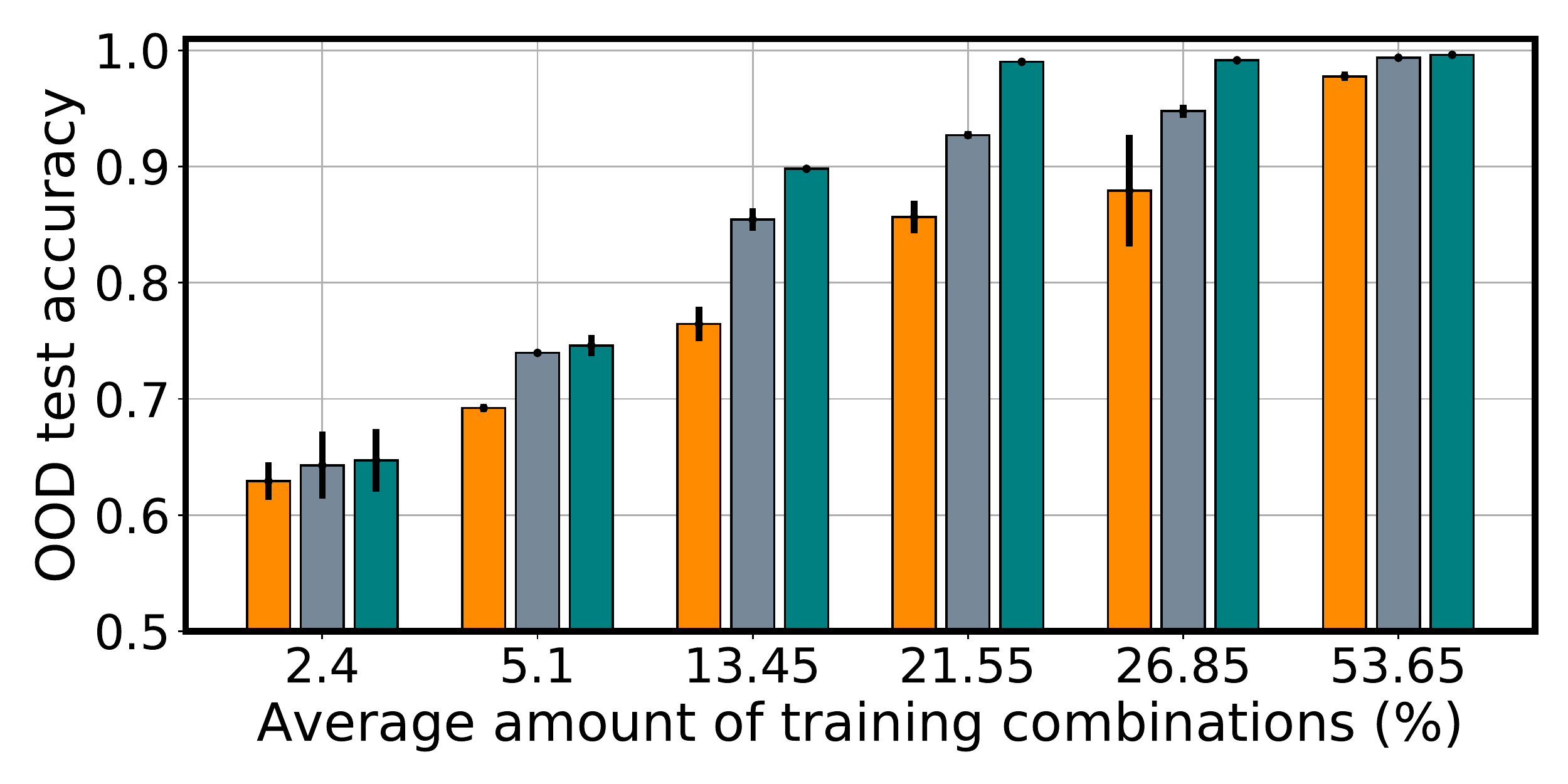} &
    \hspace{-0.5cm} \includegraphics[width=0.5\textwidth]{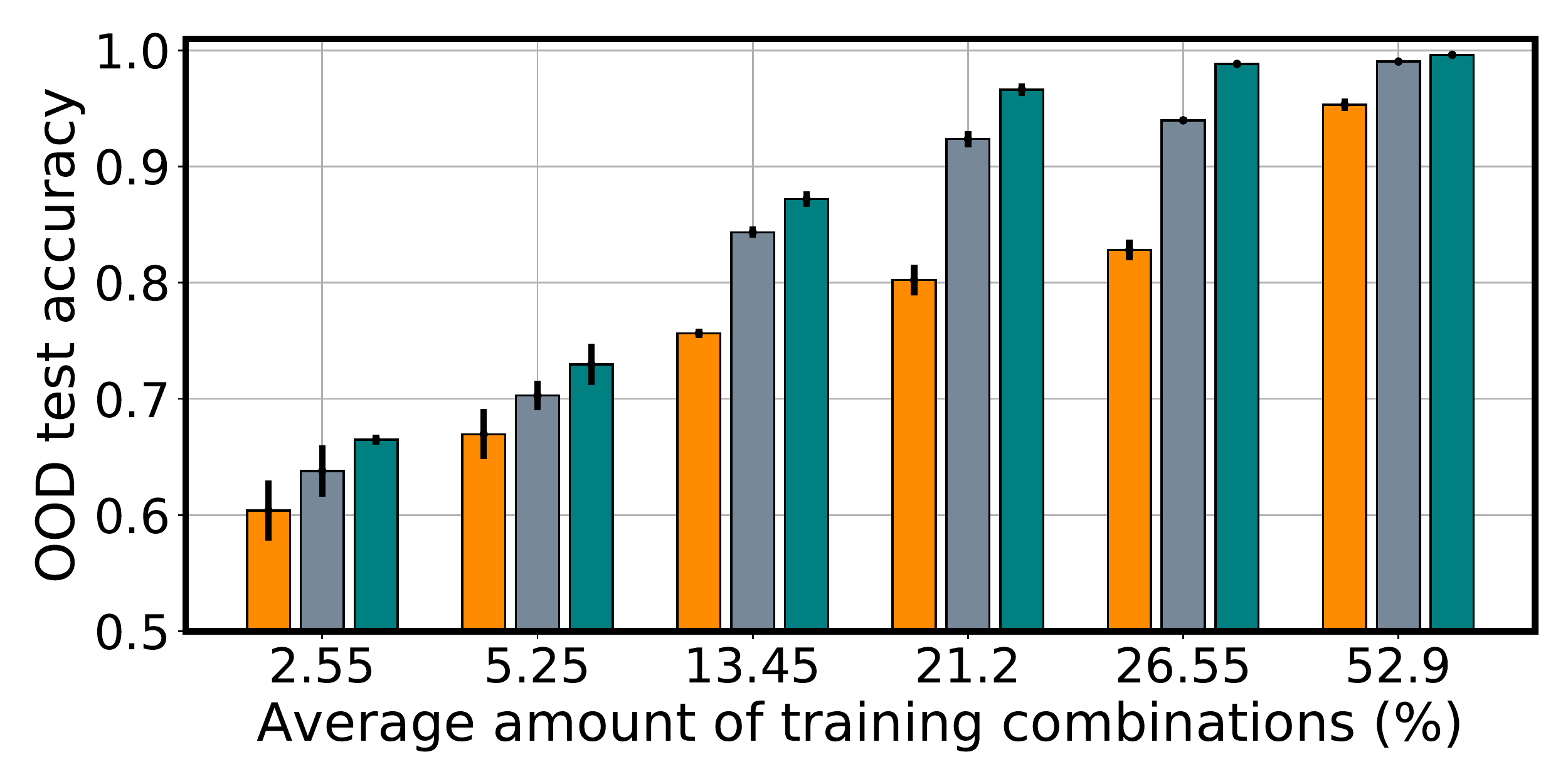} \\
    (a) & (b) \vspace{0.4cm} \\
    \includegraphics[width=0.5\textwidth]{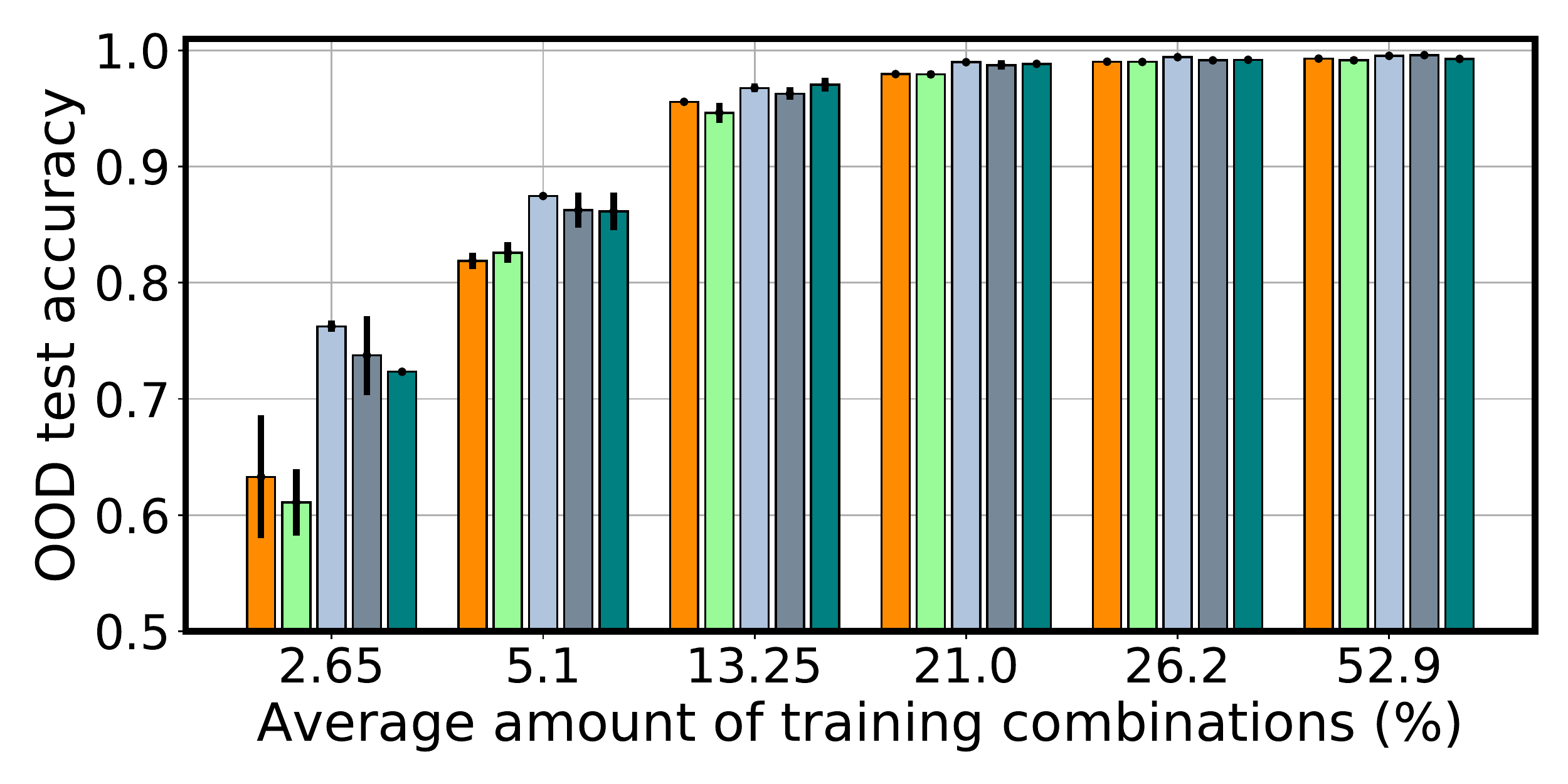} & \hspace{-0.5cm} \includegraphics[width=0.5\textwidth]{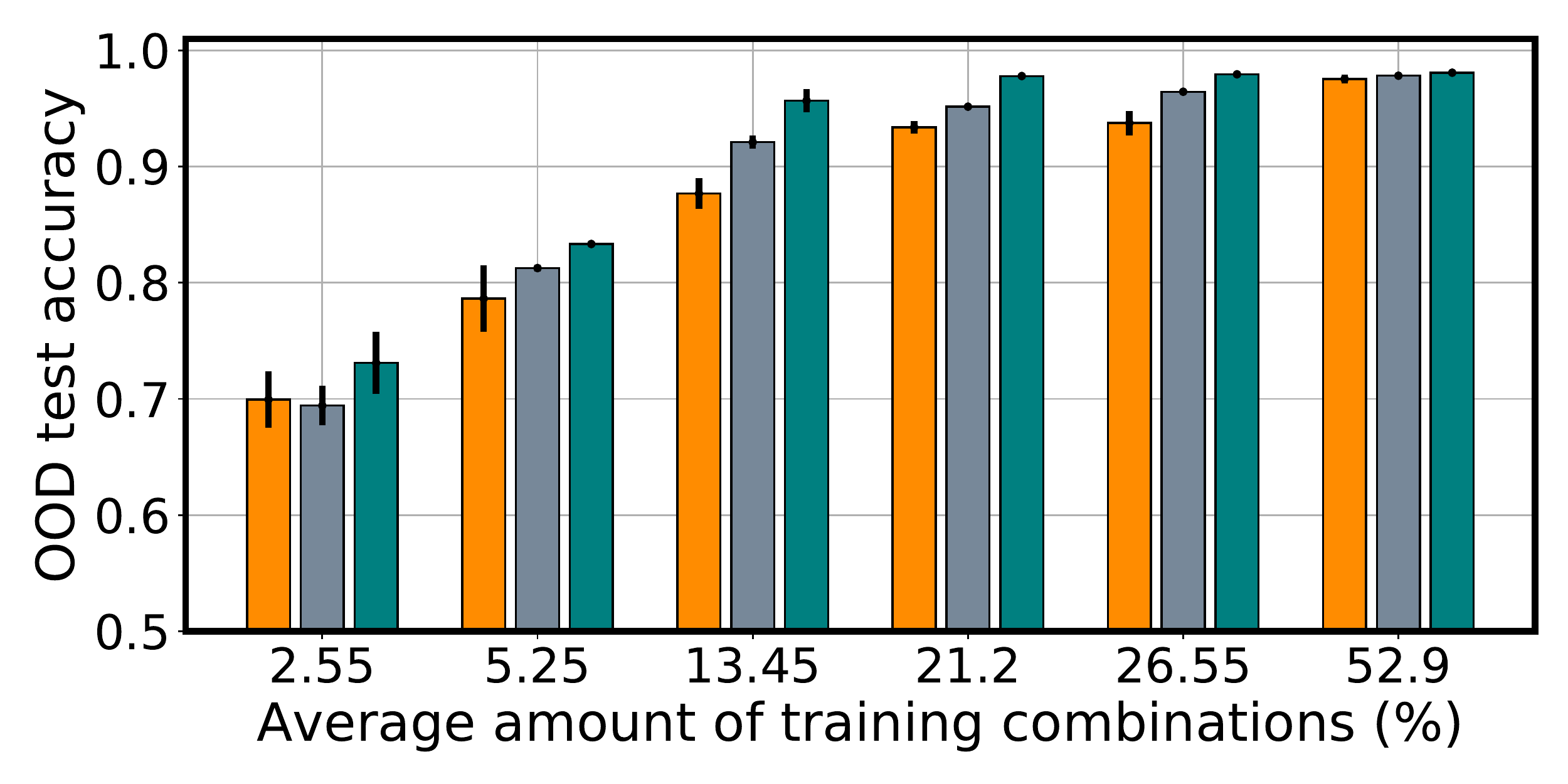}   \\
    (c) & (d)  \\ 
    \end{tabular}
    \caption{Results for comparison between libraries with different degree of modularity at the image encoder stage. OOD test accuracy refers to systematic generalization accuracy. (a-d) Systematic generalization accuracy on (a) attribute extraction from single object, (b) attribute extraction from multiple objects, (c) attribute comparison between pairs of separated objects, and (d) comparison between spatial positions.}
    \label{fig:mod_image_encoder_results}
\end{figure}

We tested the libraries in the first three columns of Figure~\ref{fig:mod_image_encoder_libraries}a on the VQA-MNIST tasks, while the libraries on the last two columns are used on the task of attribute comparison between a pair of separated objects. In Figure~\ref{fig:mod_image_encoder_results}a-d, we reported the systematic generalization performance for the tasks of (a) attribute extraction from single object, (b) attribute extraction from multiple objects, (c) attribute comparison between pairs of separated objects, and (d) comparison between spatial positions. We observe that the \emph{group - all - all} library consistently outperforms the \emph{sub-task - all - all} library across all the tasks. The \emph{sub-task - all - all} library is nonetheless superior to the standard \emph{all - all - all} library. In (c), the modularity of the image encoder library brings advantage also to the network with different degrees of modularity at the intermediate stage.

\subsection{Results for Libraries for Different Implementations of Intermediate Modules per sub-task}\label{sec:results_subtasks}

\begin{figure}[!ht]
    \centering
    \begin{tabular}{@{\hspace{-0.1cm}}cc}
    \multicolumn{2}{c}{\includegraphics[width=0.8\textwidth]{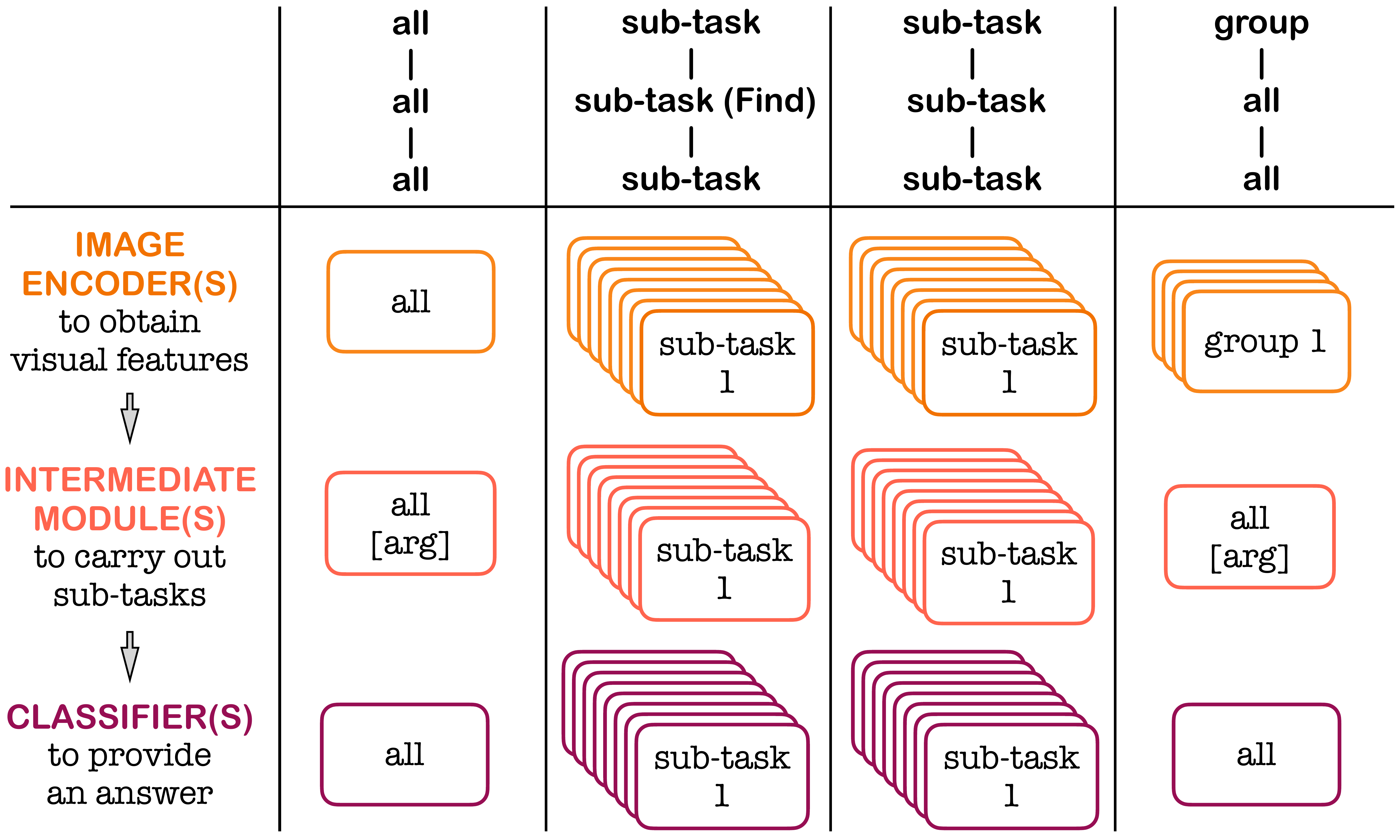}}  \\
    \multicolumn{2}{c}{(a)} \vspace{0.4cm}  \\
    \multicolumn{2}{c}{\includegraphics[width=0.96\textwidth]{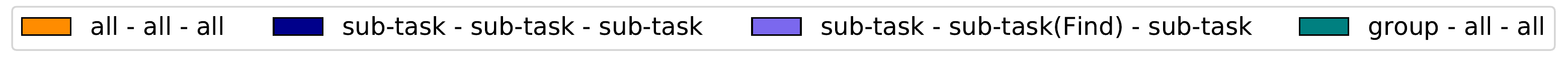}}  \\
    \includegraphics[width=0.5\linewidth]{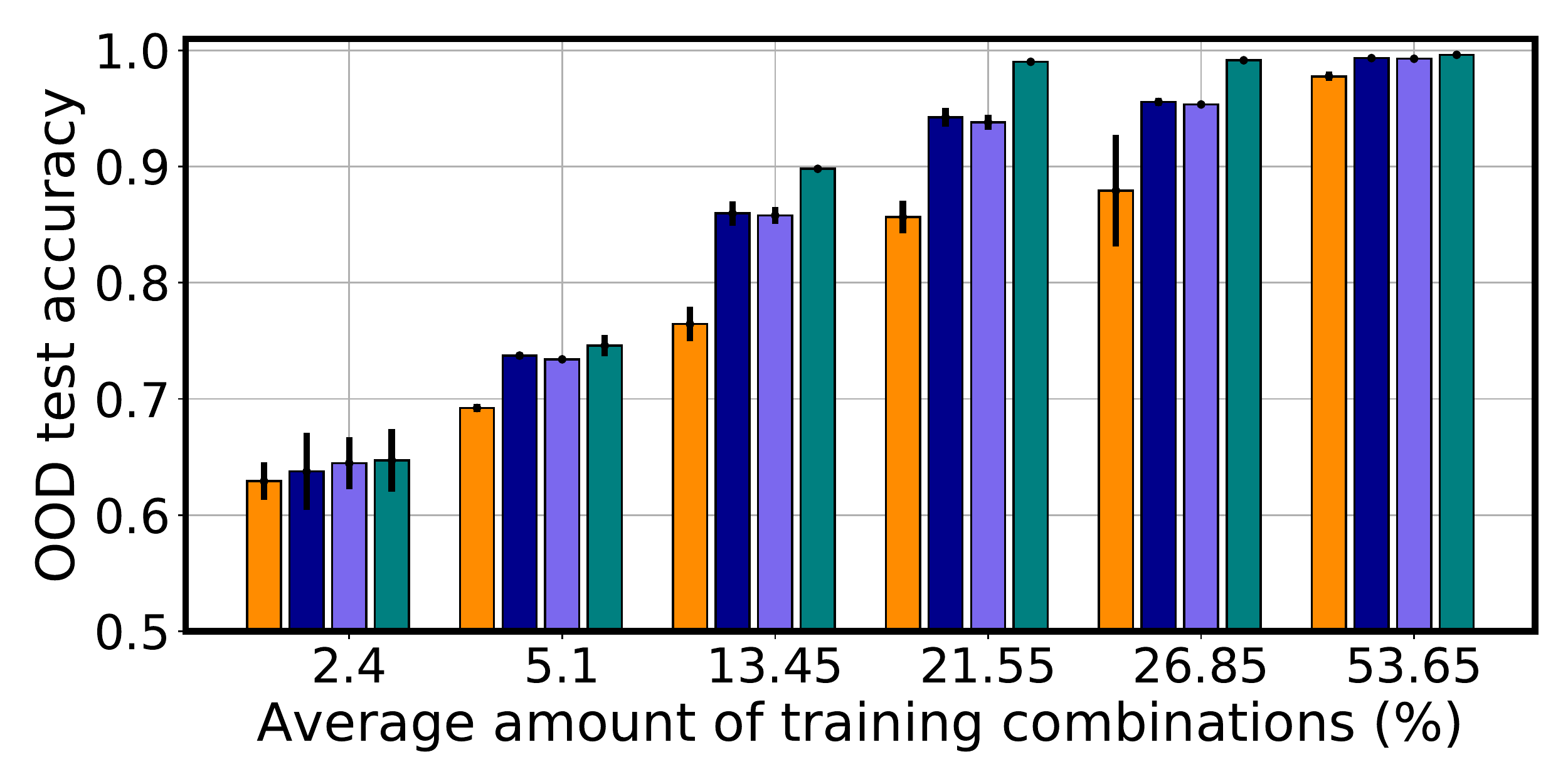} & \hspace{-0.5cm} \includegraphics[width=0.5\columnwidth]{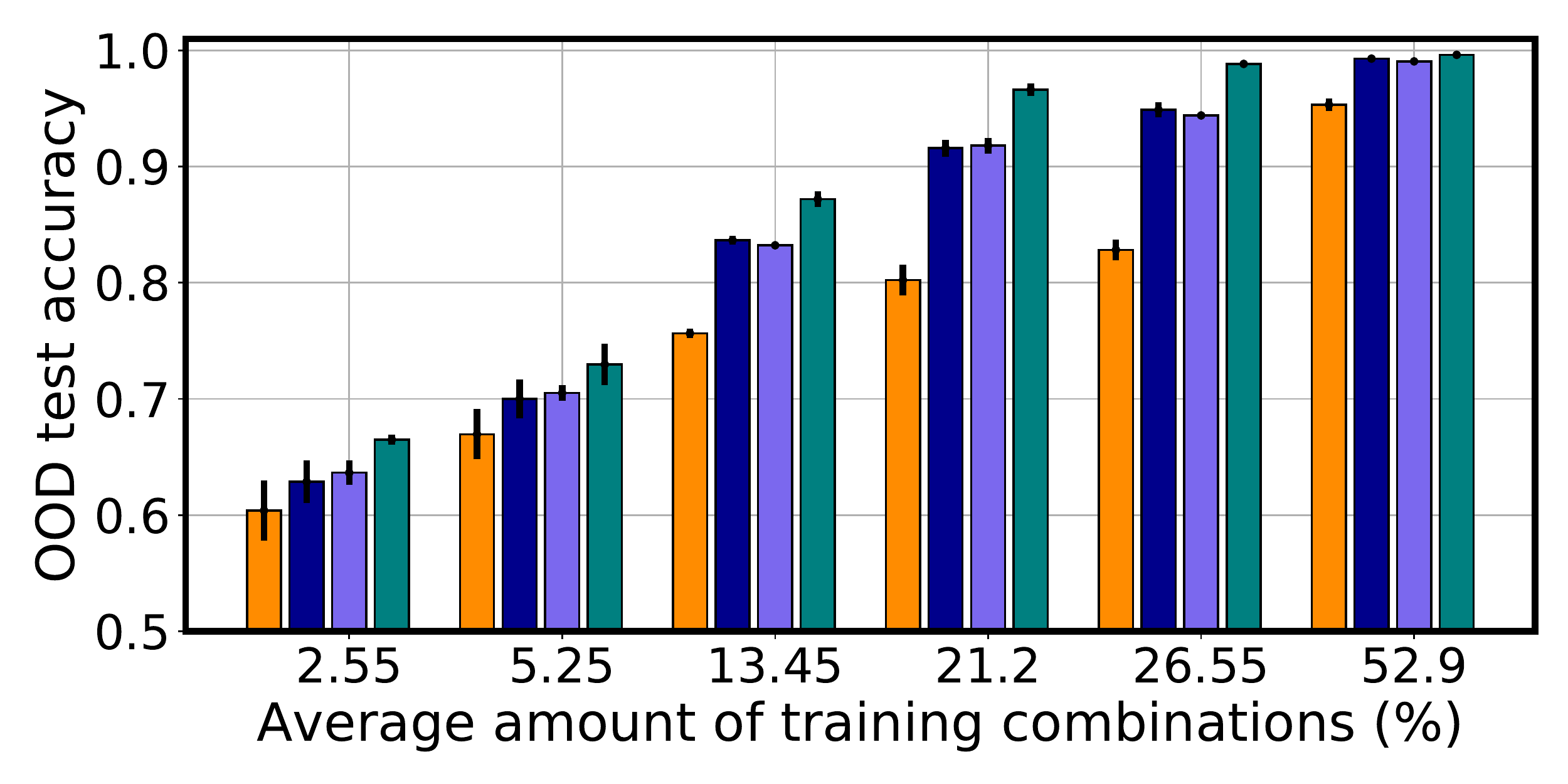} \\
    (b) & (c) \vspace{0.4cm} \\
    \includegraphics[width=0.5\textwidth]{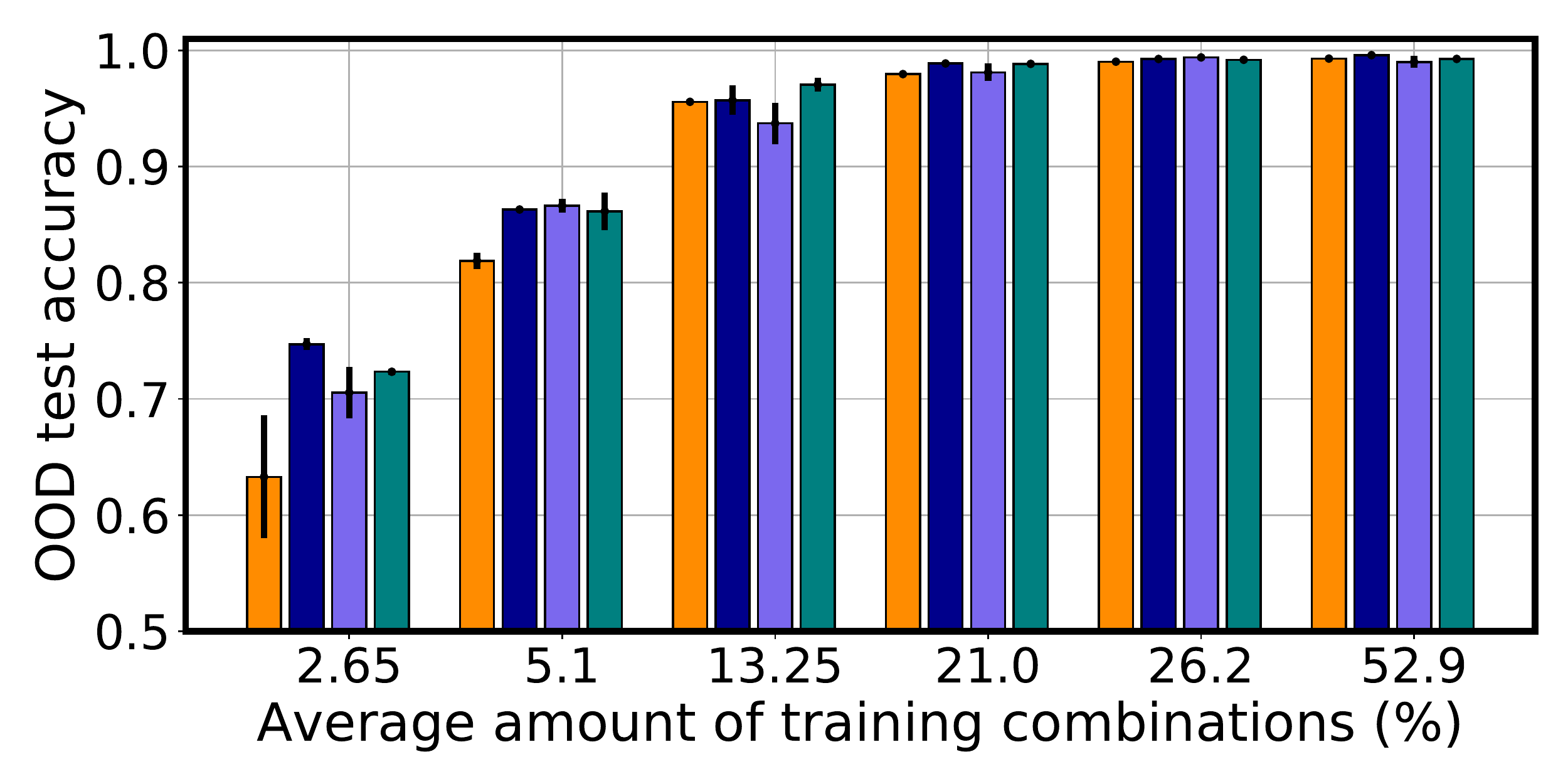} & \hspace{-0.5cm} \includegraphics[width=0.5\textwidth]{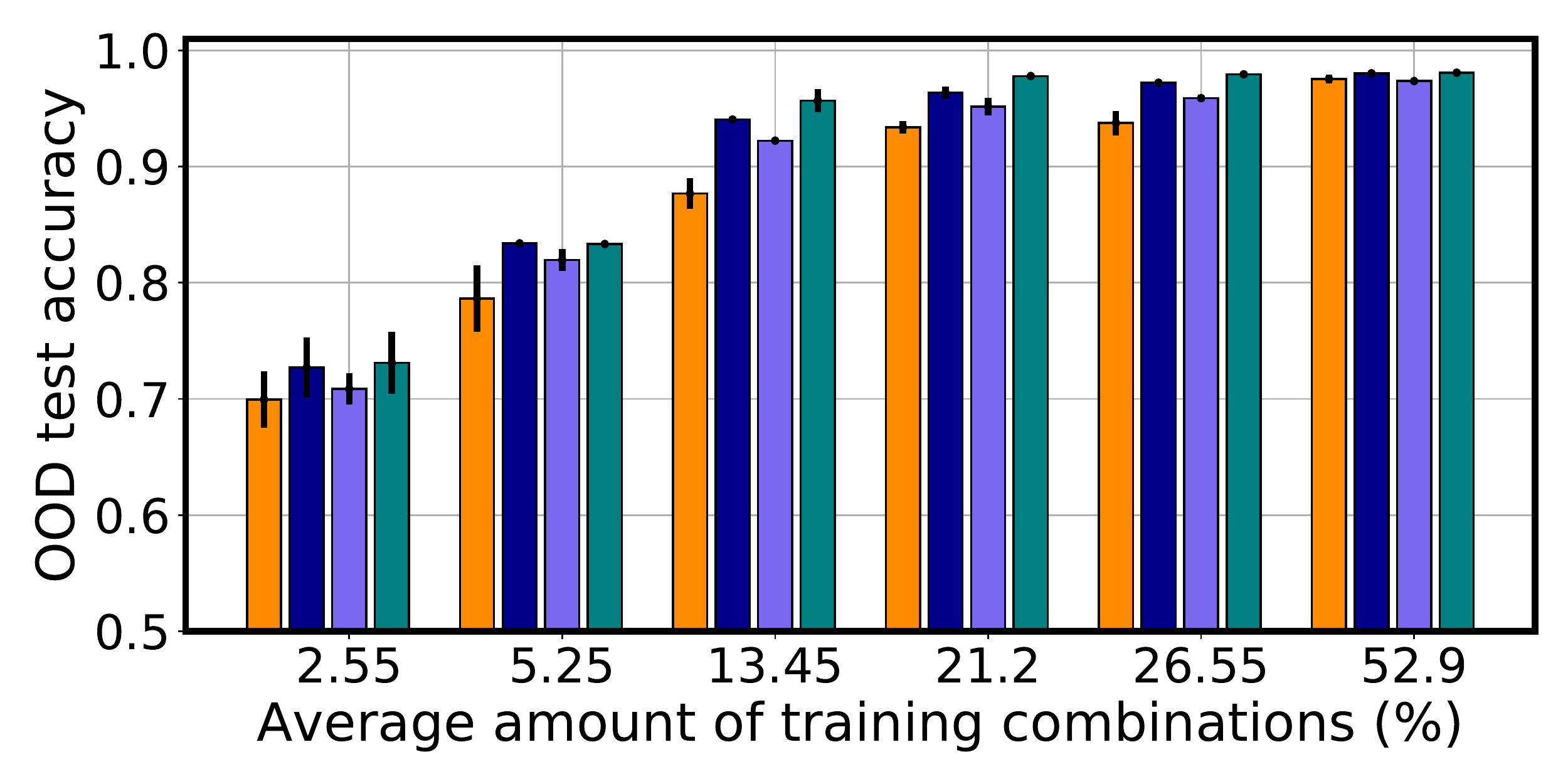}   \\
    (d) & (e)  \\ 
    \end{tabular}
    \caption{Comparison between libraries of same degree of modularity, but different implementation of the intermediate modules. OOD test accuracy refers to systematic generalization accuracy. (a) Two libraries with different implementation of the intermediate modules and  \emph{all - all - all} and  \emph{group - all - all} libraries for reference. Systematic generalization accuracy on (b) attribute extraction from single object, (c) attribute extraction from multiple objects, (d) attribute comparison between pairs of separated objects, and (e) comparison between spatial positions.}
    \label{fig:different_subtasks}
\end{figure}

To exclude that the different implementation of the Residual from the Find module could be a reason for the poorer systematic generalization performance of the \emph{sub-task - sub-task - sub-task} library, we introduce a similar library, which we call \emph{sub-task - sub-task(Find) - sub-task} library. In the latter, the implementation of the intermediate modules rely on the Find module as in Equations~\ref{eq:def_group}-\ref{eq:spec_type}, where each sub-task defines a group. 
See Figure~\ref{fig:different_subtasks}a for the depictions of these libraries of modules.

Figure~\ref{fig:different_subtasks}b-e shows the systematic generalization of the libraries of Figure~\ref{fig:different_subtasks}a, trained on (b) attribute extraction from single object, (c) attribute extraction from multiple objects, (d) attribute comparison between pairs of separated objects, and (e) comparison between spatial positions. We observe that libraries with sub-task degree of modularity at the image encoder stage perform quite similar independently from the implementation of their intermediate modules, and show consistently lower performance than the \emph{group - all - all} library.

\subsection{Results for Libraries with Modular Classifier}\label{sec:result_modular_classifier}

\begin{figure}[!ht]
    \centering
    \begin{tabular}{@{\hspace{-0.1cm}}cc}
    \multicolumn{2}{c}{\includegraphics[width=0.8\textwidth]{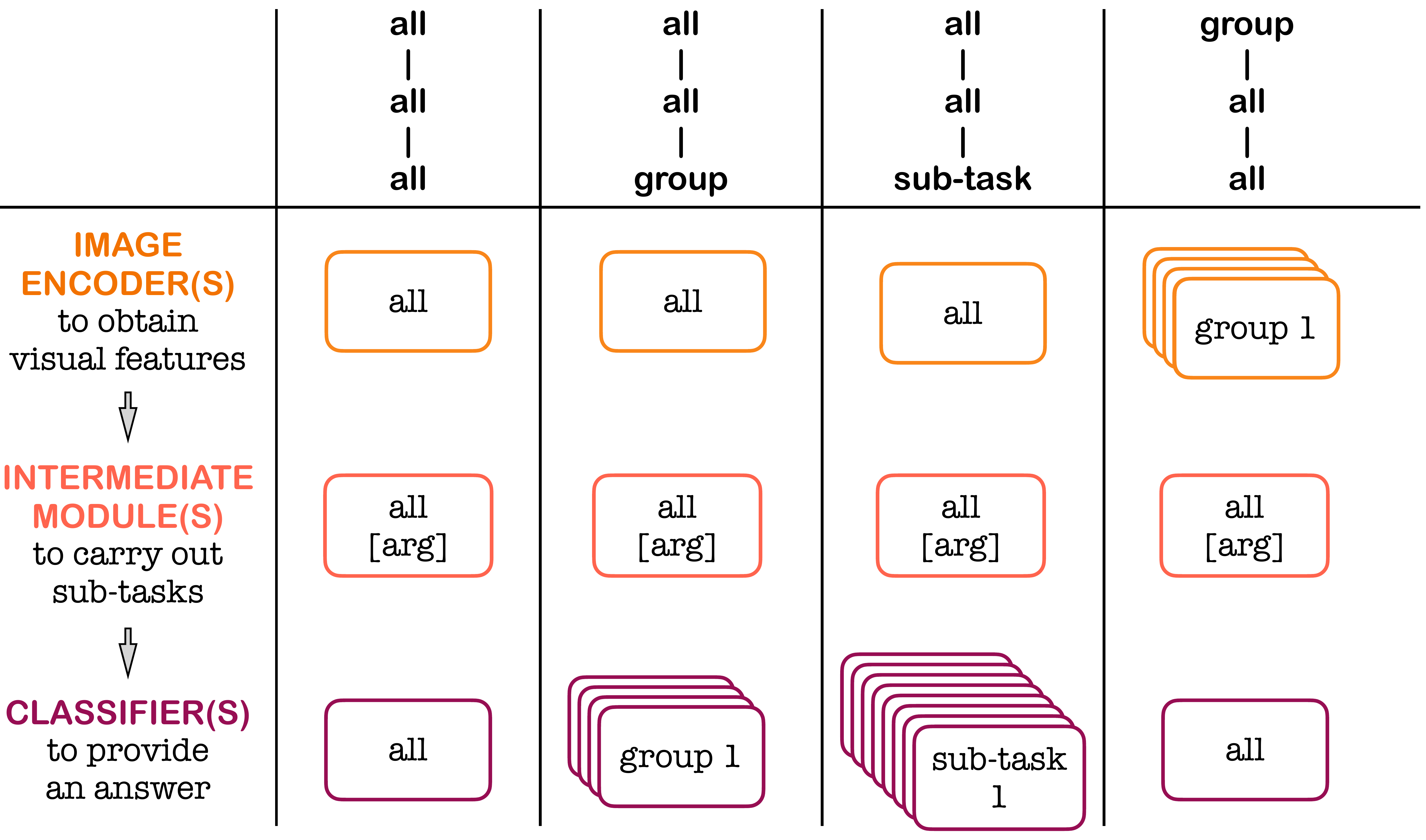}}  \\
    \multicolumn{2}{c}{(a)} \vspace{0.4cm} \\
    \multicolumn{2}{c}{\includegraphics[width=1\textwidth]{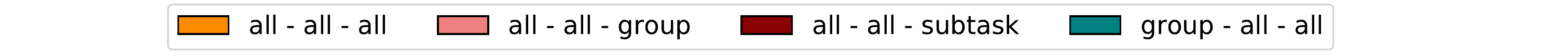}}  \\
    \includegraphics[width=0.5\textwidth]{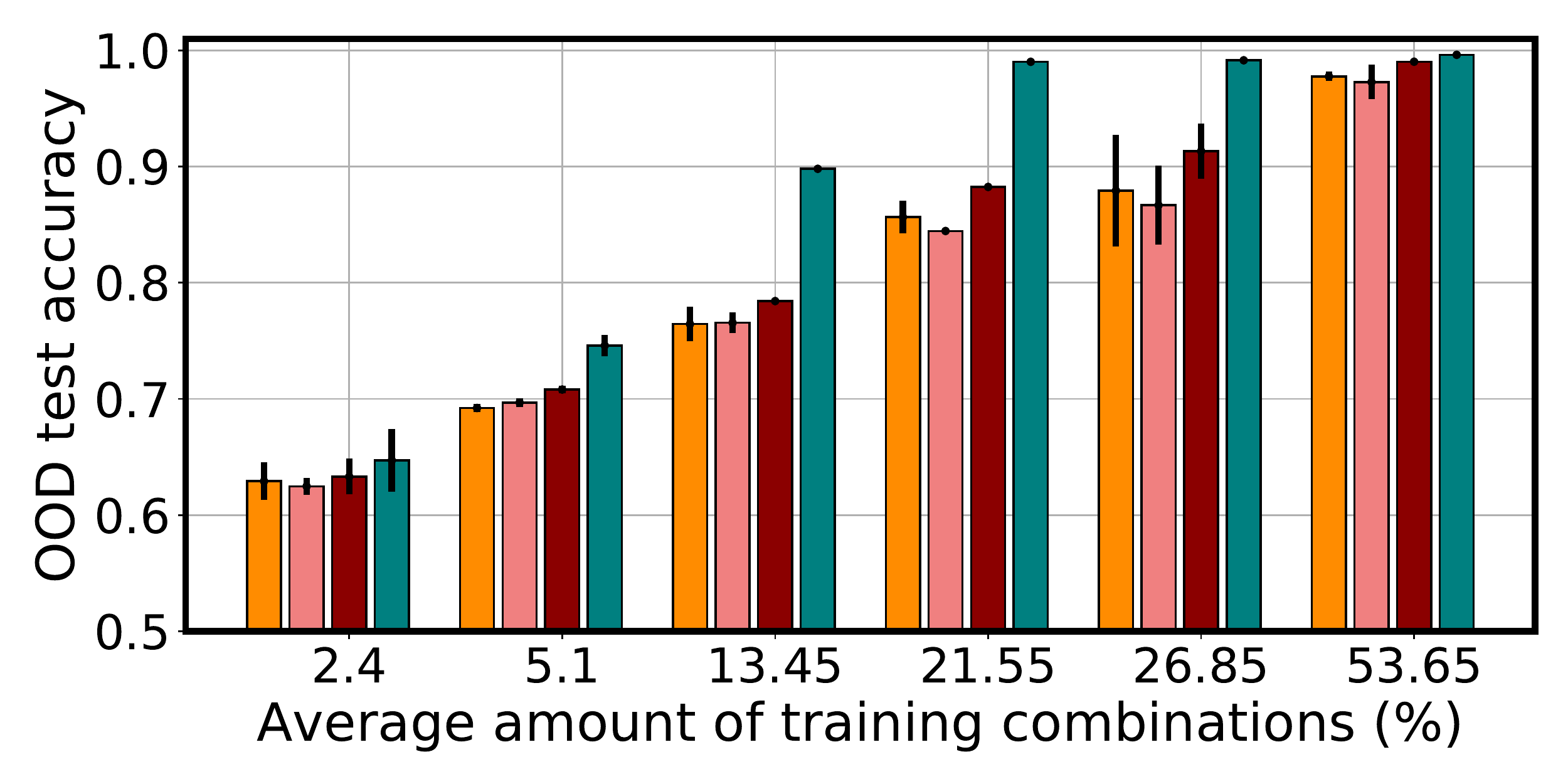} & \hspace{-0.5cm} \includegraphics[width=0.5\textwidth]{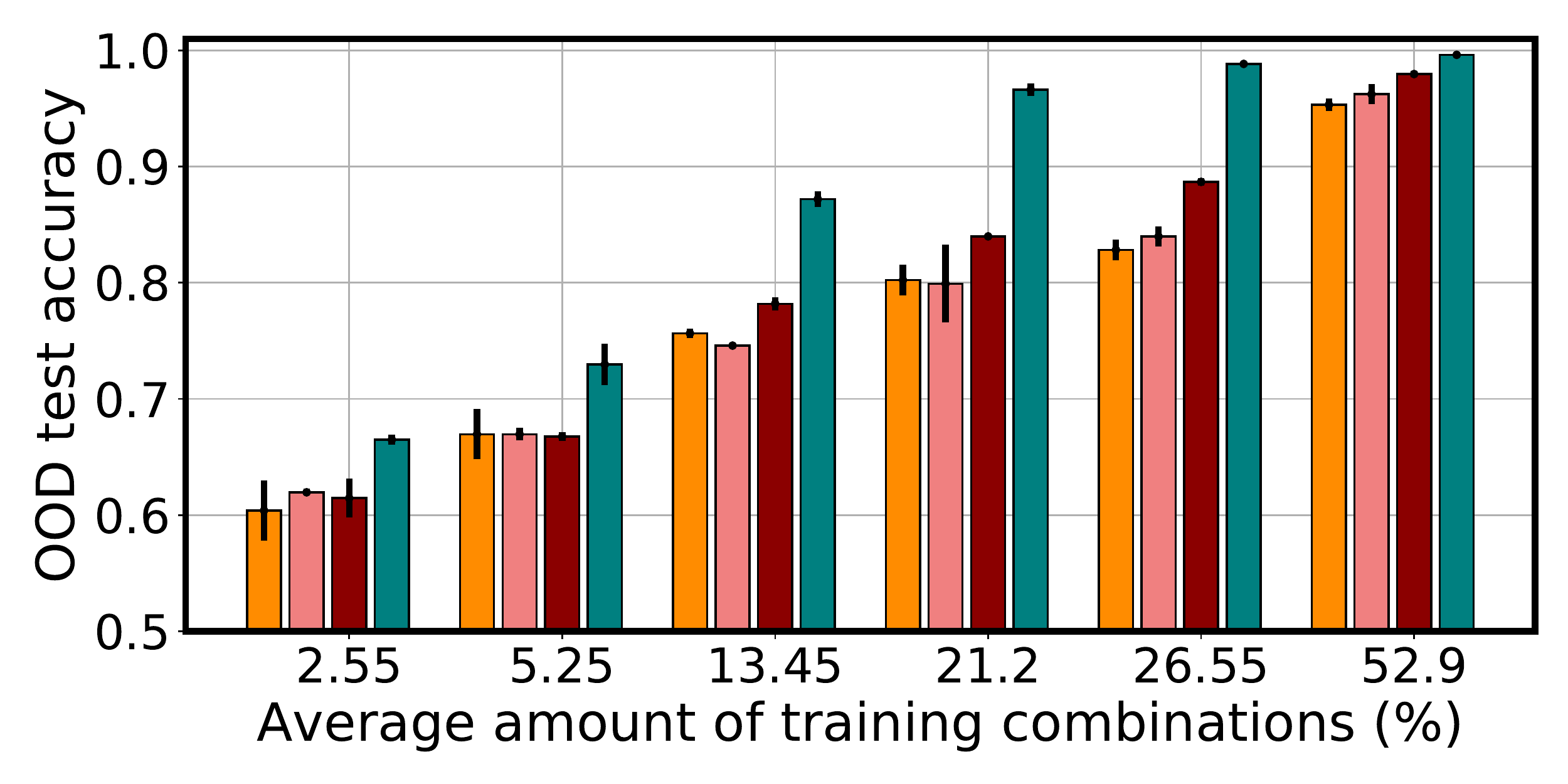} \\
    \vspace{0.4cm}
    (b) & (c) \\
    \includegraphics[width=0.5\textwidth]{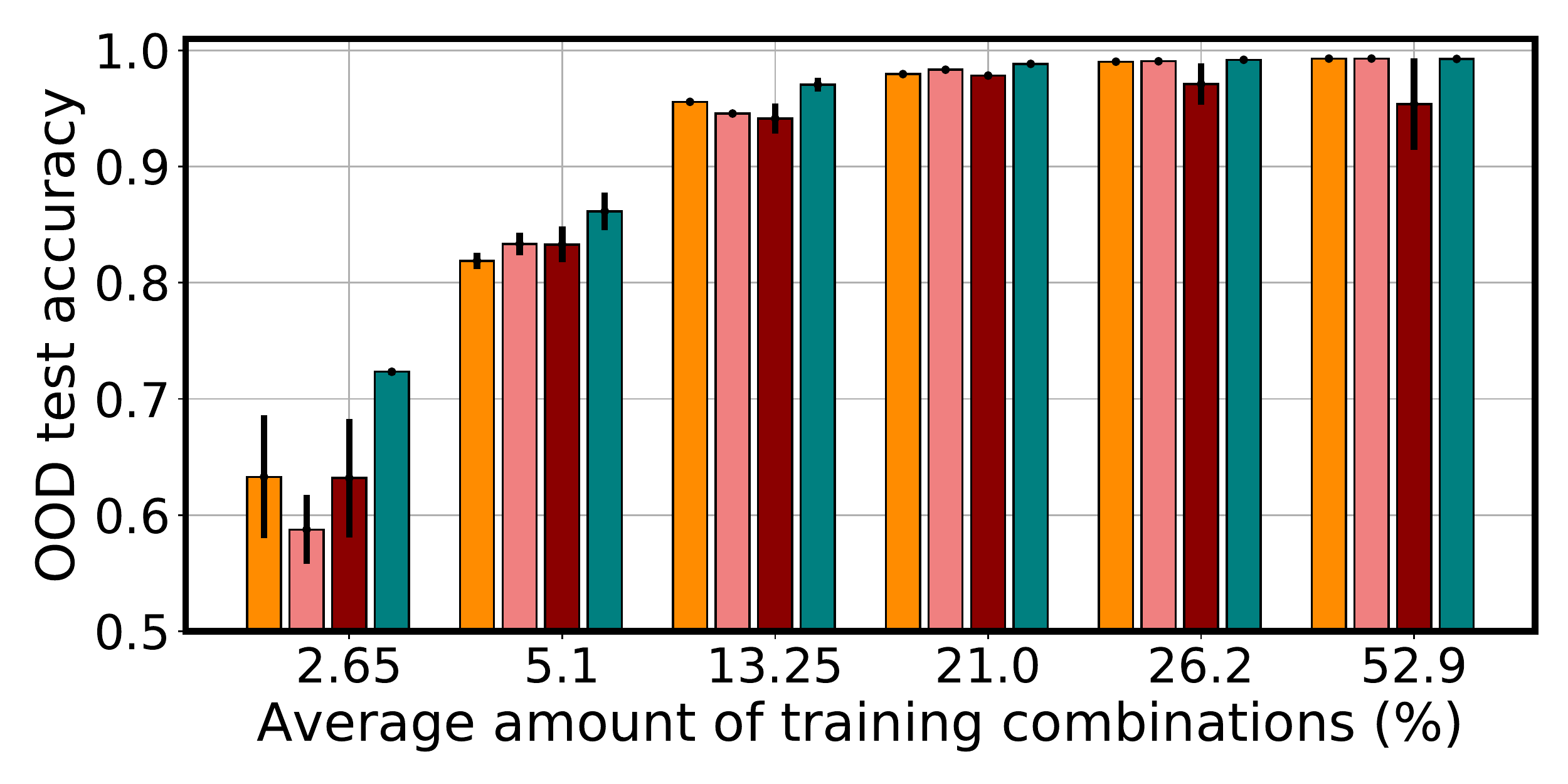} & \hspace{-0.5cm} \includegraphics[width=0.5\textwidth]{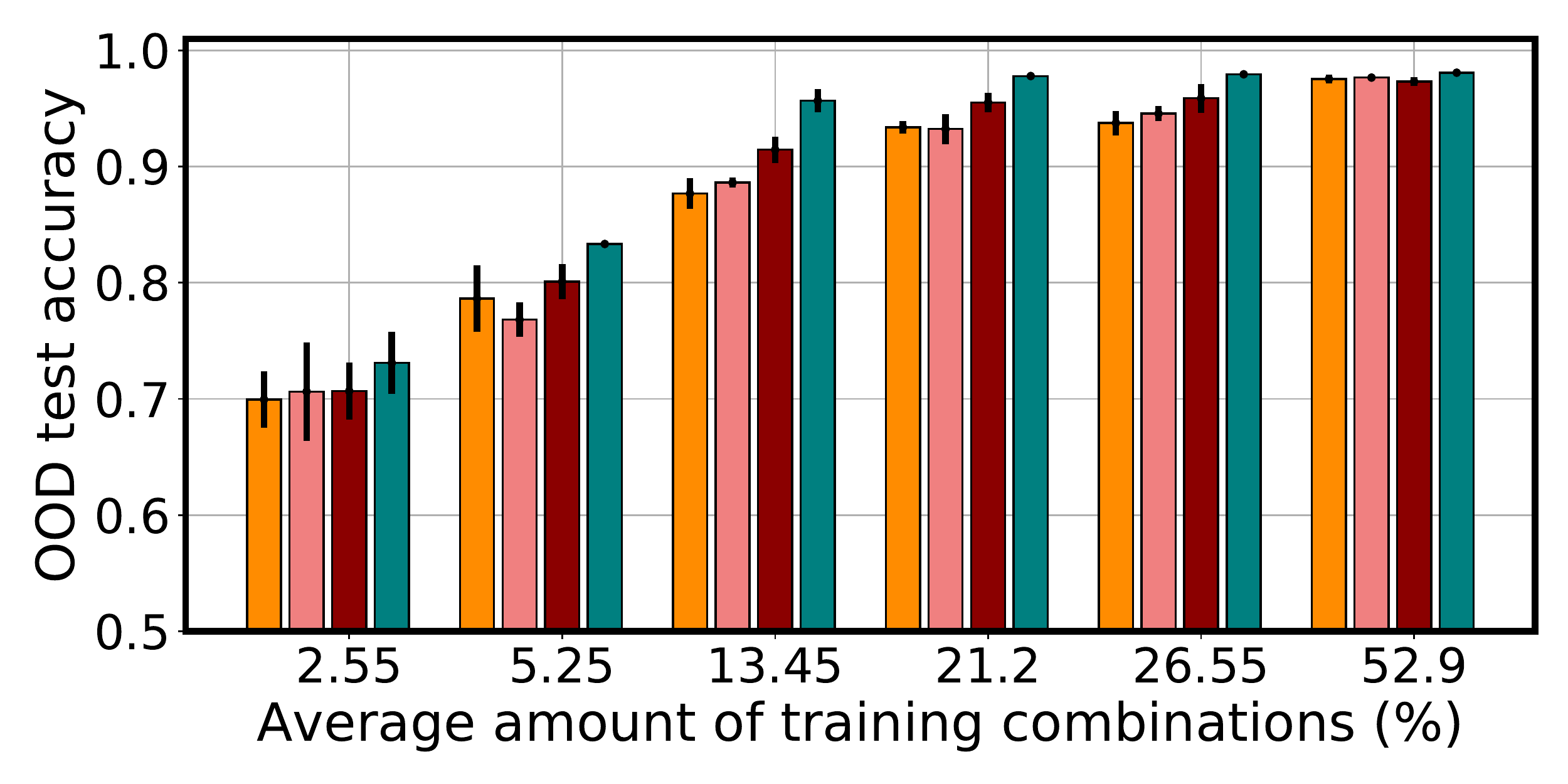}   \\
    (d) & (e)  \\ 
    \end{tabular}
    \caption{Comparison between libraries with different degrees of modularity at the classifier stage. OOD test accuracy refers to systematic generalization accuracy. (a) Three libraries with different degrees of modularity at the classifier stage and the \emph{group - all - all} library for reference. Systematic generalization accuracy on (b) attribute extraction from single object, (c) attribute extraction from multiple objects, (d) attribute comparison between pairs of separated objects, and (e) comparison between spatial positions.}
    \label{fig:modular_classifier}
\end{figure}

We further consider the effect of libraries with different degrees of modularity at the classifier stage on systematic generalization.
To do so, we introduce two libraries with a shared image encoder module and shared intermediate modules. As before,  we envision three degrees of modularity at the classifier stage: shared module, that is the \emph{all - all - all} library, per group of sub-tasks, that is, the \emph{all - all - group} library, and per sub-task, that is, the \emph{all - all - sub-task} library. Figure~\ref{fig:modular_classifier}a depicts these libraries.

The systematic generalization performance for the libraries in Figure~\ref{fig:modular_classifier}a across the VQA-MNIST tasks is reported in Figure~\ref{fig:modular_classifier}b-e. The conclusion here is that the degree of modularity of the library at the classifier stage alone does not bring any advantages across the tested tasks. 

\subsection{Results on More Training Examples}\label{app:more_datapoints}
\begin{figure*}[!ht]
    \centering
    \begin{tabular}{cc}
    \multicolumn{2}{c}{\includegraphics[width=0.9\linewidth]{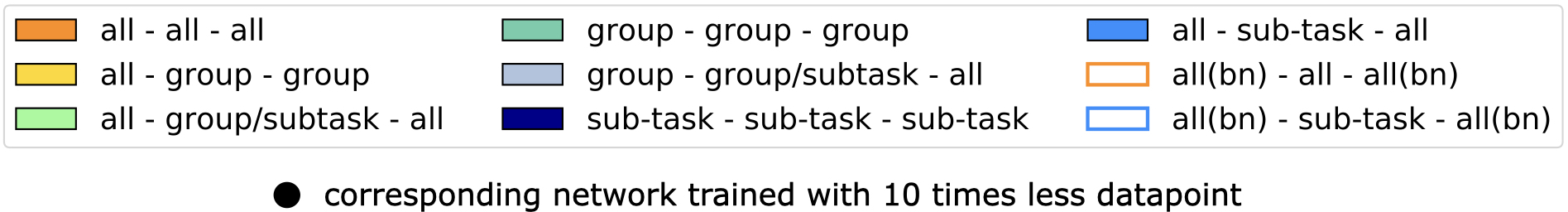}} \\
    \hspace{-0.2cm} \includegraphics[width=.4\linewidth]{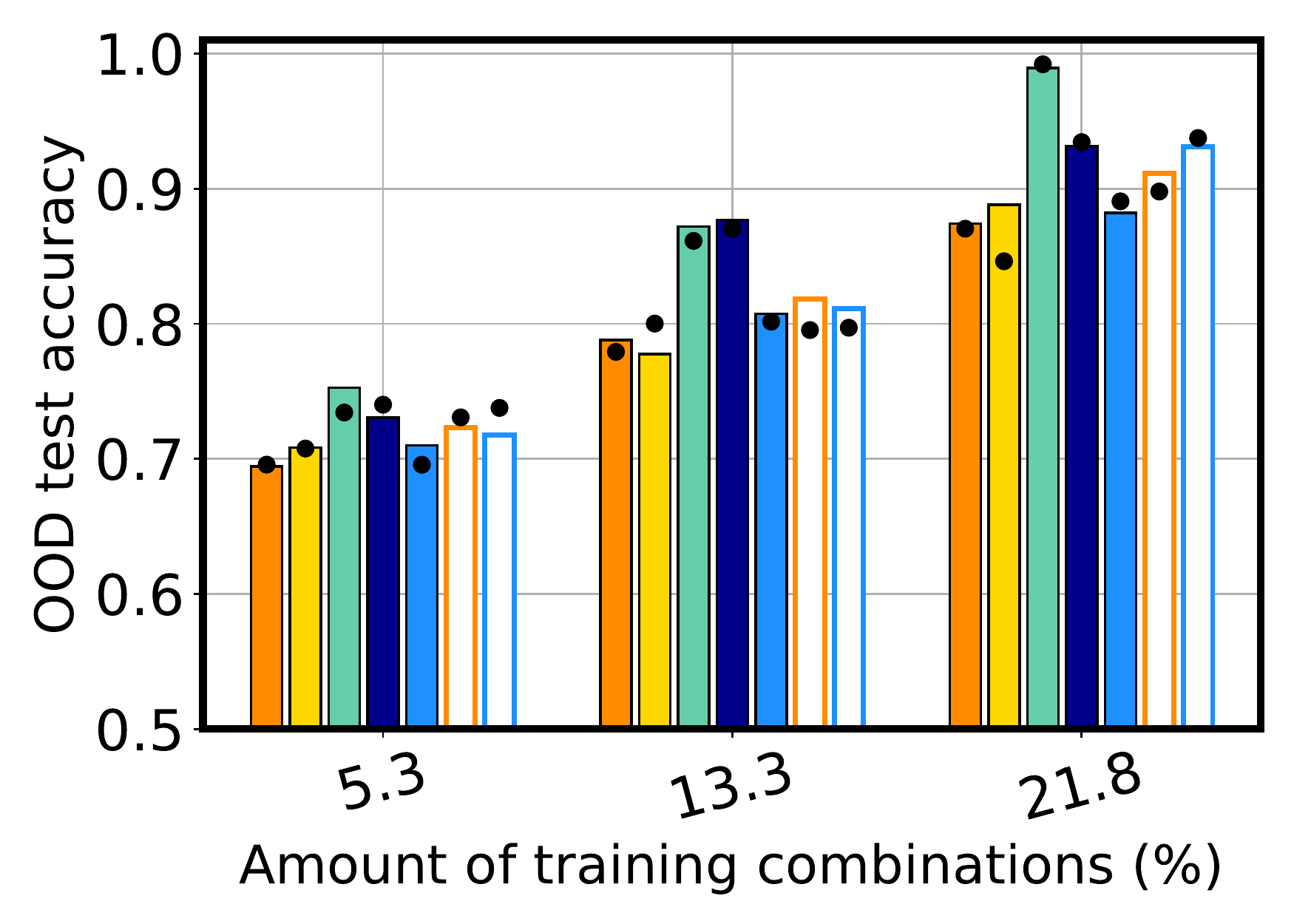} 
    & \hspace{-0.cm} \includegraphics[width=.4\linewidth]{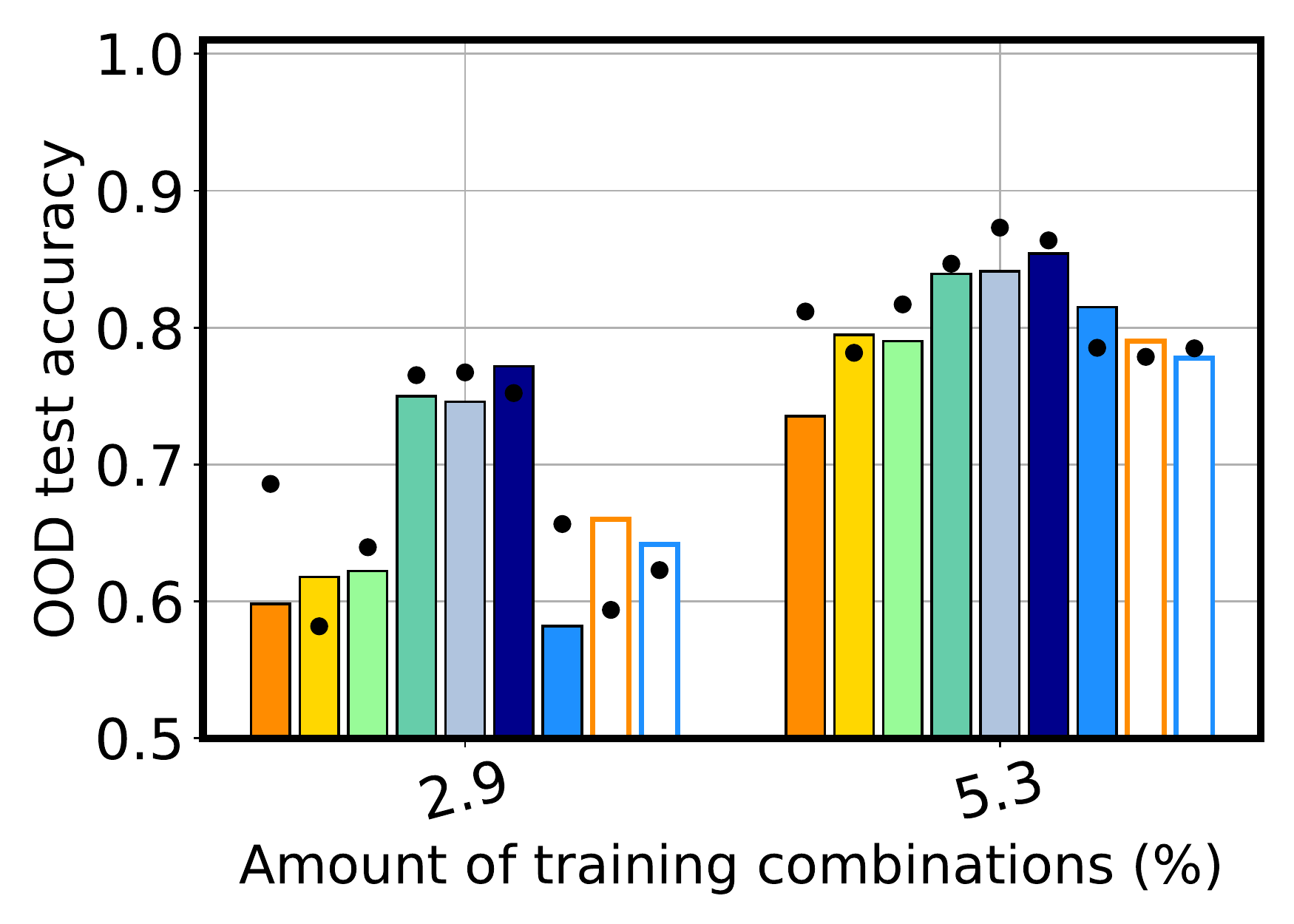} \\ 
    \vspace{-0.05cm}
    (a) & (b) \\
    \end{tabular}
    \caption{Experiments with ten times more data. OOD test accuracy refers to systematic generalization accuracy. The colored bars correspond to networks trained with $2.1$ million images, dots correspond to the case of $210K$ images. Systematic generalization accuracy for (a) attribute extraction from single object, (b) attribute comparison between pairs of separated objects.}
    \label{fig:10x_data}
\end{figure*}

We performed additional experiments to check if the amount of training examples plays a role in the trade-off among degrees of modularity. With this goal, we increased by a factor ten the amount of training examples, in Figure~\ref{fig:10x_data}, for (a) attribute extraction from single object, and  (b) attribute comparison between pairs of separated objects.

The NMNs trained on (a) have a fixed batch size to value 64 and a grid of learning rates with values $[10^{-5}, 10^{-4}, 0.001, 0.005, 0.01]$. We fixed at half a million the amount of iterations.
The NMNs trained on dataset (b) share the same set of hyper-parameters of the case with ten times less data. We fixed the number of iterations to one million, but due to the high computational cost, we ran the experiments for maximum $\sim 300$ hours on a NVIDIA DGX-1 machine. Not all the models reached the last iteration, but all their loss curves qualitatively show convergence. We report the results in Figure~\ref{fig:10x_data}. (a) shows the case of attribute extraction from a single objects, while (b) reports the performance for attribute comparison between pairs of objects.

The colored bars report the systematic generalization performance over one trial, with $2.1$ million training examples, while the black dots display the systematic generalization performance for the corresponding model trained on ten times less examples. We observe that, consistently with the previous results, the \emph{group - group - group} library outperforms the other libraries, in Figure~\ref{fig:10x_data}a.  In Figure~\ref{fig:10x_data}b, the NMN with modular library at the image encoder stage achieve the highest systematic generalization. This result shows that high systematic generalization is not reachable by increasing the amount of training data. 
This conclusion is consistent with the observation in previous works on out-of-distribution generalization~\cite{loula2018rearranging, madan2020capability}.

%% file: A3_cogent.tex
\newpage
\clearpage
\section{Supplemental Results on SQOOP}\label{sec:further_sqoop}

To perform these experiments, we rely on the networks proposed in \cite{bahdanau2018systematic}. Both tree-NMNs make use of batch normalization in their image encoder and classifier stages. 

To validate that NMNs outperform other VQA models in systematic generalization, we report in Table~\ref{tab:sqoop_2objs_all} the results obtained on SQOOP for two non-modular neural networks, namely re-implementation of MAC~\cite{hudson2018compositional} from Bahdanau \emph{et al.}~\cite{bahdanau2018systematic} and FiLM~\cite{perez2018film}, with the results of tree-NMNs we have reported in Table~\ref{table:sqoop_2objs}. We adopted the same hyper-parameters from \cite{bahdanau2018systematic}.

For each model, we observe a wide gap in systematic generalization on the two datasets. For the FiLM architecture, reducing the amount of objects in the image leads to higher systematic generalization than for the case of five objects, while the opposite happens to MAC. 

A direct comparison between non-modular networks and NMN is not possible, given that the NMN are provided with an optimal program layout. Nonetheless, the higher systematic generalization of NMNs highlights the potential benefit in the use of NMNs wherever there is a mechanism to identify a proper program, given a question.

\begin{table}[h!]
  \caption{Systematic generalization accuracy (\%) for MAC (reimplemented by Bahdanau \emph{et al.}~\cite{bahdanau2018systematic}), FiLM, and NMN architectures with tree layouts. Top row: bias in the question (five objects per image). Bottom row: bias in the co-occurrence of objects in the image (two objects per image).}
  \label{tab:sqoop_2objs_all}
  \centering
  \begin{tabularx}{\textwidth}{XXXXX}
    \toprule
    Bias type 
    &\ MAC   \newline \mbox{\footnotesize\ (Bahdanau \emph{et al.}~\cite{bahdanau2018systematic})} 
    & \quad \  FiLM 
    & \mbox{\emph{all - all - all}} \newline \mbox{\footnotesize\ with bn}
    & \mbox{\emph{all - sub-task - all}} \newline \mbox{\footnotesize\ with bn}\\
    \hline
    Five objects per image  \newline \mbox{\footnotesize (as in~\cite{bahdanau2018systematic})}
    & \mbox{\ $84 \pm 4$} 
    & \mbox{\quad \ $70 \pm 2$} 
    & \mbox{\ $\mathbf{99.8 \pm 0.2}$} 
    & \mbox{$\mathbf{99.96 \pm 0.06}$} \\
    \hline
    Two objects per image
    & \mbox{\ $66\pm 5$} 
    & \mbox{\quad \ $77\pm 3$}
    & \mbox{\ ${84\pm2}$}      
    & \mbox{$\mathbf{88.5\pm0.5}$}\\
    \hline
    \bottomrule 
    \end{tabularx}
\end{table}

\newpage
\section{Implementation Details and Supplemental Experiments on CLEVR-CoGenT}\label{app:cogent}
In Appendix~\ref{app:vectornmn} we report the implementation of the Vector-NMN as in \cite{bahdanau2019closure}, with its image encoder, intermediate module, and classifier. Then, in Appendix~\ref{sec:cogent_description} we describe the Compositional Generalization Test (CoGenT), its tasks, and our grouping of sub-tasks, in Appendix~\ref{sec:subtask_division}. Lastly, we report the results obtained on single questions for other non-modular networks, and different NMN implementations.

\subsection{CLEVR-CoGenT}\label{sec:cogent_description}
The Compositional Generalization Test (CLEVR-CoGenT) dataset~\cite{johnson2017clevr} includes training, validation, and test splits. The data generation process follows the usual criteria of CLEVR, but CoGenT is built to explicitly measure the ability of a model to generalize to novel combinations of objects and attributes. CLEVR-CoGenT has two conditions: Condition A and Condition B. In Condition A, the one used at training, cubes, cylinders and spheres appear in different sizes, colors, and positions. Cubes are gray, blue, brown, or yellow, cylinders are red, green, purple, or cyan, while spheres can have any color. In Condition B these combinations of attributes are reversed. Cubes are red, green, purple, or cyan, cylinders are gray, blue, brown, or yellow, and spheres can have any color. Condition B is the split to measure systematic generalization of a model. The models can be validated and tested in-distribution as well.

\subsection{Vector-NMN}\label{app:vectornmn}
We define Vector-NMN as in previous work \cite{bahdanau2019closure}. On the family of CLEVR datasets, the image encoder of Vector-NMN takes in input the output features from a ResNet-101 as detailed in Section 4.1 of \cite{johnson2017clevr}. The image encoder consists of two convolutional layers with ReLU activation, as detailed in Table~\ref{tab:stem_Vector-NMN}, the classifier is a multi-layer perceptron, as detailed in Table ~\ref{tab:classifier_Vector-NMN}.

\begin{table}[!h]
    \centering
    \caption{Image encoder of Vector-NMN.}
    \label{tab:stem_Vector-NMN}
    \begin{tabular}{c|c|c}
    \toprule
      \multicolumn{3}{c}{{Image encoder}} \\
        \toprule
        (0) & Conv2d &  ((1024, 64, kernel\_size=(3, 3), stride=(1, 1), padding=(1, 1)) \\
        \hline
        (1) & ReLU &  \\
        \hline
        (2) & Conv2d & (64, 128, kernel\_size=(3, 3), stride=(1, 1), padding=(1, 1))\\
        \hline
        (3) & ReLU & \\
        \hline
        \bottomrule 
    \end{tabular}
\end{table}

\begin{table}[!h]
    \centering
    \caption{Classifier of Vector-NMN.}
    \label{tab:classifier_Vector-NMN}
    \begin{tabular}{c|c|c}
    \toprule
      \multicolumn{3}{c}{{Classifier}} \\
        \toprule
        (0) & Flatten & \\
        \hline
        (1) & Linear &  (in\_features=128, out\_features=1024, bias=True) \\
        \hline
        (2) & ReLU &  \\
        \hline
        (3) & Linear & (in\_features=1024, out\_features=32, bias=True)\\
        \hline
        \bottomrule
    \end{tabular}
\end{table}

The Vector-NMN leverages on modulation through a set of weights, similarly to the FiLM network~\cite{perez2018film}. Every intermediate module of the Vector-NMN receives in input the concatenation of the embedding for the sub-task and the output from previous modules, defined as $h_c = [\text{Emb}(k); s_x; s_y]$. We define the output of the image encoder given the input features of an image as $s_{\text{img}}$. In the following, we report the definition of Vector-NMN \cite{bahdanau2019closure}:
\begin{align}
    &[\beta_j,\tilde{\gamma}_j] = W_2^j (\text{ReLU}(W_1^j h_c + b_1^j) + b_2^j), \ \text{ with } j\in[0,1],  \label{eq:film_weight}\\
    &\gamma_j = 2\tanh{(\tilde{\gamma}_j)} +1,  \\
    & \tilde{h}_1 = \text{ReLU}(U_1 * (\gamma_1 \odot s_{\text{img}} \oplus \beta_1)), \\    
    & \tilde{h}_2 = \text{ReLU}(U_2 * (\gamma_2 \odot \tilde{h}_1 \oplus \beta_2) + h_x), \\
    & h_i(k, s_x, s_y) = \text{maxpool}(\tilde{h}_2).\label{eq:vectornmn_transf}
\end{align}

\subsection{Grouping of sub-tasks}\label{sec:subtask_division} 
We decide to divide sub-tasks based on their similarity. We group together all the sub-tasks which require determining colors, shapes, materials, etc. We identify eight groups of sub-tasks.
In the following, we specify each group, each with its related sub-tasks: {counting}, with
\texttt{count}, \texttt{equal\_integer}, \texttt{exist}, \texttt{greater\_than}, \texttt{less\_than}, \texttt{unique}; colors, with \texttt{equal\_color}, \texttt{filter\_color[*color instance]}, \texttt{query\_color}, \texttt{same\_color}; {materials}, with \texttt{equal\_material}, \texttt{same\_material}, \texttt{filter\_material[*material instance]}, \texttt{query\_material}; {shapes}, with \texttt{equal\_shape}, \texttt{filter\_shape[*shape instance]}, \texttt{query\_shape}, \texttt{same\_shape}; {sizes}, with \texttt{equal\_size}, \texttt{filter\_size[*size instance]}, \texttt{query\_size}, \texttt{same\_size}; {spatial relations} \texttt{relate[* four possible directions]}; {logical operations} \texttt{intersection}, \texttt{union}; and a group neutral to sub-tasks, with \texttt{scene}.

\subsection{Supplemental Results on  CLEVR-CoGenT}\label{sec:app_cogent_more_models}
We now report the systematic generalization performance on  CLEVR-CoGenT. The results include non-modular networks, as MAC~\cite{hudson2018compositional} re-implemented by Bahdanau \etal~\cite{bahdanau2019closure}, its variant with batch normalization in the image encoder and classifier, the FiLM network~\cite{perez2018film}, the Vector-NMN and the Tensor-NMN, their variant with batch normalization in the image encoder and classifier, and an additional NMN which receives the input $s_{\text{img}}$ at every intermediate step.

We divide the questions into several question types, by considering the first sub-task of the program layout for each question, as in \cite{johnson2017clevr}. In Table \ref{tab:single_question_cogent}, we report separately in a scale from 0 to 100 (\ie percentage) the performance for each model and each question type. It is important to underline that all the NMNs are trained and tested using the ground-truth program layout.
\begin{table}
\centering
    \caption{Systematic generalization accuracy (\%) on validation split of  CLEVR-CoGenT Condition B across different question types.}
    \label{tab:single_question_cogent}
\begin{tabularx}{\textwidth}{XXXX}
\toprule
 & MAC with bn \newline (Bahdanau \etal~\cite{bahdanau2019closure})
 & MAC \newline (Bahdanau \etal~\cite{bahdanau2019closure})
 & FiLM \\
\midrule
    {\texttt{count} }
    &  $  69.5  \pm 0.5$ 
    &  $  68.8 \pm 0.9$ 
    &  $ \mathbf{73.3 \pm} \mathbf{0.9}$ \\
    \hline
    \small{\texttt{equal\_color}}
    &   $ 78.5  \pm 0.8$ 
    &   $ 79.1 \pm 0.7$ 
    &   $ 78.1 \pm 0.7$  \\
    \hline
    \small{\texttt{equal\_integer}} 
    &  $81  \pm 1$ 
    &  $79 \pm 2$ 
    &  $82 \pm 2$  \\
    \hline
    \small{\texttt{equal\_material}}
    & $77  \pm 3$ 
    & $80.6 \pm 0.8$ 
    & $82 \pm 2$  \\
    \hline
    \small{\texttt{equal\_shape}}  
    & $96 \pm 1$
    & $92 \pm 2$ 
    & $\mathbf{97.0 \pm 0.7}$ \\
    \hline
    \small{\texttt{equal\_size }}
    & $77 \pm 2$ 
    & $ 81.3 \pm 0.7$ 
    & $ 81 \pm 1$  \\
    \hline
    \small{\texttt{exist}}
    & $84.9 \pm 0.4$ 
    & $85.1 \pm 0.6$
    & $\mathbf{86.8 \pm 0.4}$ \\
    \hline
    \small{\texttt{greater\_than}}  
    &  $83.4 \pm 0.9$ 
    &  $83 \pm 1$ 
    &  $83.8 \pm 0.5$ \\
    \hline
    \small{\texttt{less\_than}}  
    &  $81.4 \pm 0.4$ 
    &  $81.4 \pm 0.7$ 
    &  $81.8 \pm 0.8$  \\
    \hline
    \small{\texttt{query\_color}}   
    &  $63 \pm 1$ 
    &  $\mathbf{68 \pm 1}$ 
    &  $ 65.5 \pm 0.3$  \\
    \hline
    \small{\texttt{query\_material}}
    &  $85 \pm 1$ 
    &  $87 \pm 1$ 
    &  $87 \pm 1$  \\
    \hline
    \small{\texttt{query\_shape}}    
    &  $34.7 \pm 0.7$ 
    &  $36 \pm 1$ 
    &  $36.2 \pm 0.6$ \\
    \hline
    \small{\texttt{query\_size}}
    &  $ 84 \pm 2$ 
    &  $86 \pm 1$ 
    &  $85.4 \pm 0.5$ \\
    \hline   
\bottomrule
\end{tabularx} 
\begin{tabularx}{\textwidth}{XXXX}
\toprule
 & Tensor simple arch
 & Tensor
 & Tensor  with bn 
 \\
\midrule
    \small{\texttt{count} }
    & $  69.7 \pm 0.5$ 
    & $  69.7 \pm 0.8$ 
    & $  70.9 \pm 0.7$ 
    \\
    \hline
    \small{\texttt{equal\_color} }
    &  $ 76 \pm 2$ 
    &  $ 75.6 \pm 0.8$
    &  $ 78.0 \pm 0.7$ 
    \\
    \hline
    \small{\texttt{equal\_integer}}
    &   $83 \pm 1$ 
    &   $82.7 \pm 0.3$ 
    &   $83.3 \pm 0.7$ 
    \\
    \hline
    \small{\texttt{equal\_material}}
    &    $74 \pm 2$ 
    &    $74 \pm 2$ 
    &    $75 \pm 1$ 
    \\
    \hline
    \small{\texttt{equal\_shape}}  
    &  $92 \pm 2$ 
    &  $91 \pm 2$ 
    &  $92 \pm 3$ 
    \\
    \hline
    \small{\texttt{equal\_size}}     
    &   $76 \pm 3$ 
    &   $75 \pm 1$ 
    &   $77 \pm 1$
    \\
    \hline
    \small{\texttt{exist}}          
    &  $83.8 \pm 0.3$ 
    &  $84.2 \pm 0.4$ 
    &  $84.9 \pm 0.2$ 
    \\
    \hline
    \small{\texttt{greater\_than}}   
    &   $84.4 \pm 0.5$ 
    &   $83.8 \pm 0.6$ 
    &   $84.4 \pm 0.5$ 
    \\
    \hline
    \small{\texttt{less\_than}}
    &   $82.2 \pm 0.3$ 
    &   $80.7 \pm 0.9$ 
    &   $82 \pm 1$ 
    \\
    \hline
    \small{\texttt{query\_color}}    
    &    $59 \pm 1$ 
    &    $58 \pm 1$ 
    &    $62.6 \pm 0.9$ 
    \\
    \hline
    \small{\texttt{query\_material}} 
    &   $85 \pm 1$ 
    &   $84.1 \pm 0.9$ 
    &   $85.7 \pm 0.7$ 
    \\
    \hline
    \small{\texttt{query\_shape}}    
    &  $37 \pm 2$ 
    &  $37 \pm 1$ 
    &  $37 \pm 1$ 
    \\
    \hline
    \small{\texttt{query\_size}}
    &   $85 \pm 1$ 
    &    $83.5 \pm 0.6$ 
    &   $84.9 \pm 0.8$ 
    \\
    \hline   
\bottomrule
\end{tabularx}
\begin{tabularx}{\textwidth}{XXXX}
\toprule
 & Vector 
 & Vector with bn
 & Vector \newline modular (\textbf{ours}) \\
\midrule
    \small{\texttt{count} }
    & $  70.4 \pm 0.4$ 
    & $ 71.5 \pm 0.9 $
    & $  71\pm 1$ \\
    \hline
    \small{\texttt{equal\_color} }
    &  $ 74 \pm 1$ 
    &  $76 \pm 2$ 
    &  $ \mathbf{80 \pm 1}$ \\
    \hline
    \small{\texttt{equal\_integer}}
    &   $78 \pm 2$ 
    &   $81.6 \pm 0.9$
    &   $\mathbf{85 \pm 2}$ \\
    \hline
    \small{\texttt{equal\_material}}
    &   $74.2 \pm 0.7$ 
    &   $77 \pm 3 $
    &   $\mathbf{84 \pm 2}$ \\
    \hline
    \small{\texttt{equal\_shape}}  
    &  $89 \pm 3$ 
    &  $86 \pm 2 $
    &  $79 \pm 2$ \\
    \hline
    \small{\texttt{equal\_size}}     
    &   $75 \pm 1$
    &   $78 \pm 3$
    &   $\mathbf{88 \pm 2}$ \\
    \hline
    \small{\texttt{exist}}          
    &  $84.4 \pm 0.4$ 
    &  $85 \pm 0.7$ 
    &  $84.4 \pm 0.5$ \\
    \hline
    \small{\texttt{greater\_than}}   
    &   $83.6 \pm 0.4$ 
    &   $84 \pm 1 $
    &   $\mathbf{89 \pm 1}$ \\
    \hline
    \small{\texttt{less\_than}}
    &   $82.0 \pm 0.5$ 
    &   $83 \pm 1$
    &   $\mathbf{87 \pm 2}$ \\
    \hline
    \small{\texttt{query\_color}}    
    &    $60 \pm 1$ 
    &    $66 \pm 5 $
    &    $67 \pm 4$ \\
    \hline
    \small{\texttt{query\_material}} 
    &   $84.7 \pm 0.4$
    &   $86 \pm 2$
    &   $\mathbf{88.2 \pm 0.8}$ \\
    \hline
    \small{\texttt{query\_shape}}    
    &  $40 \pm 3$ 
    &  $41 \pm 2$
    &  $\mathbf{52 \pm 3}$ \\
    \hline
    \small{\texttt{query\_size}}
    &  $84.7 \pm 0.7$ 
    &  $86 \pm 2$
    &  $\mathbf{89.5 \pm 0.5}$ \\
    \hline   
\bottomrule
\end{tabularx}
\end{table}

The best network for each question type is highlighted in bold. On nine over thirteen question types, our Vector-NMN with modular image encoder has higher performance than pre-existing modular networks, given the same ground-truth program layout. Modularity of the image encoder is particularly convenient on the hardest type of questions, as \texttt{query\_shape} and \texttt{query\_color}. 

\newpage
\section{Supplemental Results on  CLOSURE}\label{app:closure_results}
CLOSURE is a novel test set for models trained on the CLEVR dataset~\cite{bahdanau2019closure}. This dataset introduces seven question templates that have high overlap to the CLEVR questions but zero probability of appearing under the CLEVR data distribution. 

We trained Tensor-NMN, Vector-NMN and our Vector-NMN with modular image encoder on the CLEVR dataset by keeping the hyper-parameters (learning rate, batch size, number of iterations, network's number of layers and transformations) as in~\cite{bahdanau2019closure}. For our Vector-NMN, the separation of sub-tasks into groups is specified in Section~\ref{sec:clevr_systematic}. 

The results in Table~\ref{table:closure} shows the performance of our Vector-NMN with modular image encoder compared to the original Vector-NNM and Tensor-NMN, for novel questions templates of CLOSURE.

\begin{table}[h!]
  \caption{Systematic generalization accuracy (\%) on each question type of CLOSURE for Tensor-NMN and Vector-NMN (as reported in \cite{bahdanau2019closure}), and for our Vector-NMN, across five repetitions.}
  \label{table:closure}
  \centering
  \begin{tabularx}{\textwidth}{XXXX}
  \toprule
    & Tensor-NMN & Vector-NMN & Vector-NMN \newline with modular \newline image encoder (ours) 
    \\
    \hline
    \texttt{and\_mat\_spa}
    & $64.9 \pm 2$
    & $\mathbf{86.3 \pm 2.5}$ 
    & $65.8 \pm 2.4$  \\
    \hline
    \texttt{or\_mat }  
    & $44.8 \pm 6.8$ 
    & $\mathbf{91.5 \pm 0.77}$ 
    & $70.3 \pm 3.3$ \\
    \hline
    \texttt{or\_mat\_spa  } 
    & $47.9 \pm 5.8$ 
    & $\mathbf{88.6 \pm 1.2}$ 
    & $65.2 \pm 4.5$ \\
    \hline
    \texttt{embed\_spa\_mat  } 
    & $98.1 \pm 0.38$ 
    & $\mathbf{98.5 \pm 0.13}$ 
    & $88.6 \pm 1.8$ \\
    \hline
    \texttt{embed\_mat\_spa }  
    & $79.3 \pm 0.83$ 
    & $\mathbf{98.7 \pm 0.19}$ 
    & $94.0 \pm 1.4$ \\
    \hline
    \texttt{compare\_mat }  
    & $90.7 \pm 1.8$ 
    & $\mathbf{98.5 \pm 0.17}$ 
    & $89.8 \pm 1.7$ \\
    \hline
    \texttt{compare\_mat\_spa}
    & $91.2 \pm 1.9$ 
    & $\mathbf{98.4 \pm 0.3}$ 
    & $90.2 \pm 2.1$ \\
    \hline
    \bottomrule
  \end{tabularx}
\end{table}

The Vector-NMN outperforms our modular Vector-NMN across all questions. These two models only differ at the image encoder stage, which seem to suggest that a limited distribution of questions can have a different effect on modularity from what we observed in previous experiments. Note that this experiment is the only experiment in the paper in which modularity at the image encoder stage does not improve systematic generalization. This experiment is also the only one in which the distribution of program layouts in the training and testing is different. 
This is an interesting phenomenon which requires of further analysis (it is unclear how general this phenomenon is, what are the trends for different amounts of bias, etc.). A possible explanation is that having more modules leads to a more diverse set of possible networks, and if there is bias in the program layout, the difference between the trained networks and tested ones may become larger. This suggests that there is a trade-off between modularity and bias in the program layout.

\newpage
\section{Code and Computational Cost}\label{app:computational_cost}
All the code for data generation, data loading, training, testing, and the networks trained on CLEVR-CoGenT can be found at this link: \url{https://github.com/vanessadamario/understanding_reasoning.git}. This work stems from two forked repositories which investigated systematic generalization in VQA: \url{https://github.com/rizar/systematic-generalization-sqoop.git} \cite{bahdanau2018systematic}  and \url{https://github.com/rizar/CLOSURE.git} \cite{bahdanau2019closure}. Our code, as the one from which we took inspiration, is open access and reusable.

The experiments have been run on multiple platforms (AWS + MIT's OpenMind Computing Cluster\footnote{\url{https://openmind.mit.edu/}}).

\paragraph{VQA-MNIST}
Across all the experiments with $210$K training examples, the number of epochs is always higher than $60$ (more specifically, $(200000*64)/210000 > 60$). Across all the experiments, we observed that the training converged for such number of epochs. 

The resources and amount of time for  VQA-MNIST experiments varies: with larger batch size, the experiments of attribute extraction from single object (experiment\_1) took around ten days to complete the training of each data-run (NVIDIA Tesla K80, NVIDIA Tesla K40, NVIDIA TitanX). 

We moved the experiments for attribute comparison  and attribute extraction from multiple objects to AWS. We further parallelized those using Ray (\url{https://ray.io/}), a parallelization framework. 
On AWS, we have been using 60 instances of g4dn.2xlarge, with 32 GB RAM, Intel Custom Cascade Lake CPU, NVIDIA T4 GPU. Training a data-run for these experiments took less than 48 h.

For the experiments with $2.1$M training examples, we use an NVIDIA DGX-1. For the attribute extraction from a single object ($500$K iterations, batch size $64$), training all models (for grid search) in parallel (through Ray we were training ten NMNs at the time) took about two weeks of computation. For the attribute comparison between separated objects ($1M$ iterations, batch size $64$), training all models (for grid search) in parallel (ten experiments at the time), was taking more than two weeks. Experiments that took longer than that time were interrupted, but all the training curves reached a plateau at that point.

\paragraph{SQOOP}
The experiments on SQOOP datasets took less than $15$ hours on a single GPU (NVIDIA Tesla K80, NVIDIA Titan X).

\paragraph{CLEVR datasets}
The experiments on the CLEVR dataset have an average training time of a week, which varies depending on the network. These models are trained on single GPU (NVIDIA Tesla K80, NVIDIA Tesla K40, NVIDIA Titan X).